\pdfoutput=1

\documentclass[11pt]{article}

\usepackage{acl}

\usepackage{times}
\usepackage{latexsym}

\usepackage[T1]{fontenc}

\usepackage[utf8]{inputenc}

\usepackage{microtype}

\usepackage{inconsolata}

\usepackage{graphicx}
\usepackage{booktabs}
\usepackage{multirow}
\usepackage{subcaption}
\usepackage{tabularx}
\usepackage{CJKutf8}
\usepackage[whole]{bxcjkjatype}
\usepackage{tabularx}
\usepackage{caption}
\usepackage[T1]{fontenc}
\usepackage{textcomp}

\title{On the Multilingual Ability of Decoder-based Pre-trained Language Models: Finding and Controlling Language-Specific Neurons}

\author{%
  Takeshi Kojima, Itsuki Okimura, Yusuke Iwasawa,
  \\
  {\bf Hitomi Yanaka}, {\bf Yutaka Matsuo}
  \\
  The University of Tokyo
  \\
  \texttt{t.kojima@weblab.t.u-tokyo.ac.jp}
}

\begin{document}
\maketitle

\renewcommand{\arraystretch}{0.7}

\begin{abstract}
Current decoder-based pre-trained language models (PLMs) successfully demonstrate multilingual capabilities. 
However, it is unclear how these models handle multilingualism.
We analyze the neuron-level internal behavior of multilingual decoder-based PLMs, 
Specifically examining the existence of neurons that fire ``uniquely for each language'' within decoder-only multilingual PLMs.
We analyze six languages: English, German, French, Spanish, Chinese, and Japanese, and show that language-specific neurons are unique, with a slight overlap (< 5\%) between languages. These neurons are mainly distributed in the models' first and last few layers. 
This trend remains consistent across languages and models.
Additionally, we tamper with less than 1\% of the total neurons in each model during inference and demonstrate that tampering with a few language-specific neurons drastically changes the probability of target language occurrence in text generation.\footnote{Code and model-generated texts are available at \url{https://github.com/kojima-takeshi188/lang_neuron}}
\end{abstract}


\section{Introduction}
Recent studies have frequently demonstrated the excellent multilingual abilities of pre-trained language models (PLMs).
%
Some PLMs explicitly mix multilingual language corpus for pre-training \citep{lin2021few,scao2022bloom}, whereas others mainly use an English-dominant text corpus, with the unintentional inclusion of a low percentage of multiple language texts, which results in the acquisition of multilingual skills, such as Llama2 \citep{touvron2023llama}. 
How do they exhibit multilingual abilities?

\begin{figure}[t]
\includegraphics[width=\columnwidth]{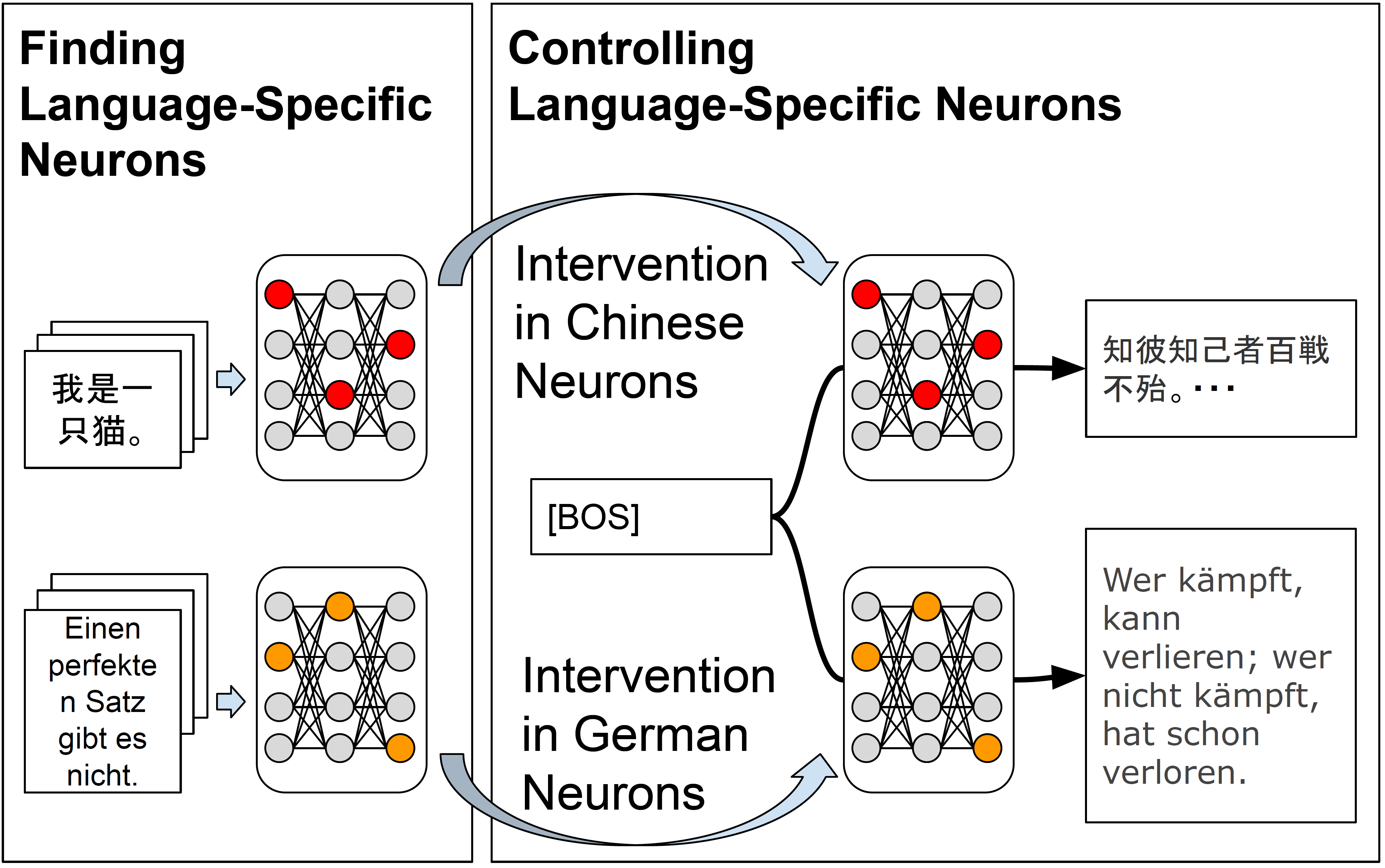}
\caption{
Overview of our proposal.
(Left) Finding language-specific neurons that tend to be activated for a target language. (Right) Controlling the detected language-specific neurons by forcing their activation during inference to manipulate the probability of target language occurrence.
}
\label{fig:overall}
\end{figure}

Prior studies have focused on language-universal neurons activated across multilingual inputs, mainly focusing on encoder-based PLMs~\citep{antverg2021pitfalls, stanczak2022same,chen2023journey,stanczak2023latent,varda2023data}. 
In contrast to encoder-based models, which might be sufficient to abstract inputs, decoder-based PLMs need to recover the language-specific information in the later part of the generation. Therefore, language-specific processing within these models should be a more complex and essential functionality compared to the encoder-based ones.
However, few studies have focused on the existence and activity of language-specific neurons in decoder-based PLMs (See Section \ref{sec:related_work}).

This study examines the behavior of language-specific neurons in decoder-based PLMs.
Specifically, we analyze multiple decoder-based PLMs, including XGLM (564M, 1.7B, 2.9B), BLOOM (560M, 1.7B, 3B), and Llama2 (7B, 15B), for six languages (English, German, French, Spanish, Chinese, and Japanese).
To investigate language-specific neurons, we adopt an approach proposed by
\citet{cuadros2022self}, which finds neurons that activate on a certain group of sentences (\textit{Positive} sentences) but do not activate on other groups (\textit{Negative} sentences).
We treat the target language texts as positive and any other language as negative, identifying language-specific neurons that statistically activate positive sentences (See Section \ref{sec:method}).
The experimental results demonstrate that the identified language-specific neurons are mainly distributed in the first and last few layers of the model. 
This trend remains consistent across multiple languages and model variants.
To verify the effect of the neurons, we intervene in language-specific neurons in the model during inference, showing that they can drastically change the probability of the target language occurrence during text generation (See Section \ref{sec:experiment}).

\section{Related Work}
\label{sec:related_work}




Previous studies analyzed the internal behavior of PLMs for multilingual tasks by observing the activation of their neurons.
Several studies have found that language-universal neurons are activated across multilingual inputs in encoder-based PLMs (mainly M-BERT, see \citet{pires2019multilingual}) for various task settings, including syntactic  or factual knowledge tasks \citep{antverg2021pitfalls,stanczak2022same,chen2023journey,stanczak2023latent,varda2023data}.
However, studies of encoder-based PLMs have not focused on the identification of language-specific neurons.
\citet{mueller2022causal,bau2018identifying} studied decoder-based language models to find the multilingually shared neurons.
Similar to the encoder-based PLMs, limited research has focused on the existence and activity of language-specific neurons in decoder-based language models.

Various methods can be used to identify and control neurons \citep{sajjad2022neuron}.
Several studies have identified and intervened in neurons for effective word editing or classification \citep{mu2020compositional,dai2021knowledge,mueller2022causal,chen2023journey,varda2023data}.
In contrast, few studies have investigated the identification and intervention of neurons for full-text generation for a desired concept, for example, \citet{bau2018identifying} for LSTM models and \citet{cuadros2022self} for pretrained transformer models.
\citet{cuadros2022self} have proposed an approach for controlling text generation on Transformer-based decoder models, and proven its effectiveness. Therefore, we use it as an analytical tool in our experiments with decoder-based PLMs.

\section{Method}
\label{sec:method}

We identified and controlled neurons specific to each language based on the approach of \citet{cuadros2022self}, with appropriate modifications for our experiments.
This approach was originally developed to identify and control neurons that respond to specific word-level concepts, such as homographs or gender biases.
However, we aimed to find neurons that grasp broader sentence-level and language-specific concepts; therefore, we modified the original approach for our purpose.

\subsection{Finding Language-specific Neurons}


First, we prepared  text for each language.
We considered a set of $|L|$ languages.
For each language $l \in L$, we prepared $N_l$ texts,
which resulted in $N = N_1 + ... + N_l + ... + N_{|L|}$ texts for all the languages.
Let $x = \{x_i\}^N_{i=1}$ be the set of all the texts.
Our goal was to find neurons that activate text in the target language $l_t \in L$ but do not activate text in other languages $L\setminus \{l_t\}$.
For each text $x_i \in x$, we assigned a label $b_i=1$ if the text was in the target language (i.e., $l = l_t$); otherwise, $b_i=0$.
Therefore, we had 
\begin{equation}
N = N^+_{l_t} + N^-_{l_t}
\end{equation}
sentences, where $N^+_{l_t}$ positive sentences consisted of texts in the target language $l$ (i.e., $b_i=1$) and $N^-_{l_t}$ negative sentences consisted of texts in other languages (i.e., $b_i=0$).
For example, if the target language $l_t$ was French, French texts were assigned label 1, and texts in other languages, such as German and Chinese were assigned label 0.

Second, we observed the activation value of each neuron inside the model given the input text.
We assigned a unique index $m \in M$ to each neuron. $|M|$ denotes the total number of neurons in the model.
Let $z_{m,i} \in z_{m}$ be the output value of neuron $m$ when text $x_i \in x$ is provided as an input to the model.
Here, we explain in detail how this value can be calculated.
Specifically, text $x_i$ is composed of a sequence of $T$ tokens $x_i = \{w_{i,1}, ..., w_{i,t}, ..., w_{i,T}\}$.
Therefore, given the input text, there exist $T$ output values $\{z_{m,i,1}, ..., z_{m,i,j}, ... z_{m,i,T}\}$ for neuron $m$ inside the decoder-based Transformer model.
We take the average of the $T$ neuron outputs to summarize the output value of neuron $m$ for the text $i$.
\begin{equation}
z_{m,i} = f(z_{m,i,1}, ..., z_{m,i,t}, ... z_{m,i,T}),
\end{equation}
where $f$ is the aggregation function of the average operator.
While the original approach \citep{cuadros2022self} defines $f$ as a max-pooling operator, our approach defines $f$ as an average operator to identify neurons that consistently activate across tokens for language identification purposes.
The output values of the [PAD] token position are excluded from the aggregation as an exception because they are regarded as noise.

Third, language-specific neurons were identified.
We regarded the dataset $\{x_i, b_i, z_{m,i}\}^N_{i=1}$ as the prediction task samples. Specifically, we regarded texts $\{x_i\}^N_{i=1}$ as inputs to the model, labels $\{b_i\}^N_{i=1}$ as their ground truth, and the output values of neurons $\{z_{m,i}\}^N_{i=1}$ as the prediction scores of the ground truth.
We can measure the performance of neuron $m$ for the task using its average precision ($AP_m = AP(z_m, b) \in [0, 1]$), which is the area under the precision-recall curve with different prediction thresholds.
We measured $AP_m$ for all neurons and ordered them in descending order.

In the original approach, only the top-$k$ neurons in descending order were defined as identified neurons.
However, this only considers strong positive correlations (i.e., the top-$k$ highest AP neurons) with labels, leaving out strong negative correlations (i.e., the top-$k$ lowest AP neurons) with labels.
We hypothesize that not only the top-$k$ neurons but also the bottom-$k$ neurons are strongly related to a specific language.
Therefore, we extended the original approach by considering not only the top-$k$ neurons but also the bottom-$k$ neurons, defining them as language-specific neurons.
We validate our claim experimentally in Section \ref{sec:experiment}.
We set $k=1000$ as the default value across the experiments.
Note that the neurons at the input layer (word embeddings) and output layer (projection layers) were excluded from the measurement because it is clear that these layers consist of language-specific modules: they consist of language-specific characters or (sub-)words.

\subsection{Controlling Language-specific Neurons}
We controlled text generation by overriding the output values of the top-$k$ and bottom-$k$ neurons with fixed values during inference.
Specifically, we calculated the fixed value for each neuron $m$ as follows: 
\begin{equation}
\bar{z}_{m} = Median(\{z_{m}|b=1\}).
\end{equation}
This is the median of the neuron outputs for the target language texts.
During inference, we intervened in the top-$k$ and bottom-$k$ neurons by replacing their outputs with fixed values in the forward pass and observed whether the models generated texts in the target language.

\section{Experiment Settings}
\label{sec:experiment}

\subsection{Models}

XGLM \citep{lin2021few}, BLOOM \citep{scao2022bloom}, and Llama2 \citep{touvron2023llama} were used in the experiments.
XGLM and BLOOM are explicitly referred to as multilingual language models. 
By contrast, Llama2 was trained almost entirely on an English text corpus, with minimal inclusion of other languages. 
Table \ref{tab:model_list} lists the models used in the experiments. All the models were downloaded from HuggingFace \citep{wolf2019huggingface}. Table \ref{tab:model_language} describes the distribution of languages in the pretraining dataset for each model\footnote{XGLM information is cited from \url{https://huggingface.co/facebook/xglm-2.9B}. BLOOM information is cited from \url{https://huggingface.co/bigscience/bloom\#languages}.}.

\begin{table}[t]
\centering
\begin{tabular}{lrrrr}\toprule
Model &\# Params &\# Layers &\# Neurons \\\midrule
XGLM &564M &24 &221,184 \\
&1.7B &24 &442,368 \\
&2.9B &48 &884,736 \\
\midrule
BLOOM &560M &24 &221,184 \\
&1.7B &24 &442,368 \\
&3B &30 &691,200 \\
\midrule
Llama2 &7B &32 &1,359,872 \\
&13B &40 &2,129,920 \\
\bottomrule
\end{tabular}
\caption{Model list used for the experiments.}
\label{tab:model_list}
\end{table}
\begin{table}[t]
\centering
\scalebox{0.95}[0.95]{
\begin{tabular}{lrrrrrrr}\toprule
&en &de &fr &es &zh &ja \\\midrule
XGLM &49.0 &5.4 &4.7 &5.3 &8.1 &4.0 \\
BLOOM &30.0 &- &12.9 &10.8 &16.2 &- \\
Llama2 &89.7 &0.2 &0.2 &0.1 &0.1 &0.1 \\
\bottomrule
\end{tabular}
}
\caption{Distribution of languages in pre-training data.}
\label{tab:model_language}
\end{table}

\subsection{Datasets}

The following six languages were used in the experiment: English (en), German (de), French (fr), Spanish (es), Chinese (zh), and Japanese (ja).
These six languages are frequently targeted in prior studies of multilingual language models; they are among the top seven languages in terms of the percentage of languages included in the XGLM pre-training data \citep{lin2021few}.
Owing to the limitations of the available computer resources, the number of languages analyzed was limited to six, as described in the limitations section.

To create a language-specific text corpus, we combined two datasets, PAWS-X \citep{yang-etal-2019-paws} and FLORES-200 \citep{costa2022no}. 
PAWS-X is a dataset for paraphrase identification between two texts for seven languages, including the aforementioned languages. 
FLORES-200 is a dataset of machine translation tasks for more than 200 languages.
The sample sentences in these tasks were of good quality, had a wide variety of text types, and covered the six languages required for our experiments. Therefore, a combination of these factors was used.
For this experiment, texts were randomly sampled in a 1:1 ratio from the two datasets to create ground- truth texts for each language. 

Following \citet{cuadros2022self}, we apply a setting of $N^-_{l_t} > N^+_{l_t}$ to account for the much larger variance of negative than positive examples. Negative samples contain texts from five language, whereas positive samples contain texts from only one language.
Specifically, we prepared 500 texts for each language, totaling 3000 texts for all six languages.
As \citet{cuadros2022self} pointed out, the choice of $N^+_{l_t}$ and $N^-_{l_t}$ is arbitrary, usually a tradeoff between the computing resources available and the quality of the representation required. In addition, \citet{cuadros2022self} set the sample sizes of both positive and negative examples between 100 and 1000. Therefore, we considered 500 to be a reasonable value.
After specifying a target language, we identified language-specific neurons for the target language using the method described in Section \ref{sec:method}.

\begin{figure}[t]
\begin{center}
\includegraphics[width=1.0\linewidth]{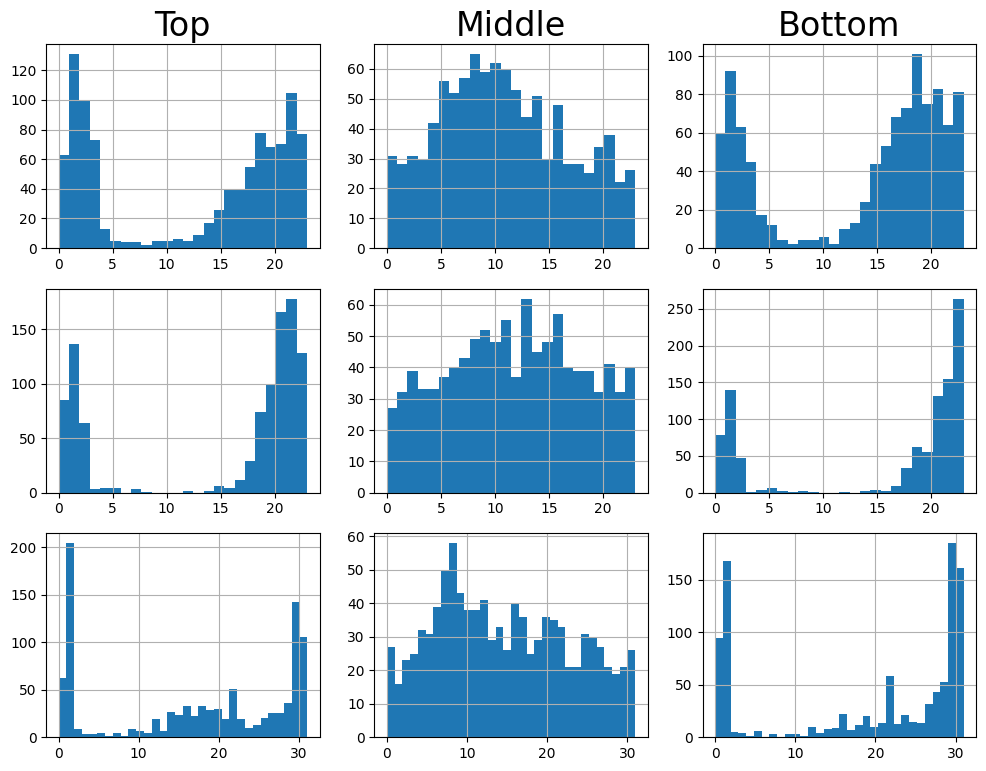}
\caption{Distribution of Top, Middle, Bottom-1000 neurons across layers.
1st row:(XGLM-564M, de). 
2nd row:(BLOOM-1b7, fr). 
3rd row:(Llama2-13b, zh).
}
\label{fig:histogram}
\end{center}
\end{figure}

\begin{table}[t]\centering
\scalebox{0.95}[0.95]{
\begin{tabular}{lrrrrrrr}\toprule
&de &en &es &fr &ja &zh \\\midrule
de &2000 &41 &74 &39 &44 &34 \\
en &41 &2000 &34 &41 &49 &40 \\
es &74 &34 &2000 &57 &77 &22 \\
fr &39 &41 &57 &2000 &21 &93 \\
ja &44 &49 &77 &21 &2000 &27 \\
zh &34 &40 &22 &93 &27 &2000 \\
\bottomrule
\end{tabular}
}
\caption{The number of overlapping language-specific neurons between languages (XGLM-564M).}
\label{tab:cross_check}
\end{table}

\section{Results and Discussion}
\label{sec:result}

The experimental results are summarized in this section.
See the Appendix for the full results across all models and languages.

\subsection{Finding Language-specific Neurons}
\label{sec:result_find}

We identified language-specific neurons using the method described in Section \ref{sec:method}.
Figure \ref{fig:histogram} shows histograms of the identified neurons for each layer in each model. 
Most of the top-1000 neurons with higher AP orders are distributed in the first and last few layers of the model. 
Similarly, most of the bottom-1000 neurons were distributed in the first and last few layers. 
In contrast, the middle-1000 neurons (around the median) in AP order were distributed mainly in the middle layers.
It was also found that the distributional property was the same across languages, model sizes, and model variants.

In addition, it was confirmed that language-specific neurons that fire in one language have little overlap with those of all other languages. 
Table \ref{tab:cross_check} shows the cross tables of the number of overlapping neurons between the six languages, indicating an overlap of less than 5 \% for every language pair.

The following interpretation is possible regarding the inner behavior of decoder-based PLMs based on the findings in Figure \ref{fig:histogram} and the prior study of multilingual models \citep{muller2021first}, which suggests that encoder-based PLMs process cross-lingual transfer in lower layers and language-agnostic semantic concepts (e.g., task prediction) in higher layers:
The first few layers of decoder-based PLMs mainly process cross-lingual transfers to transform the lexical or syntax representations of each language into language-independent semantic representations.
The middle layers of the models are primarily language-independent semantic understanding and representation processing.
The last few layers of the models translate the semantic representations back into syntax and lexical information for the target language.
This interpretation aligns with a recent study \citep{wendler2024llamas}, which suggested that the last few layers of Llama2 models are responsible for the conversion process to a target language.


\begin{figure*}[t]
  \centering
  \begin{minipage}[t]{1.0\linewidth}
      \centering
        \includegraphics[width=1.0\linewidth]{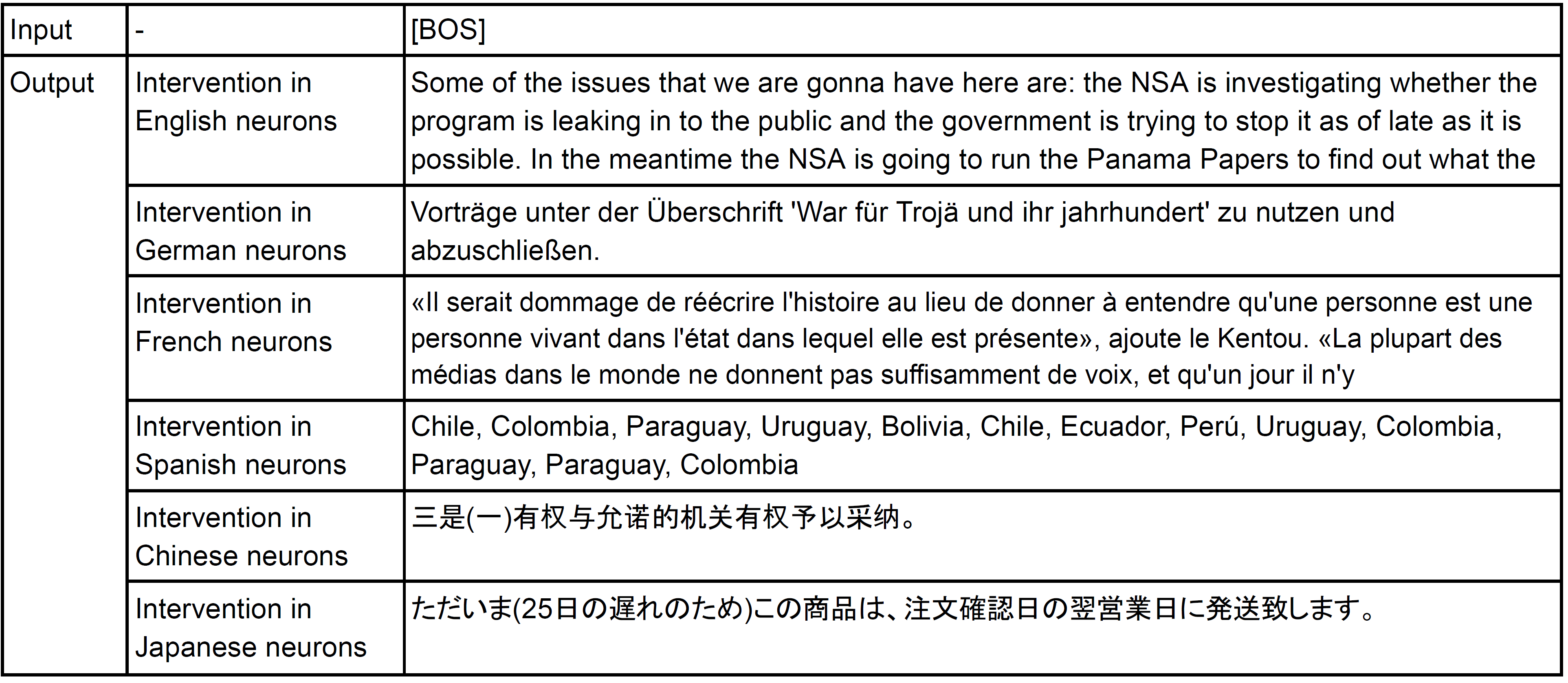}
        \vspace{-7mm}
        \caption{Model-generated text examples with unconditional text generation setting by XGLM-564M. Given a [BOS] token as input, the model generates outputs through a random sampling method.}
        \vspace{3mm}
        \label{main_text_sample_unconditional}
  \end{minipage}
  \begin{minipage}[t]{1.0\linewidth}
        \centering
        \includegraphics[width=1.0\linewidth]{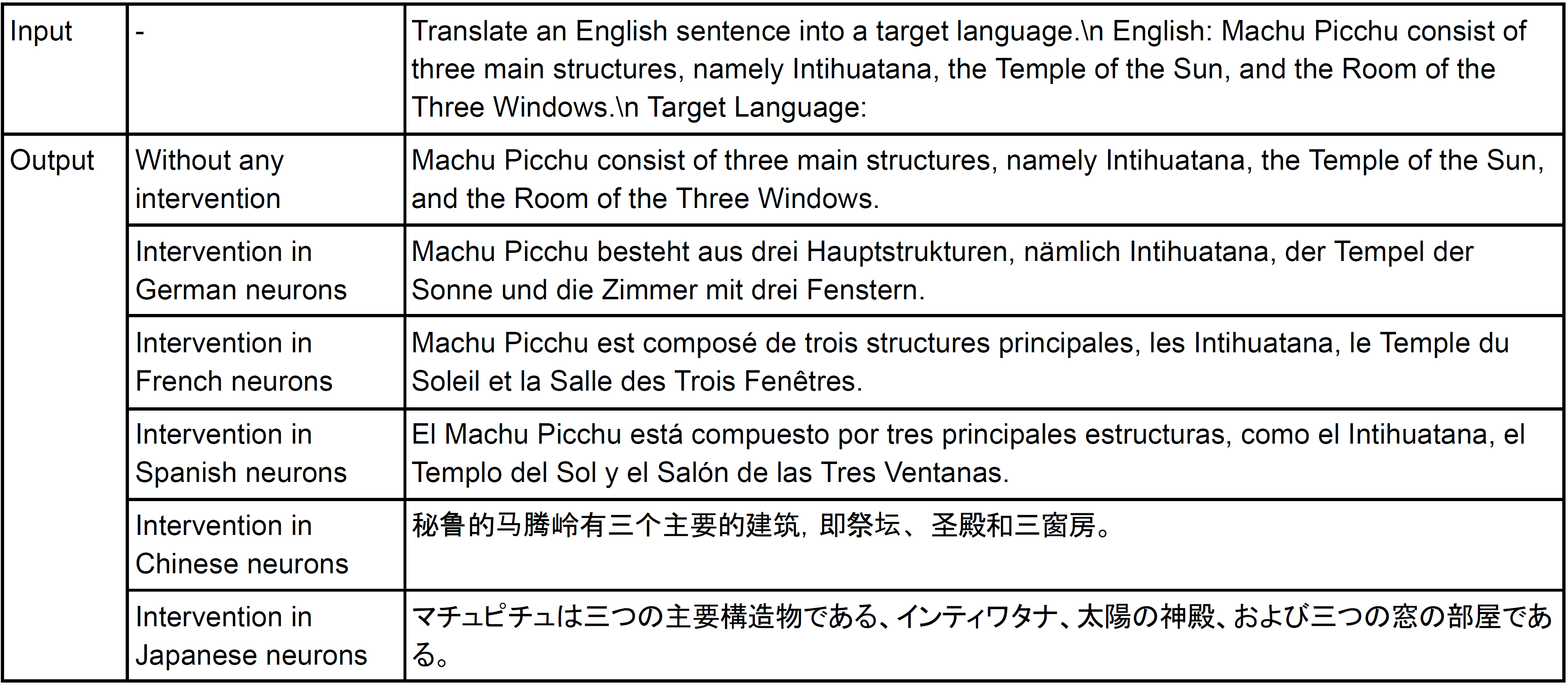}
        \vspace{-7mm}
        \caption{Model-generated text examples with conditional text generation setting by Llama-2-7b. Given a machine translation task as input, the model generates outputs through a greedy decoding method.}
        \label{main_text_sample_conditional}
  \end{minipage}
\end{figure*}

\begin{table}[t]\centering
\scalebox{0.96}[0.96]{
\begin{tabular}{lrrrrrr}\toprule
& &before &\multicolumn{3}{c}{after} \\\cmidrule{3-6}
& & &Top &Bottom &Both \\\midrule
XGLM &en &40.0 &62.0 &77.0 &\textbf{89.0} \\
(564M) &de &0.0 &89.0 &31.0 &\textbf{95.0} \\
&fr &0.0 &86.0 &7.0 &\textbf{90.0} \\
&es &2.0 &71.0 &5.0 &\textbf{78.0} \\
&zh &7.0 &\textbf{82.0} &50.0 &79.0 \\
&ja &7.0 &92.0 &61.0 &\textbf{99.0} \\
&- &9.3 &80.3 &38.5 &\textbf{88.3} \\
\midrule
BLOOM &en &37.0 &78.0 &67.0 &\textbf{88.0} \\
(1b7) &de &0.0 &60.0 &0.0 &\textbf{86.0} \\
&fr &13.0 &80.0 &72.0 &\textbf{98.0} \\
&es &18.0 &44.0 &94.0 &\textbf{97.0} \\
&zh &6.0 &1.0 &89.0 &\textbf{90.0} \\
&ja &0.0 &67.0 &35.0 &\textbf{97.0} \\
&- &12.3 &55.0 &59.5 &\textbf{92.7} \\
\midrule
Llama2 &en &83.0 &82.0 &\textbf{89.0} &\textbf{89.0} \\
(7b) &de &0.0 &2.0 &6.0 &\textbf{23.0} \\
&fr &2.0 &1.0 &\textbf{8.0} &7.0 \\
&es &1.0 &4.0 &4.0 &\textbf{35.0} \\
&zh &0.0 &2.0 &4.0 &\textbf{50.0} \\
&ja &1.0 &1.0 &\textbf{12.0} &10.0 \\
&- &14.5 &15.3 &20.5 &\textbf{35.7} \\
\bottomrule
\end{tabular}
}
\caption{Probability of language occurrence within the generated texts before and after intervention. Values in the "-" rows are the average values across six languages.}
\label{tab:intervention_language}
\end{table}

\begin{figure}[t]
  \centering
  \vspace{-2mm}
  \begin{minipage}[t]{1.0\linewidth}
        \centering
        \hspace{-6mm}
        \includegraphics[width=0.75\linewidth]{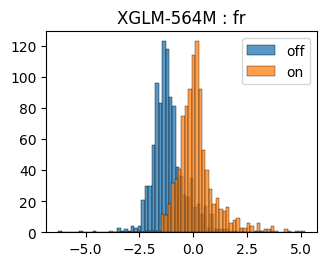}
  \end{minipage}
  \vspace{-3mm}
  \begin{minipage}[t]{1.0\linewidth}
        \centering  
        \hspace{-6mm}
        \includegraphics[width=0.75\linewidth]{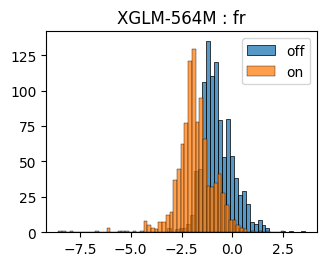}
  \end{minipage}
  \caption{(Top) Distributional difference of activation values of the top-1000 neurons between target (on) and non-target languages (off). (Bottom) Distributional difference of activation value of the bottom-1000 neurons.}
  \vspace{3mm}
  \label{fig:activation_histogram_main}
  \begin{minipage}[t]{1.0\linewidth}
        \centering
        \includegraphics[width=0.9\linewidth]{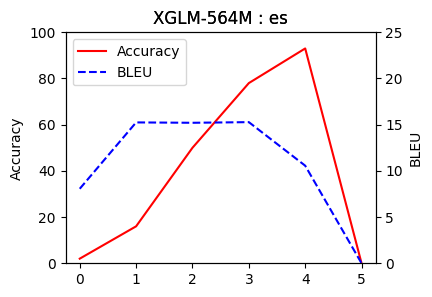}
        \vspace{-2mm}
        \caption{Ablation study of text generation by varying the number of neurons for intervention. x-axis: $\log_{10}(k)$
        }
        \label{fig:ablation_study_main}
  \end{minipage}
\end{figure}


\subsection{Controlling Language-specific Neurons}
\label{sec:result_control}


To show the effectiveness of the identified language-specific neurons, we investigated whether the models could control language in text generation by intervening with language-specific neurons.
We conducted the investigation using unconditional and conditional (i.e., machine translation) text-generation settings.

\subsubsection{Unconditional text generation}

In the experiments on unconditional text generation, we do not provide models with any input prompt, i.e., only a [BOS] token as a prompt.
Each model repeated text generation 100 times with random sampling decoding (temperature=0.8, top p=0.9) by changing the random seed from 1 to 100 each time the model started to generate text.
Figure \ref{main_text_sample_unconditional} illustrates model-generated text examples with the intervention setting in each class of language-specific neurons.
It was shown that by changing language-specific neurons for intervention, we can change the language of the output texts.

To quantitatively measure the probability of occurrence of a target language, we classified the generated texts into language categories using the language identification classifier FastText \citep{joulin2016bag,joulin2016fasttext}.
We classified each text into the target language if the classification score exceeded a threshold of 0.5 \citep{wenzek2019ccnet,touvron2023llama} and calculated the probability of the target language occurrence, i.e., the evaluation metric was accuracy.

Table \ref{tab:intervention_language} summarizes the results. This demonstrates that intervention in language-specific neurons increases the probability of the target language occurrence in unconditional text generation for each language.
In other words, the desired language could be generated by intentionally igniting target neurons.
It should be noted that the BLOOM models achieve a high probability of German and Japanese text occurrence by intervention, although the models do not explicitly include German and Japanese in their pre-training datasets, as described in Table \ref{tab:model_language}.
It is possible for a small number of these languages to be unintentionally mixed, leading to unintentional ability acquisition. For example, an English text and its translation to language may be present in a single document \citep{briakou2023searching}.

We conducted a study by intervening in only the top-1000 neurons, only the bottom-1000 neurons, and both groups of neurons. 
Interestingly, some languages did not respond to control by intervening only in the top-1000 or only the bottom-1000 neurons.
This suggests that it is possible to effectively control language by intervening in both groups of neurons.
In principle, the top-$k$ neurons are correlated with positive activation values. 
Conversely, the bottom-$k$ neurons were correlated with negative activation values. 
Figure \ref{fig:activation_histogram_main} validates this hypothesis.
These findings align with those of \citet{wang2022finding}, who suggested that neurons with both positive and negative correlations with labels are important for identifying target neurons.


\begin{table*}[t]\centering
\scalebox{0.86}{
{
\begin{tabular}{
    lr
    wr{7mm}wr{1mm}wr{4mm}
    wr{7mm}wr{1mm}wr{4mm}
    wr{7mm}wr{1mm}wr{4mm}
    wr{7mm}wr{1mm}wr{4mm}
    wr{7mm}wr{1mm}wr{4mm}
    wr{7mm}wr{1mm}wr{4mm}
}\toprule
& &\multicolumn{6}{c}{FLORES200} &\multicolumn{6}{c}{IWSLT2017} &\multicolumn{6}{c}{WMT} \\\cmidrule{3-20}
& &\multicolumn{3}{c}{Accuracy} &\multicolumn{3}{c}{BLEU} &\multicolumn{3}{c}{Accuracy} &\multicolumn{3}{c}{BLEU} &\multicolumn{3}{c}{Accuracy} &\multicolumn{3}{c}{BLEU} \\\midrule
XGLM-564M &de &0.0 &$\to$&\textbf{38.0} &0.0 &$\to$&0.0 &0.0 &$\to$&\textbf{15.0} &0.0 &$\to$&0.0 &0.0 &$\to$&\textbf{17.0} &0.0 &$\to$&0.0 \\
XGLM-564M &es &0.0 &$\to$&\textbf{3.0} &0.0 &$\to$&0.0 & &$\to$& & &$\to$& & &$\to$&\textbf{} & &$\to$& \\
XGLM-564M &ja &0.0 &$\to$&0.0 &0.0 &$\to$&0.0 &0.0 &$\to$&0.0 &0.0 &$\to$&0.0 & &$\to$&\textbf{} & &$\to$& \\
XGLM-564M &fr &0.0 &$\to$&0.0 &0.0 &$\to$&0.0 &0.0 &$\to$&\textbf{3.0} &0.0 &$\to$&0.0 &0.0 &$\to$&\textbf{1.0} &0.0 &$\to$&0.0 \\
XGLM-564M &zh &0.0 &$\to$&\textbf{1.0} &0.0 &$\to$&0.0 &0.0 &$\to$&2.0 &0.0 &$\to$&0.0 &0.0 &$\to$&\textbf{2.0} &0.0 &$\to$&0.0 \\
\midrule
BLOOM-1b7 &de &0.0 &$\to$&\textbf{56.0} &1.3 &$\to$&1.3 &0.0 &$\to$&\textbf{35.0} &1.0 &$\to$&\textbf{1.8} &0.0 &$\to$&\textbf{37.0} &\textbf{2.9} &$\to$&1.7 \\
BLOOM-1b7 &es &0.0 &$\to$&\textbf{2.0} &1.2 &$\to$&1.2 & &$\to$&\textbf{} & &$\to$&\textbf{} & &$\to$&\textbf{} & &$\to$&\textbf{} \\
BLOOM-1b7 &ja &0.0 &$\to$&\textbf{6.0} &\textbf{0.2} &$\to$&0.1 &0.0 &$\to$&\textbf{8.0} &0.1 &$\to$&\textbf{0.2} & &$\to$&\textbf{} & &$\to$&\textbf{} \\
BLOOM-1b7 &fr &0.0 &$\to$&\textbf{16.0} &1.7 &$\to$&\textbf{2.8} &0.0 &$\to$&\textbf{2.0} &1.0 &$\to$&\textbf{1.5} &0.0 &$\to$&\textbf{9.0} &1.7 &$\to$&\textbf{2.7} \\
BLOOM-1b7 &zh &0.0 &$\to$&\textbf{21.0} &\textbf{0.3} &$\to$&0.2 &0.0 &$\to$&\textbf{3.0} &0.2 &$\to$&\textbf{0.3} &0.0 &$\to$&\textbf{34.0} &0.5 &$\to$&\textbf{0.6} \\
\midrule
Llama2-7b &de &0.0 &$\to$&\textbf{66.0} &2.6 &$\to$&\textbf{17.7} &0.0 &$\to$&\textbf{48.0} &1.2 &$\to$&\textbf{12.5} &2.0 &$\to$&\textbf{53.0} &5.3 &$\to$&\textbf{15.2} \\
Llama2-7b &es &4.0 &$\to$&\textbf{77.0} &3.3 &$\to$&\textbf{16.6} & &$\to$&\textbf{} & &$\to$&\textbf{} & &$\to$&\textbf{} & &$\to$&\textbf{} \\
Llama2-7b &ja &0.0 &$\to$&\textbf{58.0} &0.3 &$\to$&\textbf{10.4} &1.0 &$\to$&\textbf{57.0} &0.2 &$\to$&\textbf{4.5} & &$\to$&\textbf{} & &$\to$&\textbf{} \\
Llama2-7b &fr &1.0 &$\to$&\textbf{58.0} &4.1 &$\to$&\textbf{21.5} &0.0 &$\to$&\textbf{32.0} &1.0 &$\to$&\textbf{11.1} &0.0 &$\to$&\textbf{36.0} &2.1 &$\to$&\textbf{13.2} \\
Llama2-7b &zh &1.0 &$\to$&\textbf{76.0} &1.0 &$\to$&\textbf{11.5} &3.0 &$\to$&\textbf{82.0} &0.6 &$\to$&\textbf{7.8} &12.0 &$\to$&\textbf{86.0} &2.4 &$\to$&\textbf{11.3} \\
\midrule
Llama2-13b &de &0.0 &$\to$&\textbf{22.0} &1.5 &$\to$&\textbf{8.8} &0.0 &$\to$&\textbf{37.0} &0.6 &$\to$&\textbf{10.0} &4.0 &$\to$&\textbf{32.0} &3.3 &$\to$&\textbf{9.7} \\
Llama2-13b &es &2.0 &$\to$&\textbf{14.0} &1.8 &$\to$&\textbf{4.3} & &$\to$&\textbf{} & &$\to$&\textbf{} & &$\to$&\textbf{} & &$\to$&\textbf{} \\
Llama2-13b &ja &7.0 &$\to$&\textbf{54.0} &2.4 &$\to$&\textbf{11.0} &4.0 &$\to$&\textbf{75.0} &0.7 &$\to$&\textbf{6.1} & &$\to$&\textbf{} & &$\to$&\textbf{} \\
Llama2-13b &fr &0.0 &$\to$&\textbf{23.0} &1.6 &$\to$&\textbf{10.5} &0.0 &$\to$&\textbf{9.0} &0.7 &$\to$&\textbf{4.7} &1.0 &$\to$&\textbf{15.0} &2.2 &$\to$&\textbf{6.6} \\
Llama2-13b &zh &20.0 &$\to$&\textbf{93.0} &4.4 &$\to$&\textbf{19.1} &40.0 &$\to$&\textbf{96.0} &5.8 &$\to$&\textbf{9.6} &57.0 &$\to$&\textbf{99.0} &13.5 &$\to$&\textbf{18.9} \\
\bottomrule
\end{tabular}
}
}
\captionsetup{width=1.0\linewidth}
\caption{Results of conditional text generation. Values on the left side of the arrows ($\to$) were measured without intervention on the language-specific neurons; values on the right side were measured during intervention on the neurons. FLORES200 includes translation tasks of English to the other five languages, while IWSLT2017 does not include tasks of English to Spanish, and WMT does not include tasks of English to Spanish or Japanese.}
\label{tab:conditional_generation_main}
\end{table*}

\begin{table*}[t]\centering

\begin{tabular}{lrrrrrrrrrrrrr}\toprule
&\multicolumn{6}{c}{``Translate a sentence} &\multicolumn{6}{c}{``Translate an English sentence} \\
&\multicolumn{6}{c}{from English to a target language.''} &\multicolumn{6}{c}{into a target language.''} \\
\cmidrule{2-13}
&\multicolumn{3}{c}{Accuracy} &\multicolumn{3}{c}{BLEU} &\multicolumn{3}{c}{Accuracy} &\multicolumn{3}{c}{BLEU} \\
\midrule
de &0.0 &$\to$ &\textbf{62.0} &2.8 &$\to$ &\textbf{16.5} &0.0 &$\to$ &\textbf{66.0} &2.6 &$\to$ &\textbf{17.7} \\
es &5.0 &$\to$ &\textbf{78.0} &4.0 &$\to$ &\textbf{16.5} &4.0 &$\to$ &\textbf{77.0} &3.3 &$\to$ &\textbf{16.6} \\
ja &0.0 &$\to$ &\textbf{55.0} &0.3 &$\to$ &\textbf{9.2} &0.0 &$\to$ &\textbf{58.0} &0.3 &$\to$ &\textbf{10.4} \\
fr &0.0 &$\to$ &\textbf{58.0} &3.4 &$\to$ &\textbf{21.3} &1.0 &$\to$ &\textbf{58.0} &4.1 &$\to$ &\textbf{21.5} \\
zh &1.0 &$\to$ &\textbf{79.0} &1.2 &$\to$ &\textbf{12.7} &1.0 &$\to$ &\textbf{76.0} &1.0 &$\to$ &\textbf{11.5} \\
\midrule
&\multicolumn{6}{c}{``Translate an English sentence} &\multicolumn{6}{c}{``Translate an English sentence} \\
&\multicolumn{6}{c}{into German.''} &\multicolumn{6}{c}{into Japanese.''} \\
\cmidrule{2-13}
&\multicolumn{3}{c}{Accuracy} &\multicolumn{3}{c}{BLEU} &\multicolumn{3}{c}{Accuracy} &\multicolumn{3}{c}{BLEU} \\
\midrule
de &96.0 &$\to$ &\textbf{99.0} &\textbf{32.8} &$\to$ &24.4 &0.0 &$\to$ &\textbf{2.0} &0.3 &$\to$ &\textbf{1.2} \\
es &0.0 &$\to$ &\textbf{1.0} &2.0 &$\to$ &\textbf{2.6} &0.0 &$\to$ &\textbf{2.0} &0.1 &$\to$ &\textbf{0.4} \\
ja &\textbf{0.0} &$\to$ &\textbf{0.0} &0.3 &$\to$ &\textbf{0.4} &\textbf{100.0} &$\to$ &99.0 &\textbf{24.3} &$\to$ &19.7 \\
fr &0.0 &$\to$ &\textbf{3.0} &2.6 &$\to$ &\textbf{3.1} &0.0 &$\to$ &\textbf{3.0} &0.2 &$\to$ &\textbf{1.0} \\
zh &0.0 &$\to$ &\textbf{2.0} &\textbf{0.8} &$\to$ &0.4 &0.0 &$\to$ &\textbf{96.0} &1.3 &$\to$ &\textbf{14.9} \\
\bottomrule
\end{tabular}

\captionsetup{width=1.0\linewidth}
\caption{Results of conditional text generation with different prompt settings for Llama2-7b.}
\label{tab:conditional_generation_ablation}
\end{table*}

We conducted an ablation study by changing the number of intervening neurons and analyzed its effect on the probability of target language occurrence.
Additionally, we verified the quality of each model-generated text using the BLEU-4 score (BLEU).
We evaluated BLEU only for texts identified by the language identifier as belonging to the target language.
Specifically, for each model-generated text identified as the target language, we set the text as a hypothesis and all positive texts as references, and measured the BLEU score.
We  averaged the BLEU scores across all model-generated texts that were identified as the target language.
Figure \ref{fig:ablation_study_main} shows the results of the intervention in Spanish neurons for XGLM-564M.
This shows that increasing the number of intervening neurons up to 1000-10000 (3-4 on the logarithm of 10 in this figure) generally increases the probability of target language occurrence, but increasing beyond that degrades text quality. 
Eventually, the sentence collapses and both language identification and quality significantly decrease. This tendency exists regardless of language or model variation.
See Appendix \ref{appx:ablation_study} for the complete results.

\subsubsection{Conditional text generation}
In experiments of conditional text generation, models were given machine translation tasks and required to solve them in a zero-shot prompting setting, but with an unusual prompt format: ``Translate an English sentence into a target language. English: \{source text\} Target Language:''. 
In other words, it is a prompt for a translation task that does not concretely specify the target language.
The aim of the prompt was to accurately check whether the manipulation of language-specific neurons could lead to the output of the target language.
Using this prompt as an input, the models started to generate text using a greedy decoding strategy.
For this experiment, we randomly selected 100 machine translation samples from FLORES200, IWSLT2017 \citep{cettolo2017overview}, and WMT \citep{bojar-EtAl:2014:W14-33,bojar-EtAl:2016:WMT1,bojar-EtAl:2018:WMT1}\footnote{We used En $\to$ Fr tasks from WMT14, En $\to$ De tasks from WMT16, and En $\to$ Zh tasks from WMT18.}, respectively.
Two evaluation metrics were used to measure translation quality: Accuracy of measuring the probability that the target language text is generated, and BLEU.
In the unconditional text-generation setting, we measured the quality of only the generated texts in the target language. However, in the conditional text-generation setting, we calculated the BLEU score between each generated text and the corresponding ground-truth text by following the standard method of BLEU evaluation in machine translation tasks \citep{papineni2002bleu}.

Table \ref{tab:conditional_generation_main} summarizes the experimental results of the conditional text generation.
There were two main findings from these results. First, interventions in language-specific neurons tend to increase the probability of producing the target language (accuracy). Second, the translation quality (BLEU) of Llama2 models increased drastically along with accuracy. In contrast, the translation quality of XGLM and BLOOM did not significantly improve compared to the accuracy improvement.
We investigated the reason for this by qualitatively analyzing the generated texts. XGLM and BLOOM were forced to output the target languages to some degree via intervention, but the output texts were not related to  translation.
For instance, when we intervened in German neurons, XGLM tended to output a word ``Deutsch''.
BLOOM tended to generate text unrelated to the translation or simply repeated the source text in English.
Conversely, Llama2 tended to output translated text in the correct target language, resulting in improved accuracy and BLEU scores.
This experiment showed that intervention in language-specific neurons can guide some models in the right direction, even when the models promptly receive an ambiguous translation task.
Figure \ref{main_text_sample_conditional} shows examples of model-generated text for Llama2-7b model. See Section \ref{appx:generated_text_samples_conditional} for additional examples.

We conducted several baseline experiments by changing the prompts to validate the robustness of the model outputs against prompts for the machine translation settings.
Specifically, we tried the following four prompts: 1.``Translate a sentence from English into a target language. English: \{source text\} Target Language:'', 2.``Translate an English sentence into a target language. English: \{source text\} Target Language:'', 3.``Translate an English sentence into German. English: \{source text\} German:'', 4.``Translate an English sentence into Japanese. English: \{source text\} Japanese:''.

The first and second prompts are ambiguous because they do not explicitly specify the target language. The second prompt is the same as that in Table 5. Regarding the third and fourth prompts, we explicitly describe the target languages in the prompts: German from Western languages and Japanese from Eastern languages. Here, we focus on the Llama-2-7b model because it has significantly improved both accuracy and BLEU scores, as described in Table 5. Similar to the experiment shown in Table 5, we conducted experiments in which the model was asked to solve a translation task under a specified prompt, while intervening in language-specific neurons. 

The experimental results are presented in Table \ref{tab:conditional_generation_ablation}. 
The first and second prompts significantly increased the probability of target language occurrence and BLEU scores with intervention in language-specific neurons for all languages. In contrast, the third and fourth prompts caused few changes when we intervened in language-specific neurons for most languages. One possible reason is that explicitly specifying the target language in a prompt automatically fires specific neurons in that language, which may offset the effects of other language-specific neurons. 

The only exception was the intervention in Chinese neurons under the fourth prompt ``Translate an English sentence into Japanese.'', which increases the probability of Chinese text generation and BLEU scores. 
One possible reason is that some Japanese and Chinese neurons have similar firing patterns within the Llama-2-7b model.
As shown in Table 7 in the Appendix, the Llama-2-7b model had a higher language-specific neuron overlap between Japanese and Chinese than the other pairs.
As these two languages share many characters with similar surface morphologies, the high similarity between the two languages may have contributed to these results.
However, it should be noted that this is not universally true across models; in some cases, the overlap of neurons in this language pair is not always high in models other than Llama2, as described in Table 7 in the Appendix.

\section{Conclusion}

This study provides new insights into the activity of language-specific neurons in decoder-based multilingual pre-trained language models: the existence of neurons that fire uniquely for each language.
The experimental results demonstrate that language-specific neurons mainly exist in the first and last few layers, regardless of the language, model size, and model variants. 
We further analyzed the effectiveness of the identified neurons by intervening in the neurons, that is, by replacing the output values with fixed activation values at inference with both unconditional and conditional settings. 
Using this approach, we can change the probability of the target language occurrence.

We hope that this study facilitates a deeper understanding of decoder-based PLMs and provides new insights for future research on multilingual decoder-based PLMs.
Future research should include proposing language-specific model-compression methods.
Future research also includes proposing new fine-tuning methods for downstream tasks to facilitate generalization to languages that are not included in the training dataset. For example, only fine-tuning the middle-layer parameters in decoder-based PLMs.


\section{Limitation}
\label{sec:limitation}

This study only analyzes open models whose parameters were publicly available.
It is not possible to analyze closed models with parameters that are not publicly available, such as ChatGPT or GPT-4 \citep{openai2023gpt}.
Although we focused our analysis on six languages, other languages need to be examined in future studies.
Analysis of encoder-decoder-based PLMs, such as mT5 \citep{xue2020mt5}, remains important but is beyond the scope of this study due to the fundamental differences in model architecture from decoder-only PLMs.

\section*{Acknowledgements}
We thank three anonymous reviewers for their helpful comments and feedback.
This study was partially supported by PRESTO, JST Grant Number JPMJPR21C8, Japan.

\bibliography{custom}


\appendix

\section{Text Examples from Language-Specific Text Corpus}

\begin{figure*}[t]
\begin{center}
\fbox{
\includegraphics[width=0.95\linewidth]{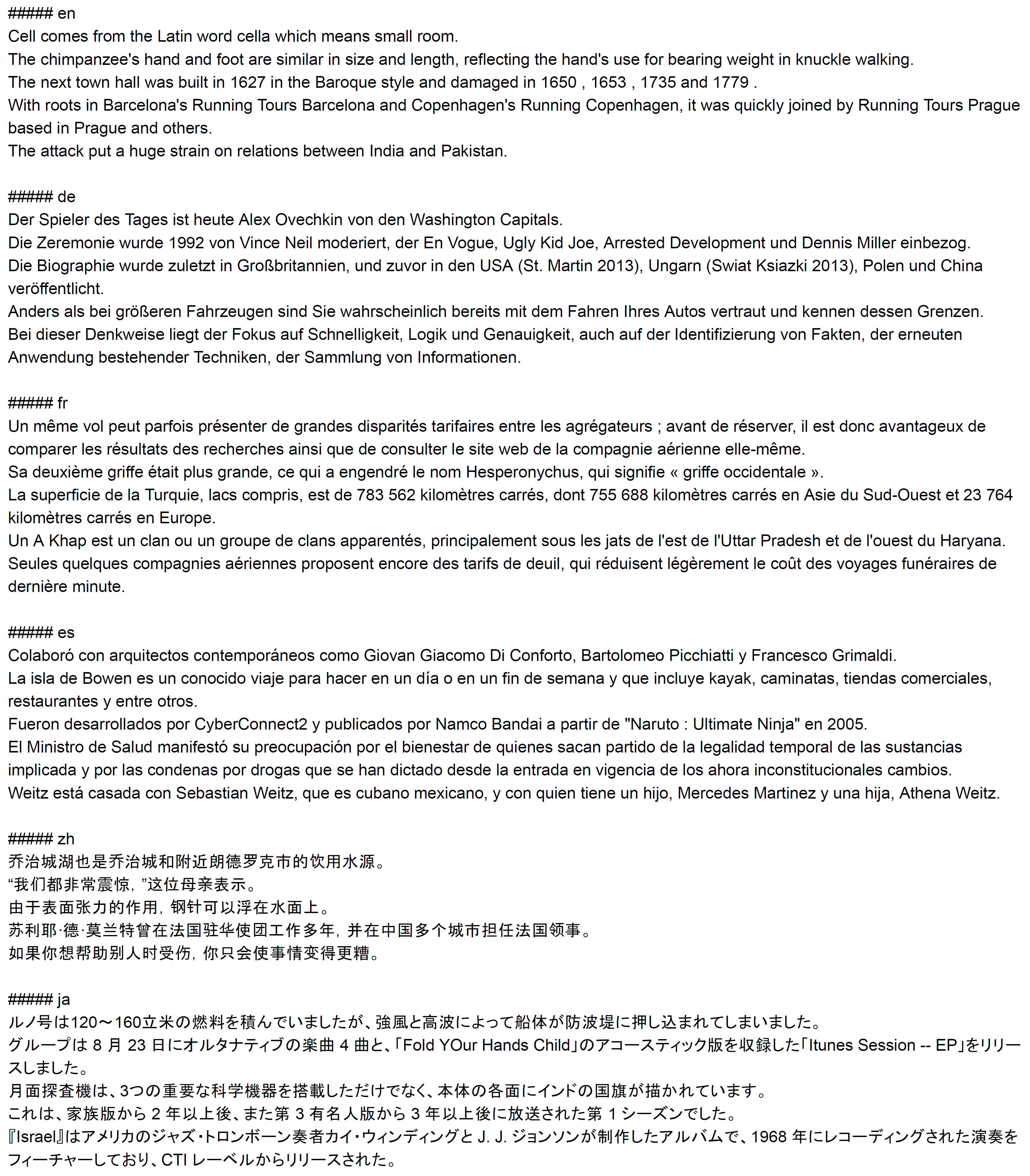}
}
\caption{Text examples from language-specific text corpus, which is a mixture of FLORES-200 and PAWS-X.}
\label{text_samples_gt}
\end{center}
\end{figure*}

Figure \ref{text_samples_gt} lists text examples from the language-specific corpus, which is a combination of FLORES-200 and PAWS-X.

\section{Identification of Neurons}

\subsection{Distribution of Language-Specific Neurons Across Layers}
\label{appx:histogram}

\subsubsection{Histogram}

\begin{figure*}[t]
\begin{center}
\begin{flushleft}
\includegraphics[width=0.85\linewidth]{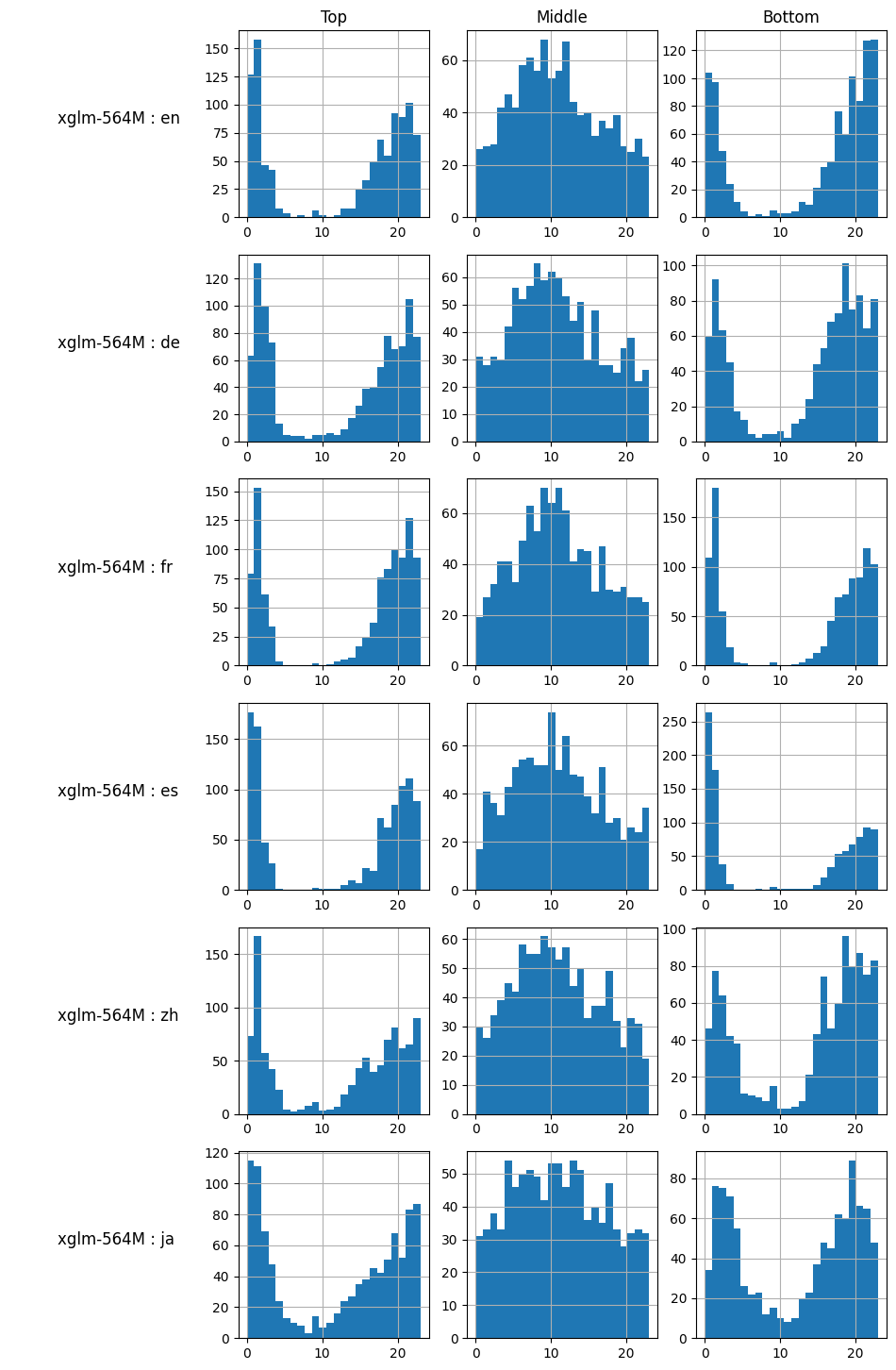}
\end{flushleft}
\caption{Histogram of language neurons across layers in xglm-564M.}
\label{appendix_histogram_xglm564m}
\end{center}
\end{figure*}

\begin{figure*}[t]
\begin{center}
\begin{flushleft}
\includegraphics[width=0.85\linewidth]{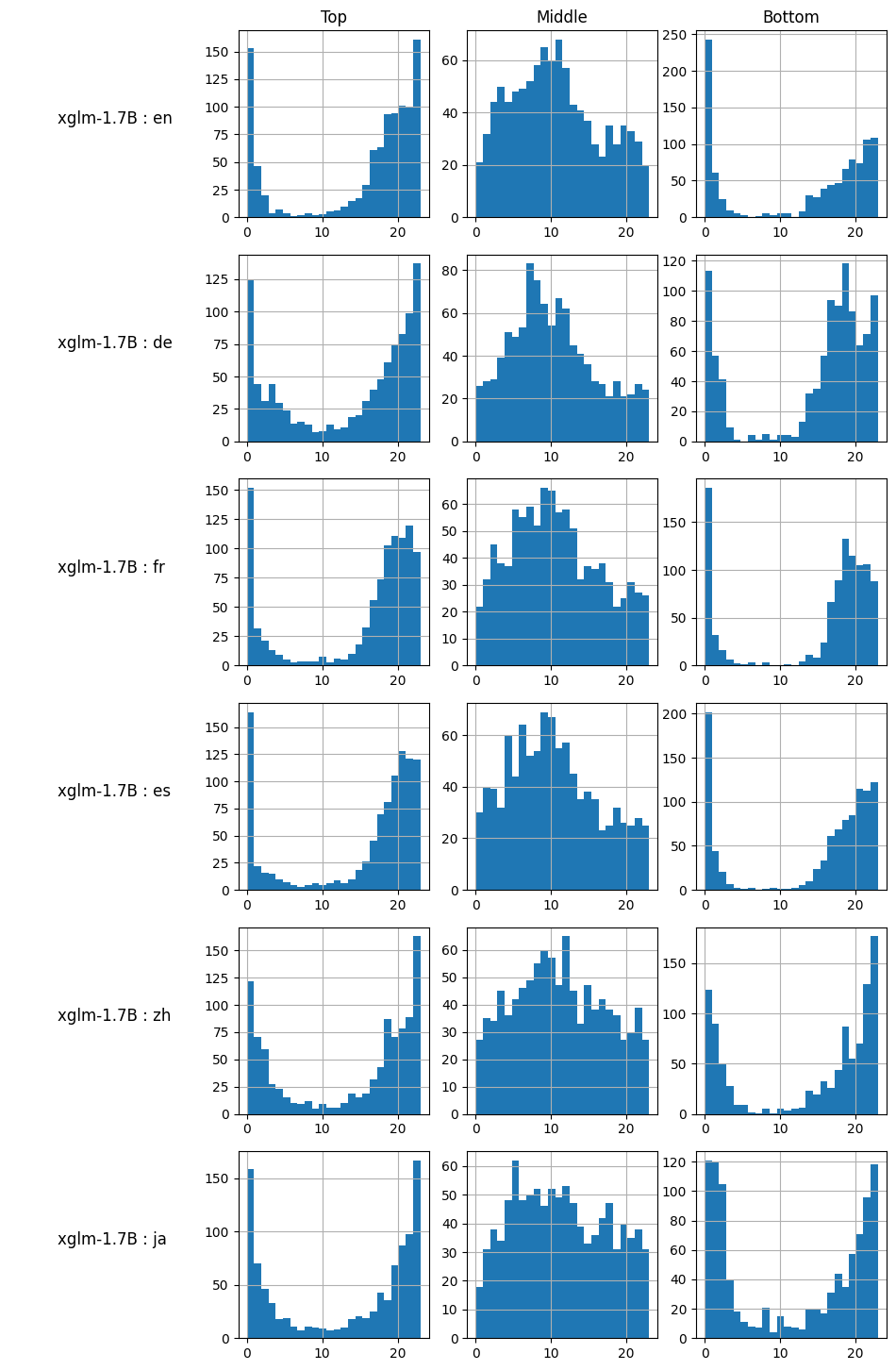}
\end{flushleft}
\caption{Histogram of language neurons across layers in xglm-1.7B.}
\label{appendix_histogram_xglm1B7}
\end{center}
\end{figure*}

\begin{figure*}[t]
\begin{center}
\begin{flushleft}
\includegraphics[width=0.85\linewidth]{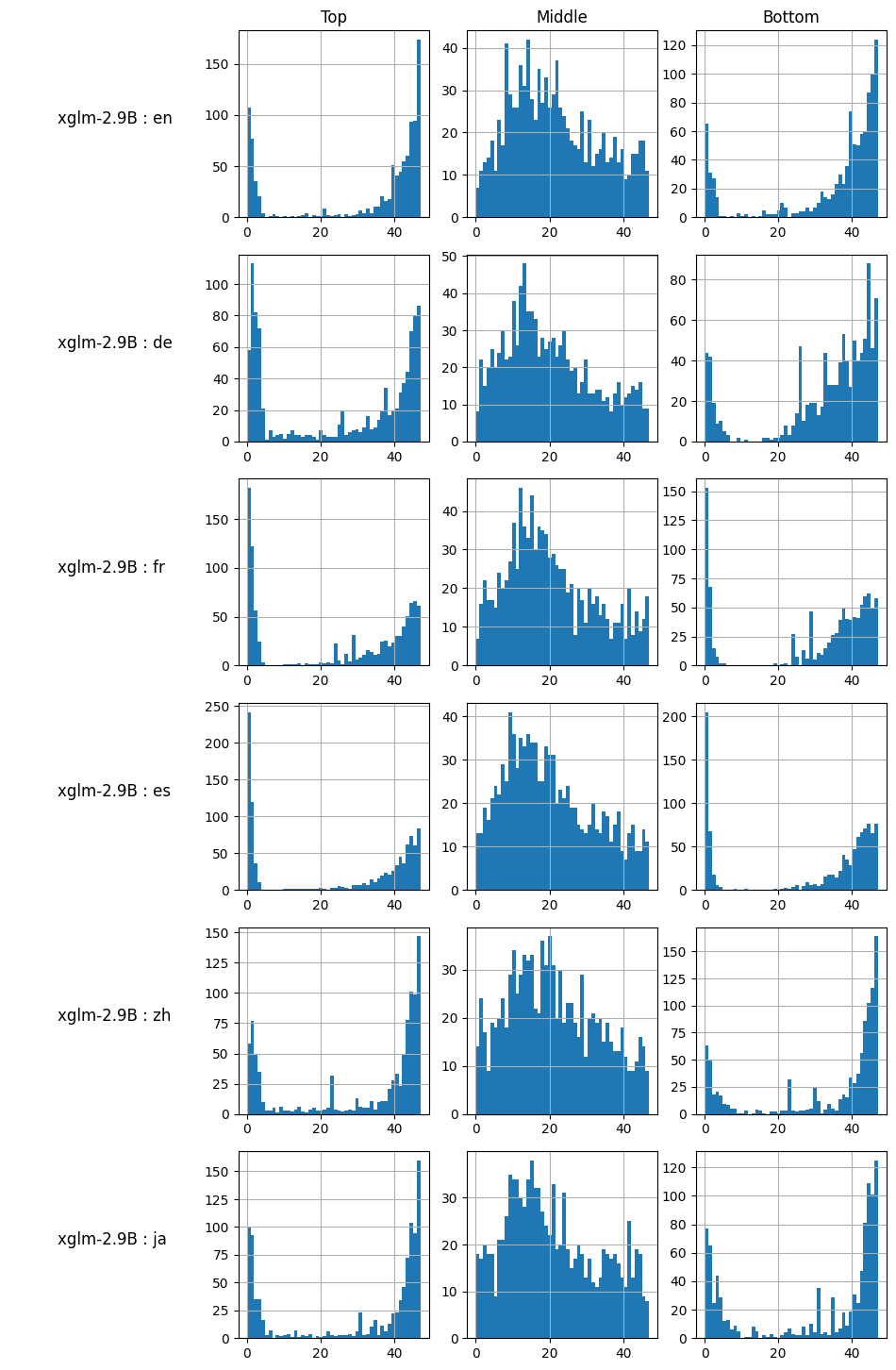}
\end{flushleft}
\caption{Histogram of language neurons across layers in xglm-2.9B.}
\label{appendix_histogram_xglm2B9}
\end{center}
\end{figure*}

\begin{figure*}[t]
\begin{center}
\begin{flushleft}
\includegraphics[width=0.85\linewidth]{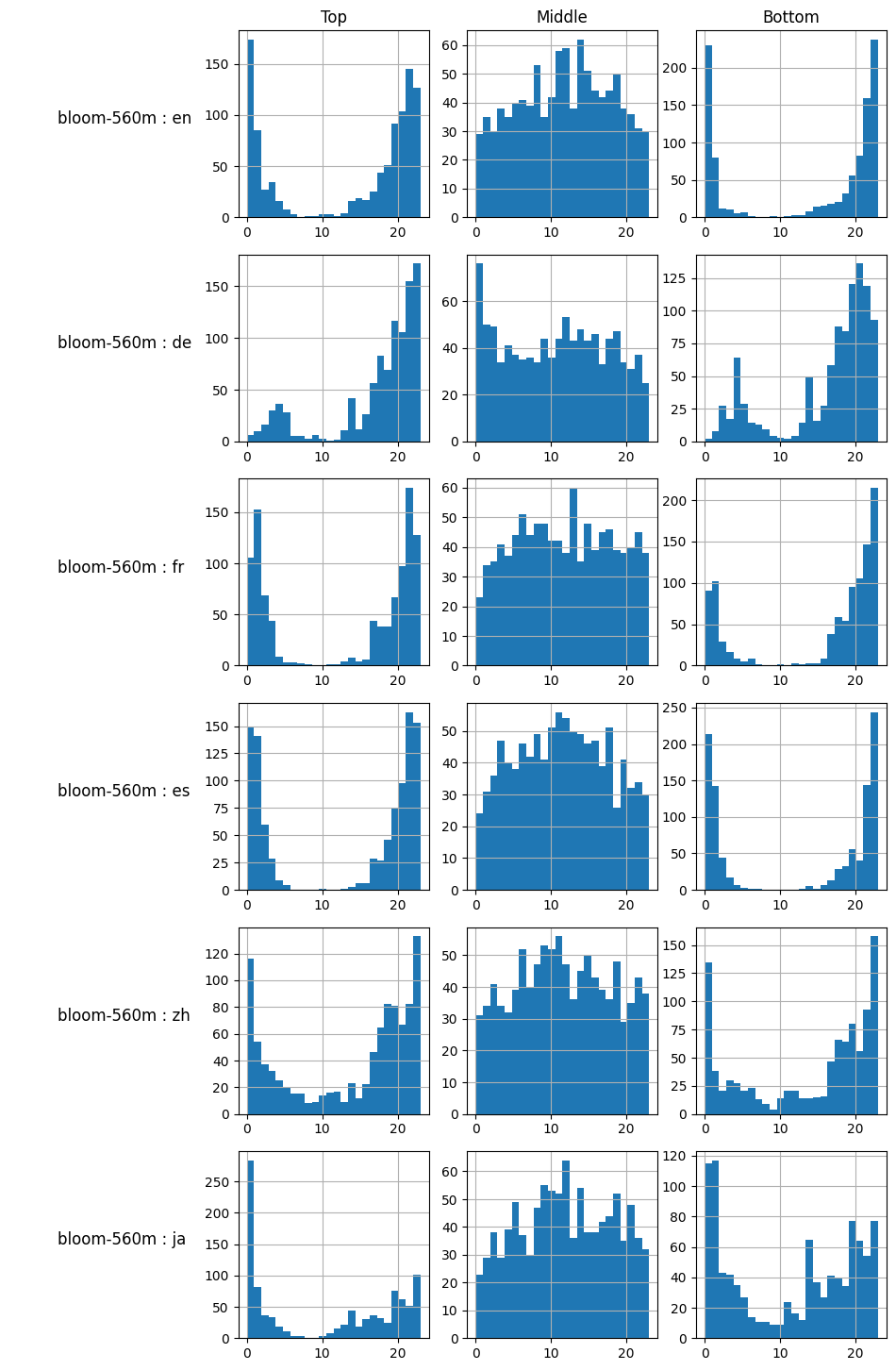}
\end{flushleft}
\caption{Histogram of language neurons across layers in bloom-560m.}
\label{appendix_histogram_bloom560m}
\end{center}
\end{figure*}

\begin{figure*}[t]
\begin{center}
\begin{flushleft}
\includegraphics[width=0.85\linewidth]{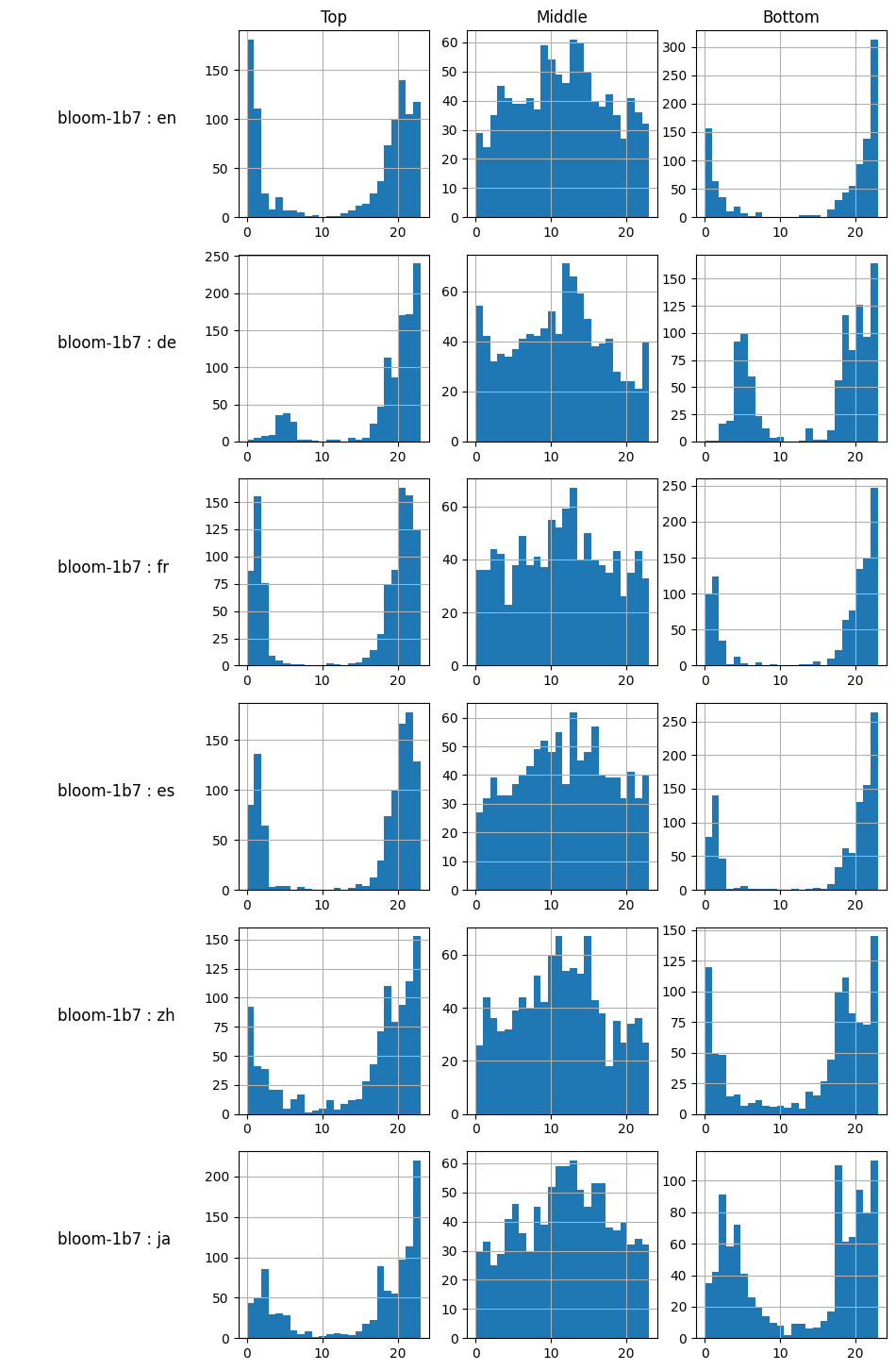}
\end{flushleft}
\caption{Histogram of language neurons across layers in bloom-1b7.}
\label{appendix_histogram_bloom1b7}
\end{center}
\end{figure*}

\begin{figure*}[t]
\begin{center}
\begin{flushleft}
\includegraphics[width=0.85\linewidth]{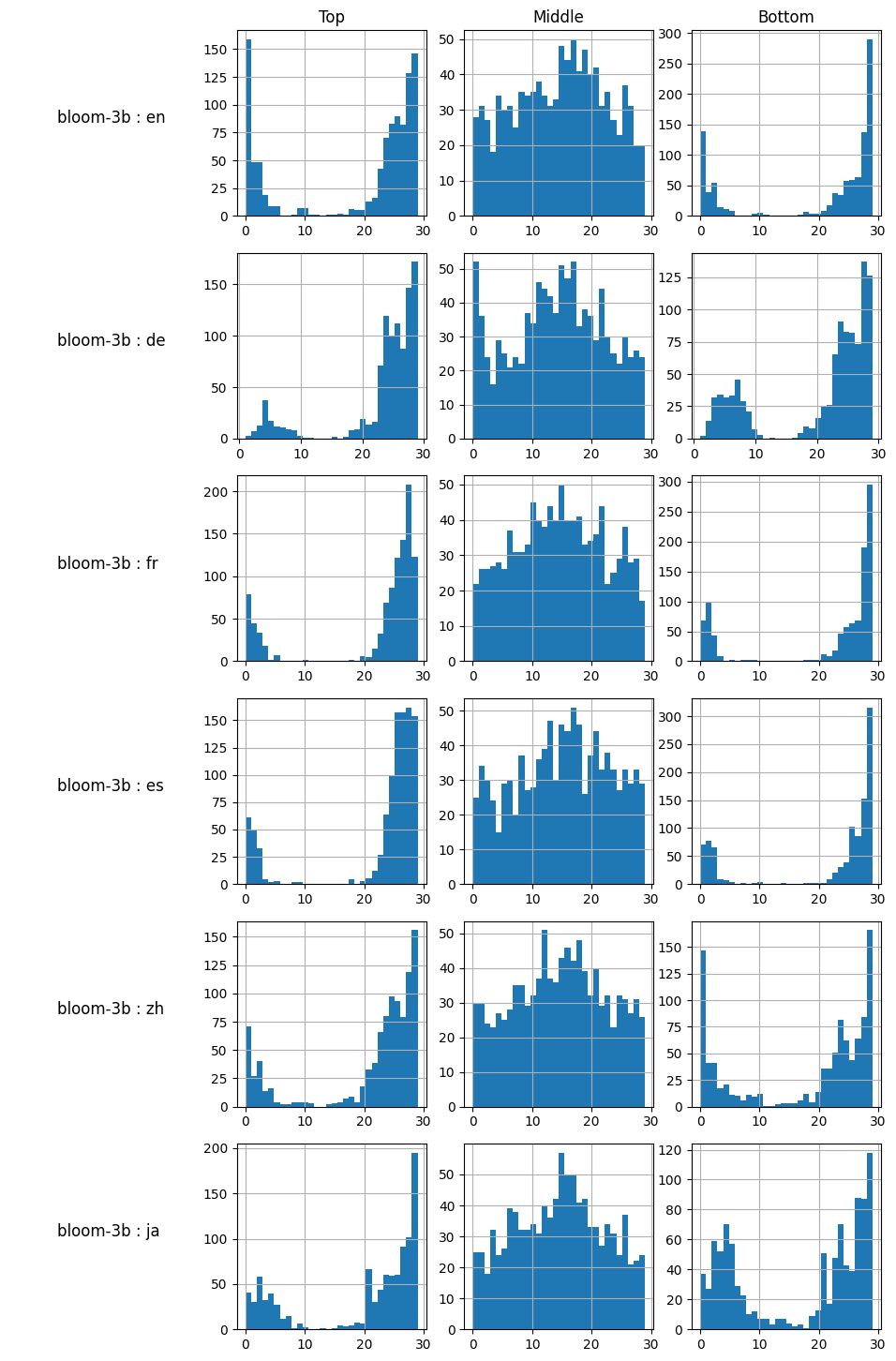}
\end{flushleft}
\caption{Histogram of language neurons across layers in bloom-3b.}
\label{appendix_histogram_bloom3b}
\end{center}
\end{figure*}

\begin{figure*}[t]
\begin{center}
\begin{flushleft}
\includegraphics[width=0.85\linewidth]{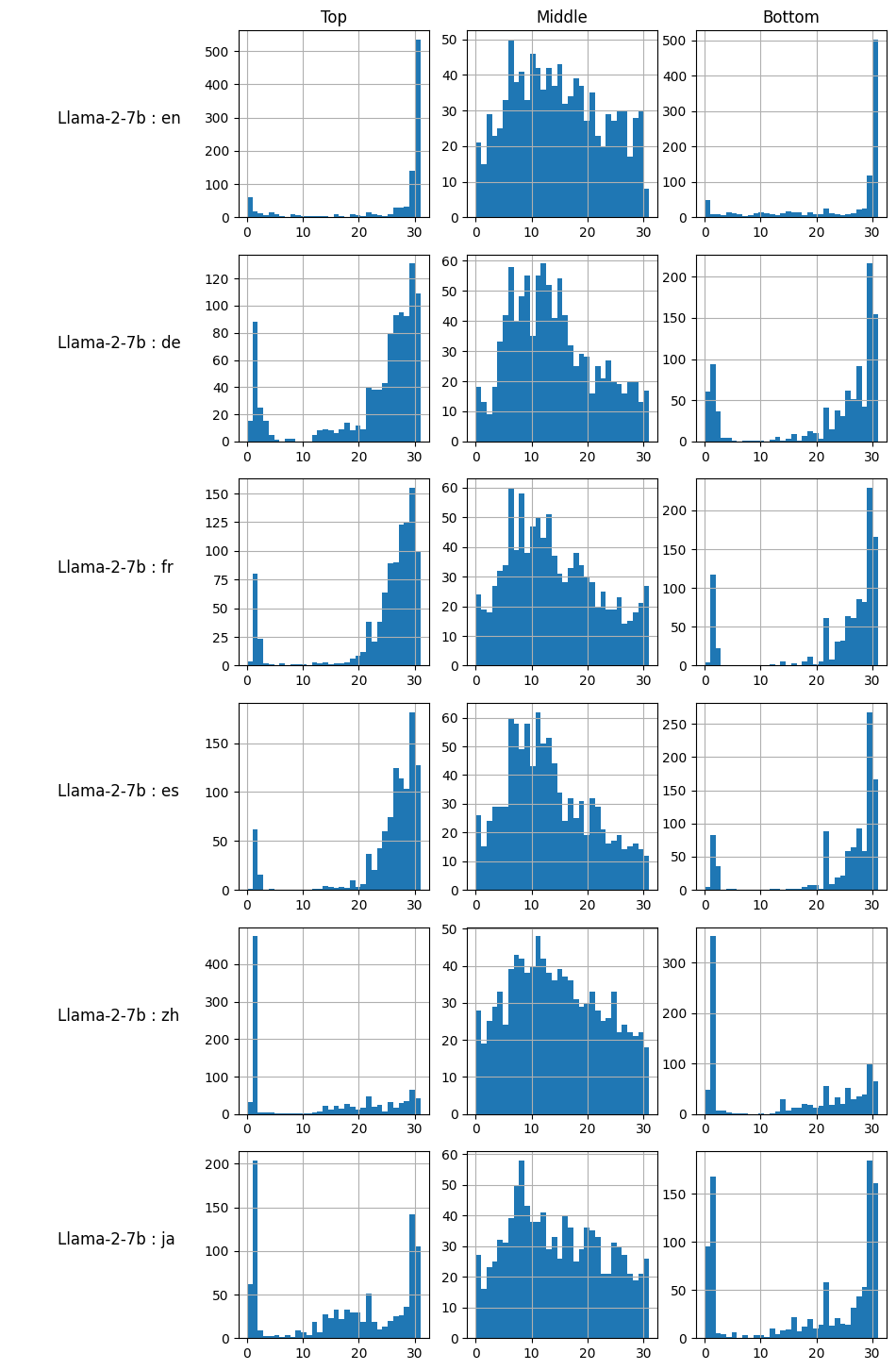}
\end{flushleft}
\caption{Histogram of language neurons across layers in Llama2-7b.}
\label{appendix_histogram_Llama2_7b}
\end{center}
\end{figure*}

\begin{figure*}[t]
\begin{center}
\begin{flushleft}
\includegraphics[width=0.85\linewidth]{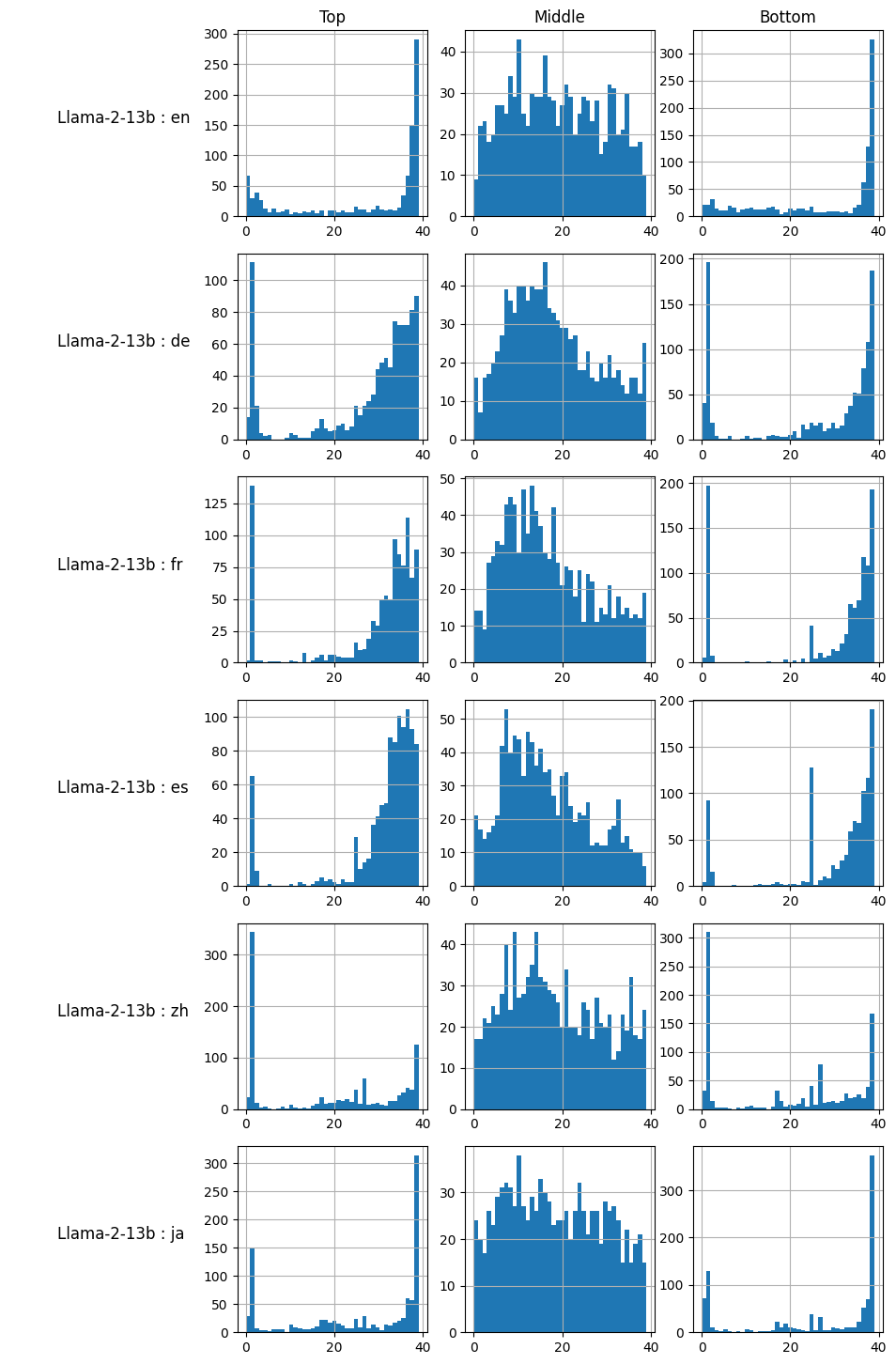}
\end{flushleft}
\caption{Histogram of language neurons across layers in Llama-2-13b.}
\label{appendix_histogram_Llama2_13b}
\end{center}
\end{figure*}

Figure \ref{appendix_histogram_xglm564m}, \ref{appendix_histogram_xglm1B7}, \ref{appendix_histogram_xglm2B9}, \ref{appendix_histogram_bloom560m}, \ref{appendix_histogram_bloom1b7}, \ref{appendix_histogram_bloom3b}, \ref{appendix_histogram_Llama2_7b}, and \ref{appendix_histogram_Llama2_13b}
describes the distribution of language-specific neurons across layers in each model using a histogram.

\subsubsection{Estimation of Beta Distribution Parameters}

To quantitatively analyze the shape of the distribution, we regarded it as a sampling subset from a beta distribution and estimated its parameters of that beta distribution. The beta distribution has two parameters, $\alpha$ and $\beta$. If $\alpha < 1$ and $\beta < 1$, then the distribution becomes convex downward. If $\alpha > 1$ and $\beta > 1$, then the distribution becomes convex upward. Table \ref{tab:beta_distribution} lists the parameter-estimation results. It is clearly shown that the distribution of the top-1000 and bottom-1000 neurons generally has the parameters $\alpha < 1$ and $\beta < 1$, supporting the claim that language-specific neurons exist in the first and last few layers. In contrast, the estimated parameters of the middle N neurons are $\alpha > 1$ and $\beta > 1$, indicating that language-independent neurons exist in the middle layers.

\subsection{Overlapping language-specific neurons between languages.}
\label{appx:overlap}

Table \ref{tab:cross_check_1} -  
\ref{tab:cross_check_6} describe cross-table check results to count the number of overlapping language-specific neurons between languages.

\subsection{Activation Values of Top and Bottom-1000 Neurons}

\begin{figure*}[t]
\begin{center}
\includegraphics[width=0.85\linewidth]{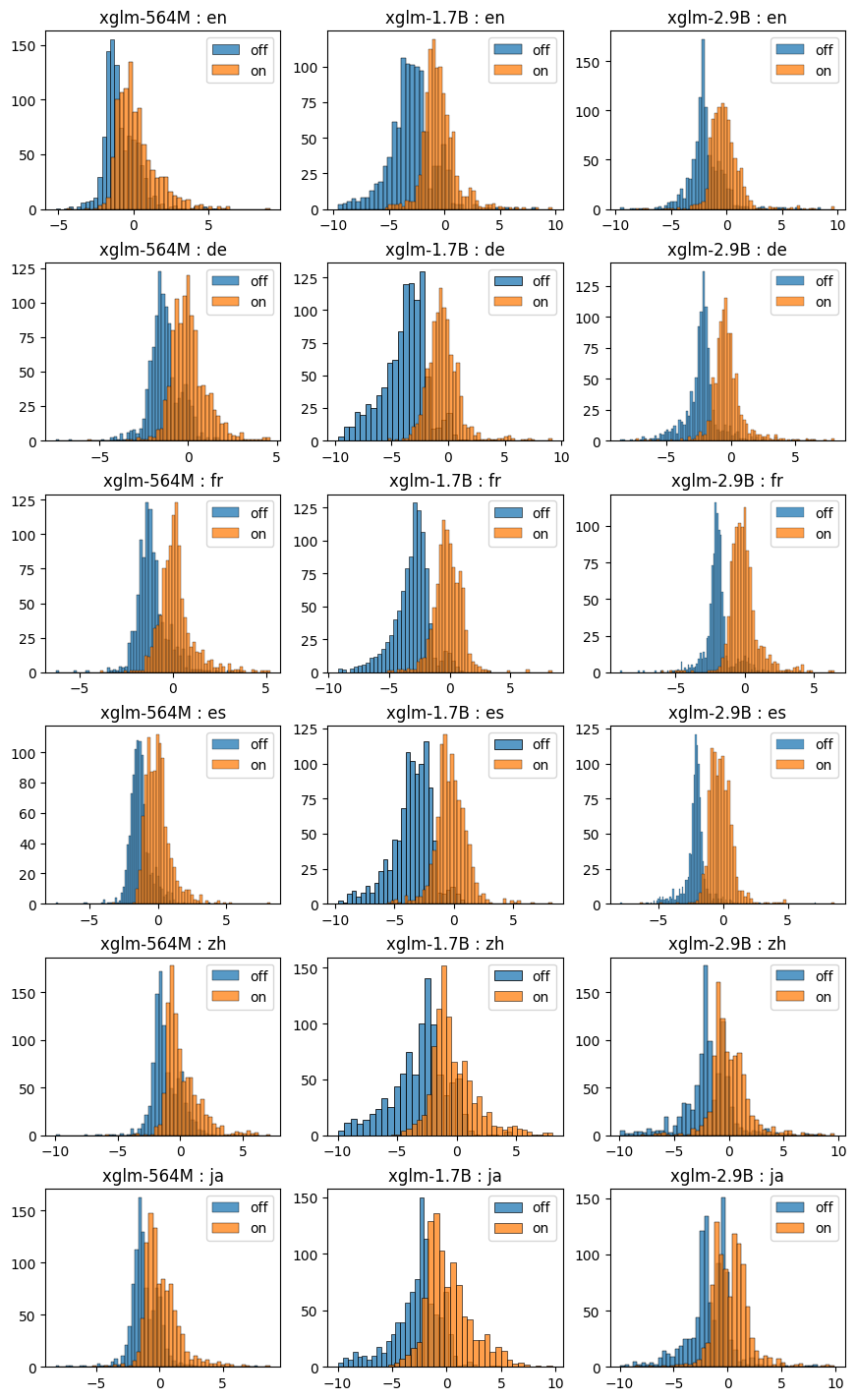}
\centering
\caption{Activation value difference of top-1000 neurons between target language(on) and non-target languages(off). x-axis: activation value of neurons. y-axis: frequency.}
\label{fig:activation_histogram_appendix_top_1}
\end{center}
\end{figure*}

\begin{figure*}[t]
\begin{center}
\includegraphics[width=0.85\linewidth]{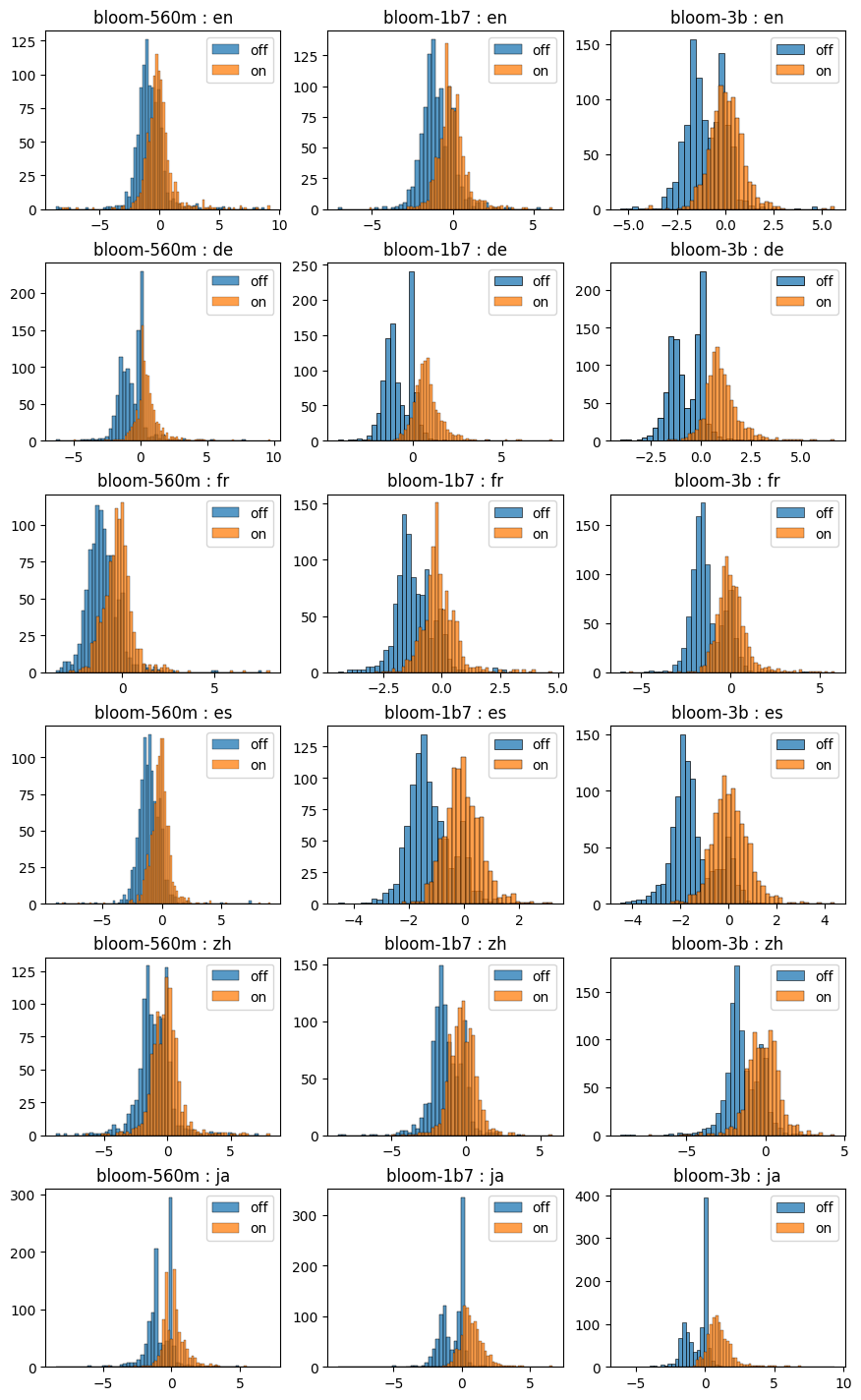}
\caption{Activation value difference of top-1000 neurons between target language(on) and non-target languages(off). x-axis: activation value of neurons. y-axis: frequency.}
\label{fig:activation_histogram_appendix_top_2}
\end{center}
\end{figure*}

\begin{figure*}[t]
\begin{center}
\includegraphics[width=0.60\linewidth]{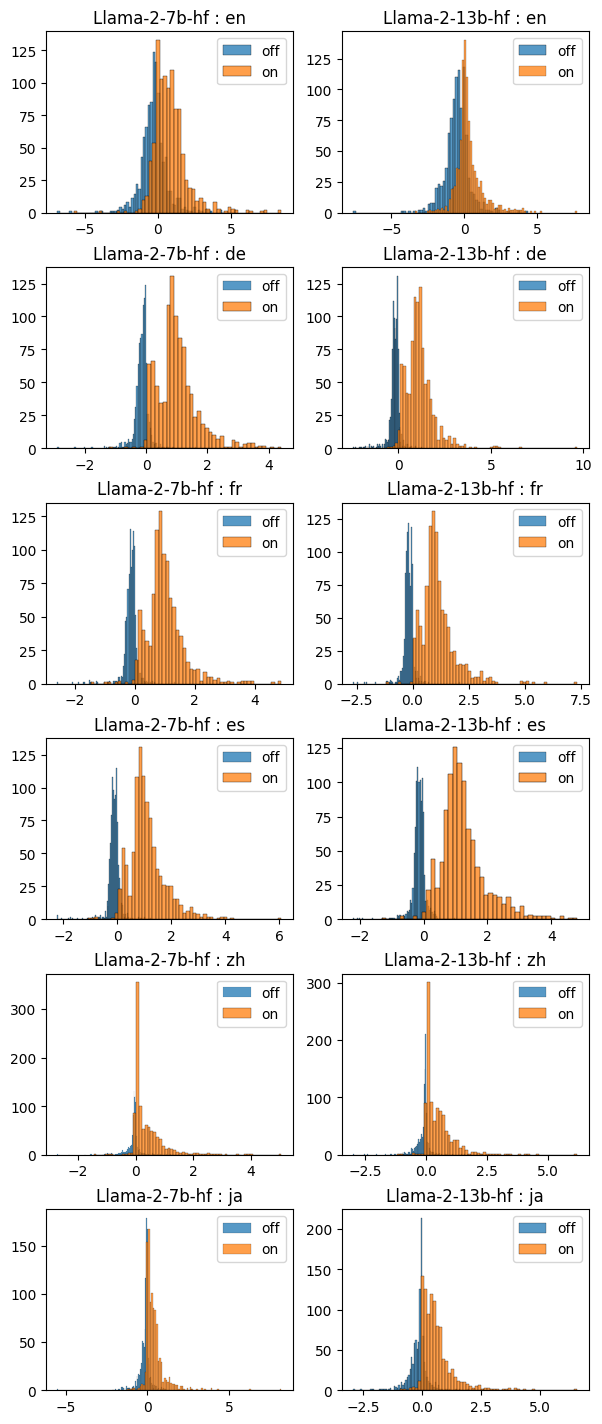}
\caption{Activation value difference of top-1000 neurons between target language(on) and non-target languages(off). x-axis: activation value of neurons. y-axis: frequency.}
\label{fig:activation_histogram_appendix_top_3}
\end{center}
\end{figure*}

\begin{figure*}[t]
\begin{center}
\includegraphics[width=0.85\linewidth]{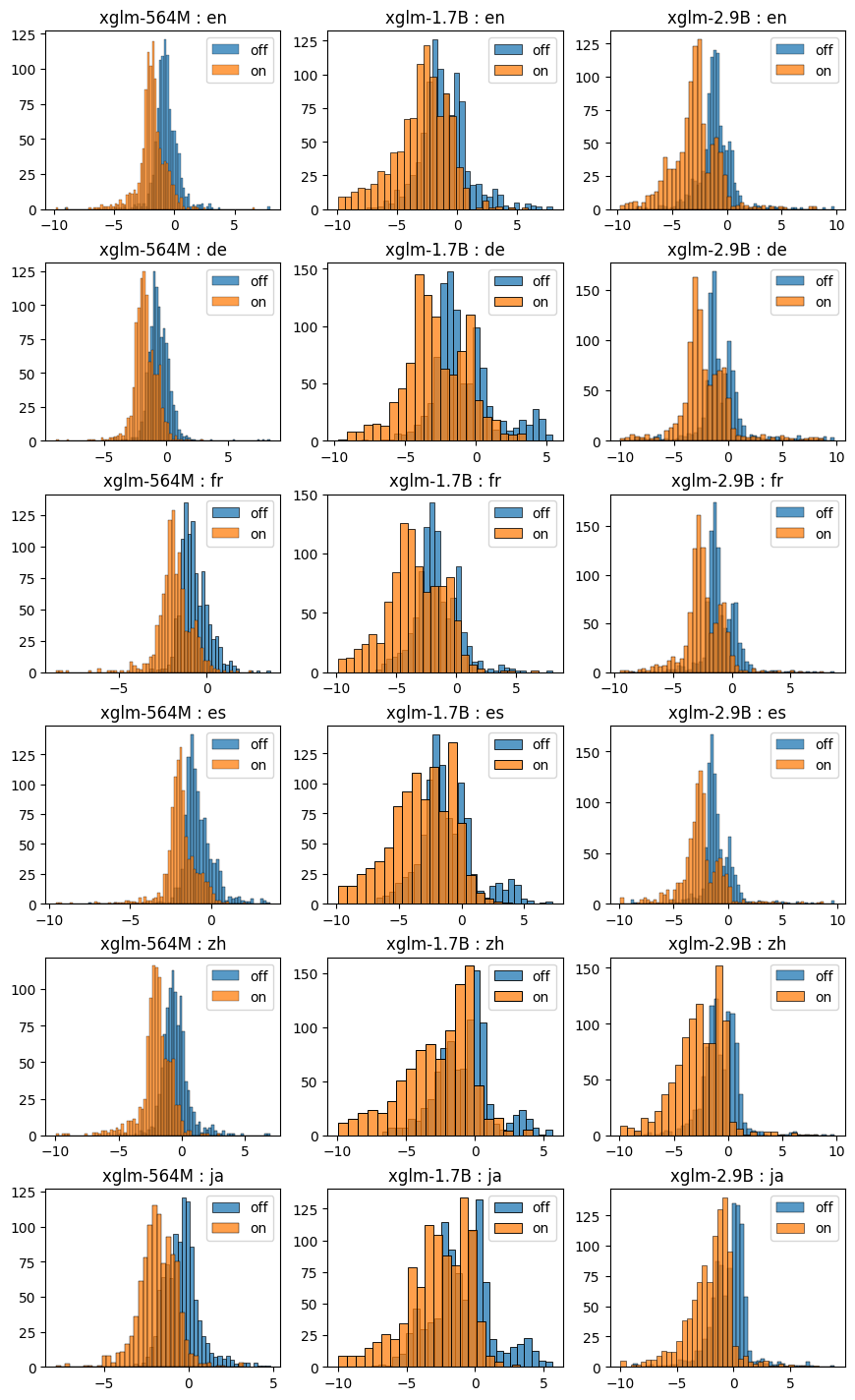}
\caption{Activation value difference of bottom-1000 neurons between target language(on) and non-target languages(off). x-axis: activation value of neurons. y-axis: frequency.}
\label{fig:activation_histogram_appendix_bottom_1}
\end{center}
\end{figure*}

\begin{figure*}[t]
\begin{center}
\includegraphics[width=0.85\linewidth]{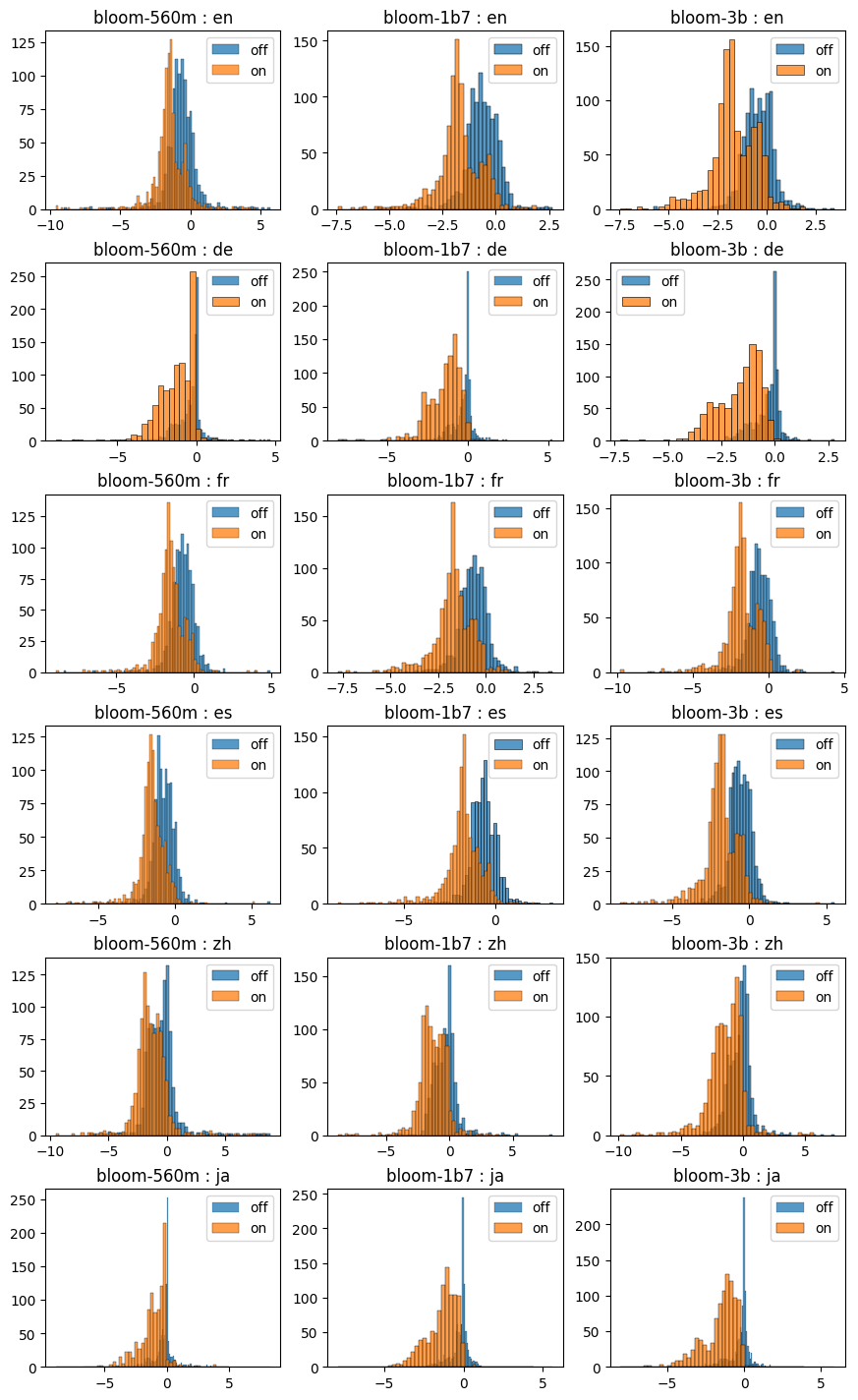}
\caption{Activation value difference of bottom-1000 neurons between target language(on) and non-target languages(off). x-axis: activation value of neurons. y-axis: frequency.}
\label{fig:activation_histogram_appendix_bottom_2}
\end{center}
\end{figure*}

\begin{figure*}[t]
\begin{center}
\includegraphics[width=0.60\linewidth]{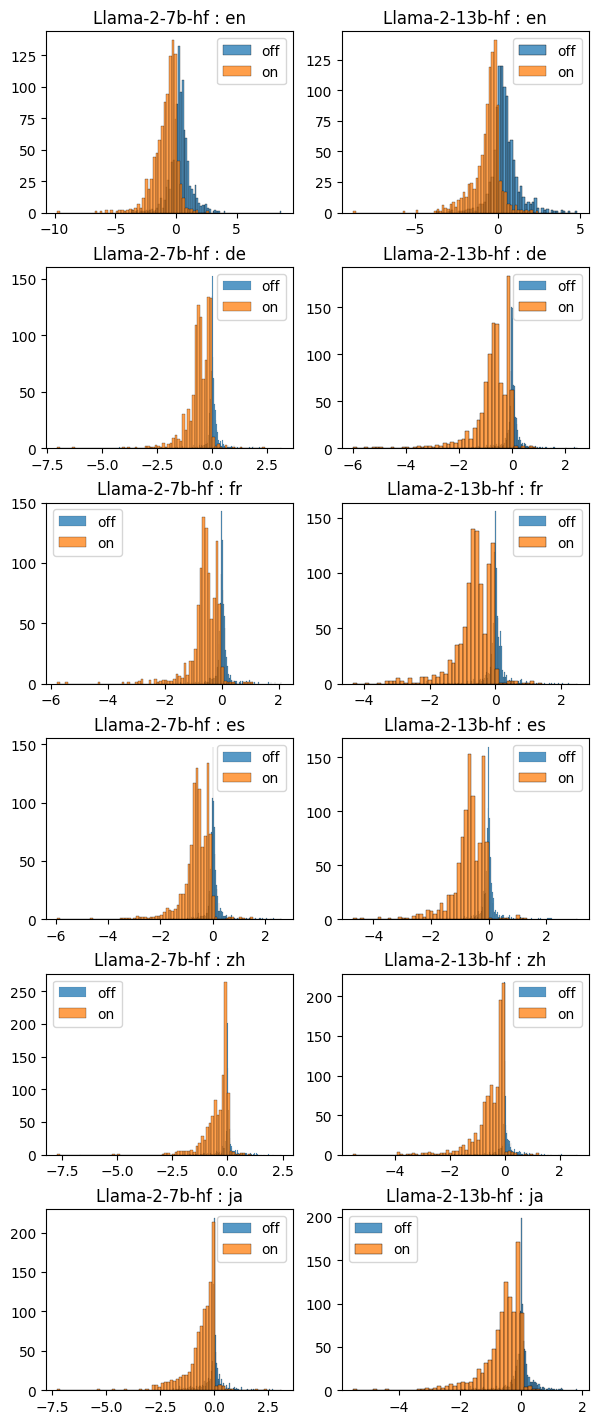}
\caption{Activation value difference of bottom-1000 neurons between target language(on) and non-target languages(off). x-axis: activation value of neurons. y-axis: frequency.}
\label{fig:activation_histogram_appendix_bottom_3}
\end{center}
\end{figure*}

Figure \ref{fig:activation_histogram_appendix_top_1}, \ref{fig:activation_histogram_appendix_top_2}, \ref{fig:activation_histogram_appendix_top_3}, \ref{fig:activation_histogram_appendix_bottom_1}, \ref{fig:activation_histogram_appendix_bottom_2}, and \ref{fig:activation_histogram_appendix_bottom_3} are histograms of the activation values for the top and bottom-1000 neurons. It was found that the top-1000 neurons activate positive values when we inputted positive text (on). In contrast, the bottom-1000 neurons tended to activate negative values when negative texts were given (off).

\section{Intervention in Neurons for Unconditional Text Generation}

\subsection{Effect of Intervention on Generated Language}

Table \ref{tab:intervention_language_appx} summarizes the probability of the target language occurrence in the generated texts before and after the intervention.

\subsection{Ablation Study of Changing the Number of Neurons for Intervention}
\label{appx:ablation_study}

\begin{figure*}[t]
\begin{center}
\includegraphics[width=0.85\linewidth]{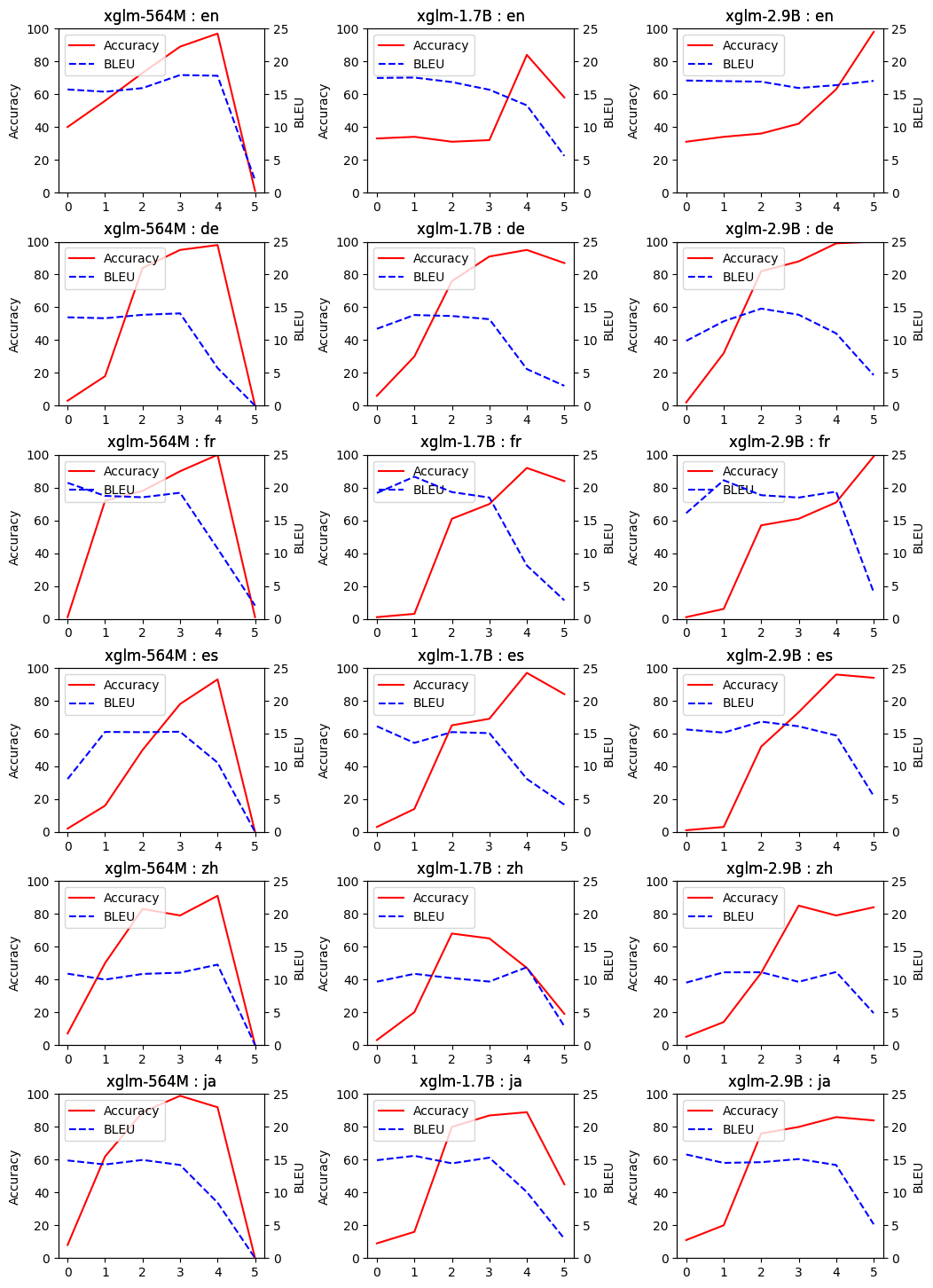}
\caption{Ablation study of changing the number of neurons to intervene. x-axis: $\log_{10}(k)$}
\label{fig:ablation_study_appendix1}
\end{center}
\end{figure*}

\begin{figure*}[t]
\begin{center}
\includegraphics[width=0.85\linewidth]{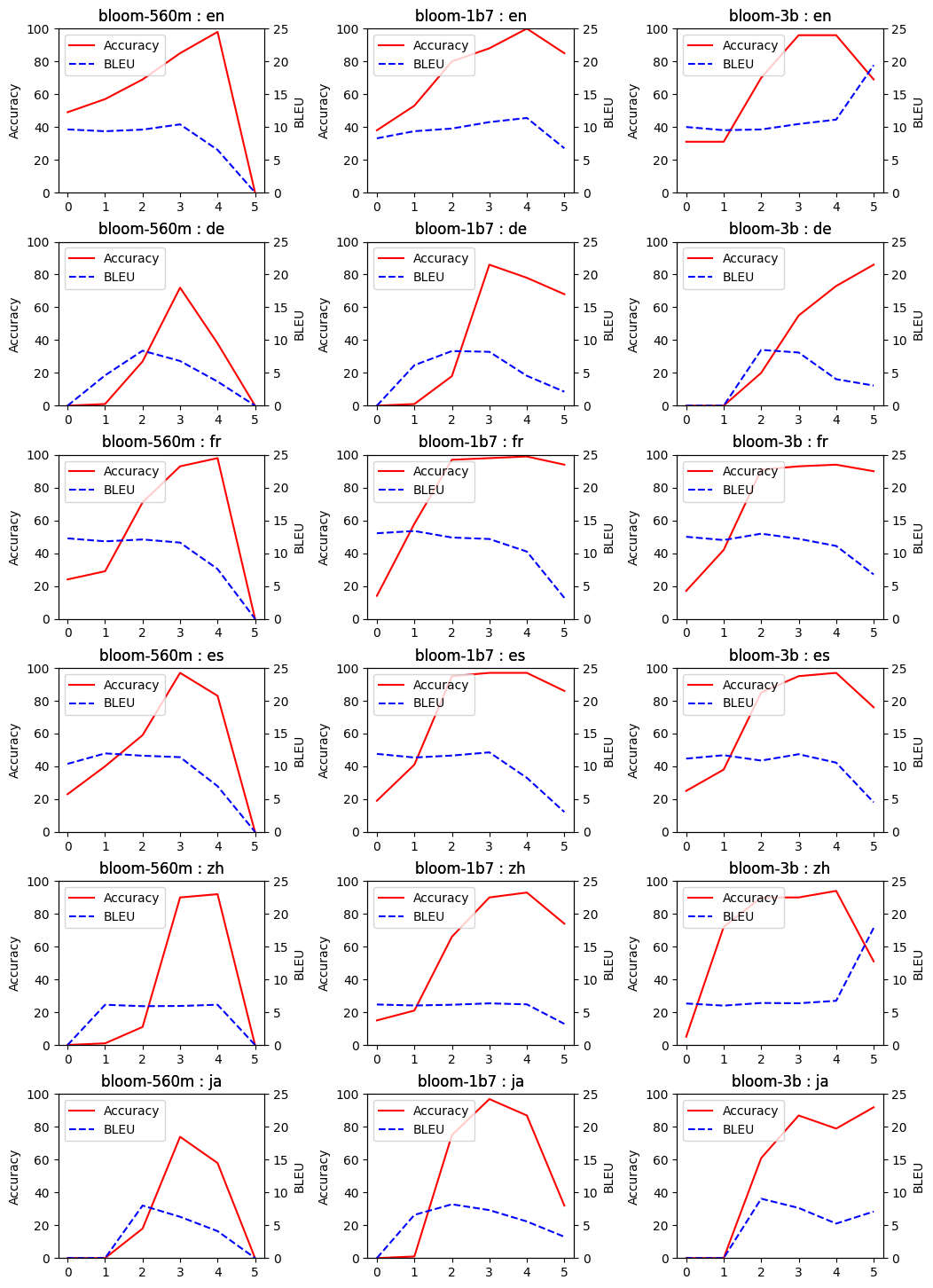}
\caption{Ablation study of changing the number of neurons to intervene. x-axis: $\log_{10}(k)$}
\label{fig:ablation_study_appendix2}
\end{center}
\end{figure*}

\begin{figure*}[t]
\begin{center}
\includegraphics[width=0.60\linewidth]{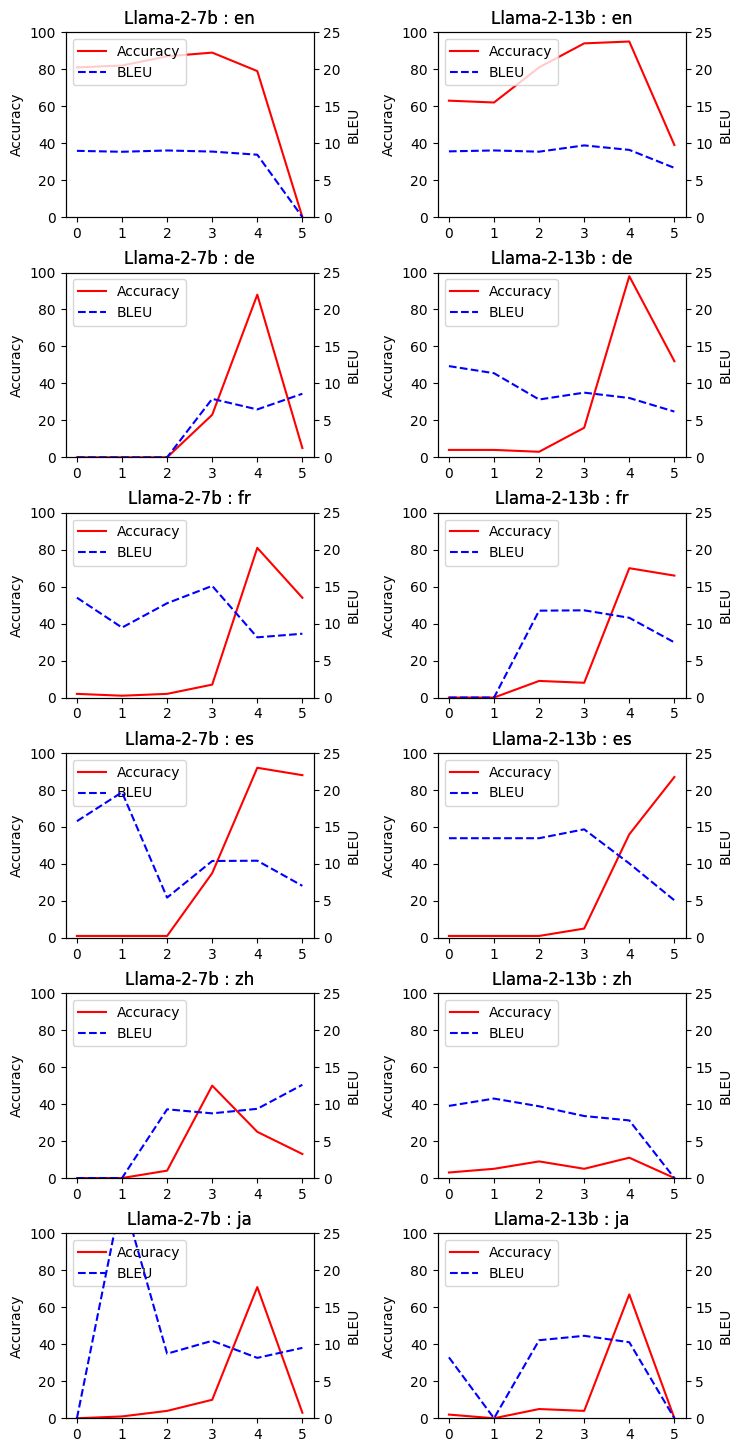}
\caption{Ablation study of changing the number of neurons to intervene. x-axis: $\log_{10}(k)$}
\label{fig:ablation_study_appendix3}
\end{center}
\end{figure*}

Figures \ref{fig:ablation_study_appendix1}, \ref{fig:ablation_study_appendix2}, and \ref{fig:ablation_study_appendix3} show the results of the ablation study when changing the number of neurons interventions. 

\subsection{Model-Generated Text Examples For Unconditional Setting}
\label{appx:generated_text_samples}

\begin{figure*}[t]
\begin{center}
\fbox{
\includegraphics[width=0.90\linewidth]{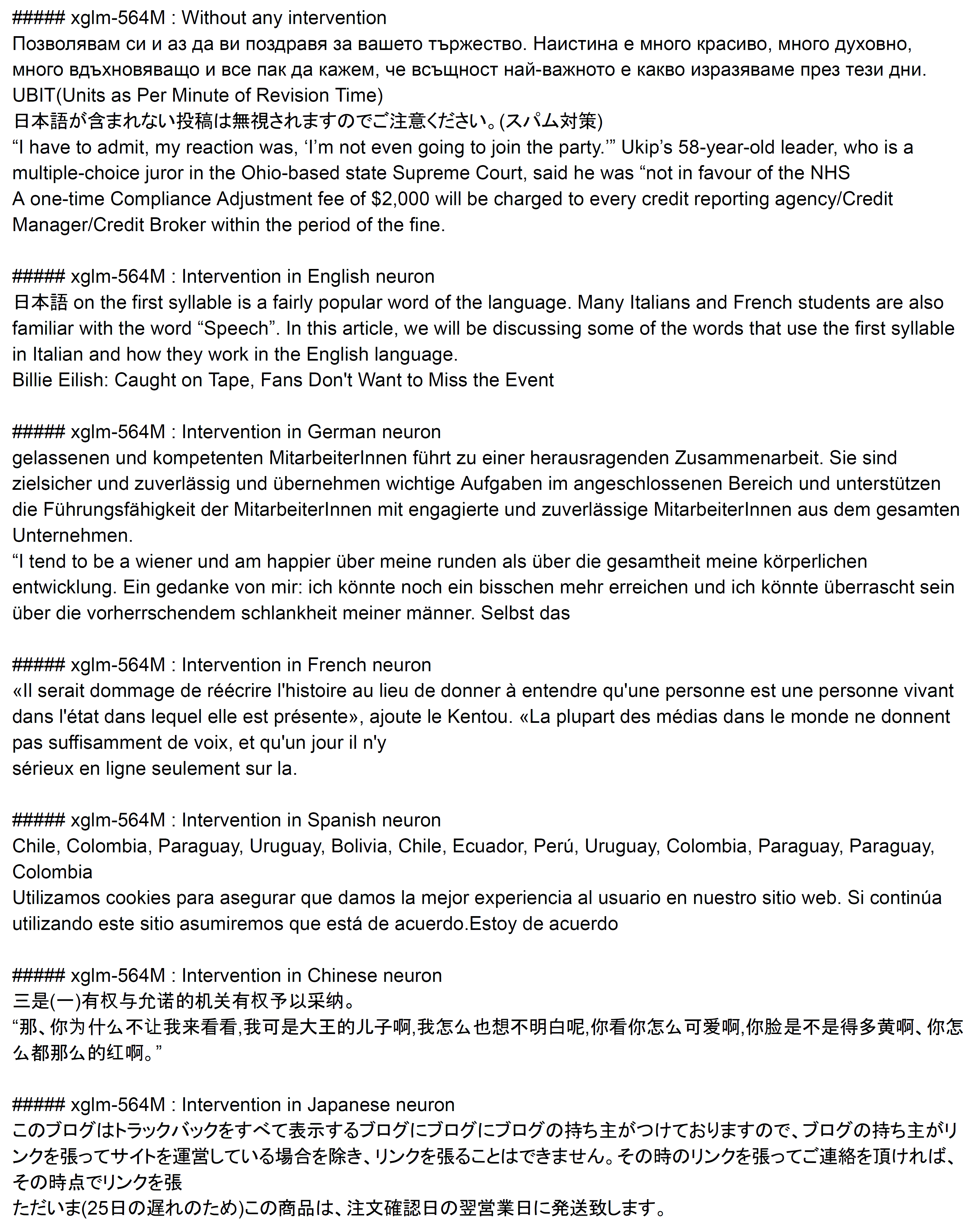}
}
\caption{Summary of Model-generated text examples by unconditional text generation setting}
\label{summary_text_sample_unconditional}
\end{center}
\end{figure*}

\begin{figure*}[t]
\begin{center}
\fbox{
\includegraphics[width=0.90\linewidth]{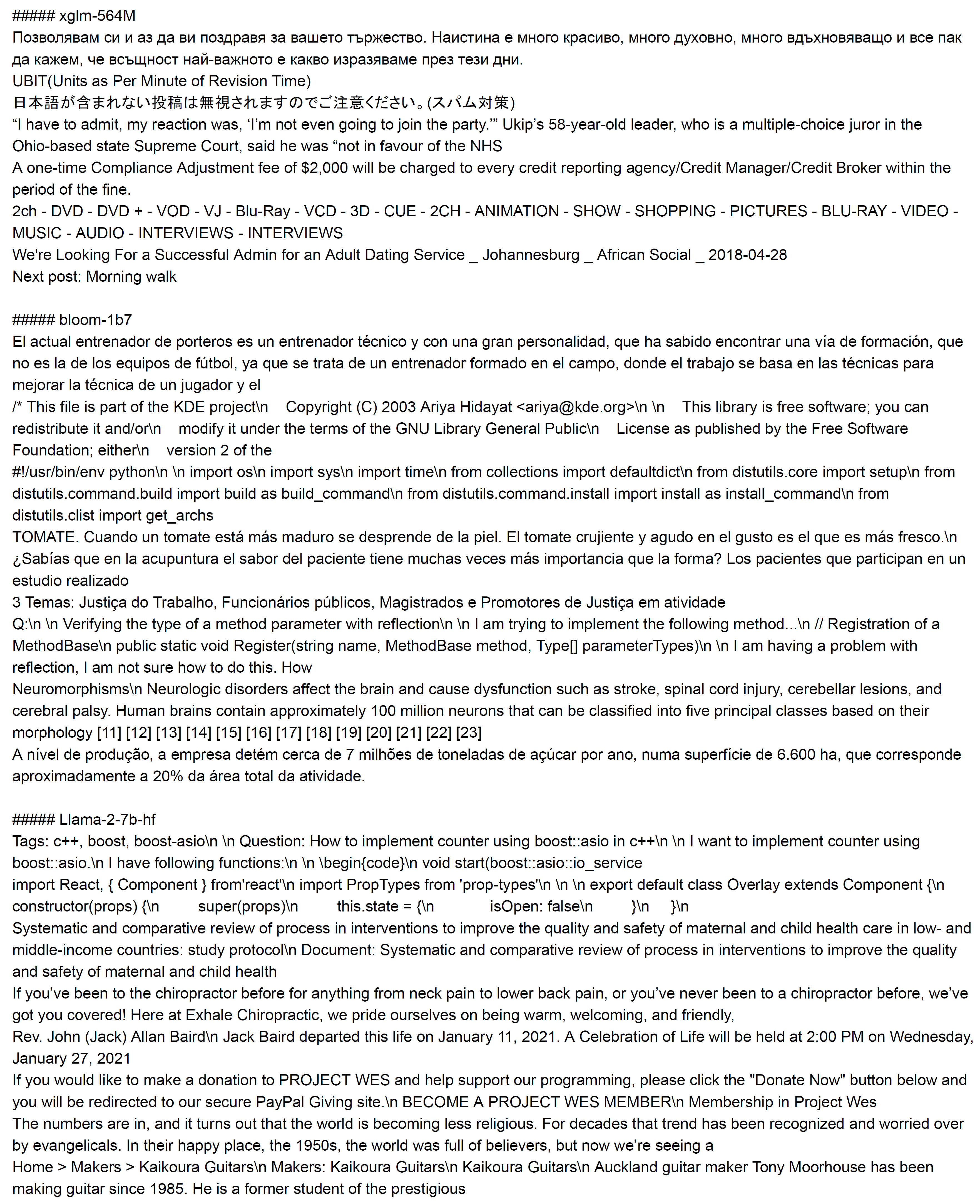}
}
\caption{Unconditionally model-generated text examples without interventions.}
\label{text_samples_natural}
\end{center}
\end{figure*}

\begin{figure*}[t]
\begin{center}
\fbox{
\includegraphics[width=0.90\linewidth]{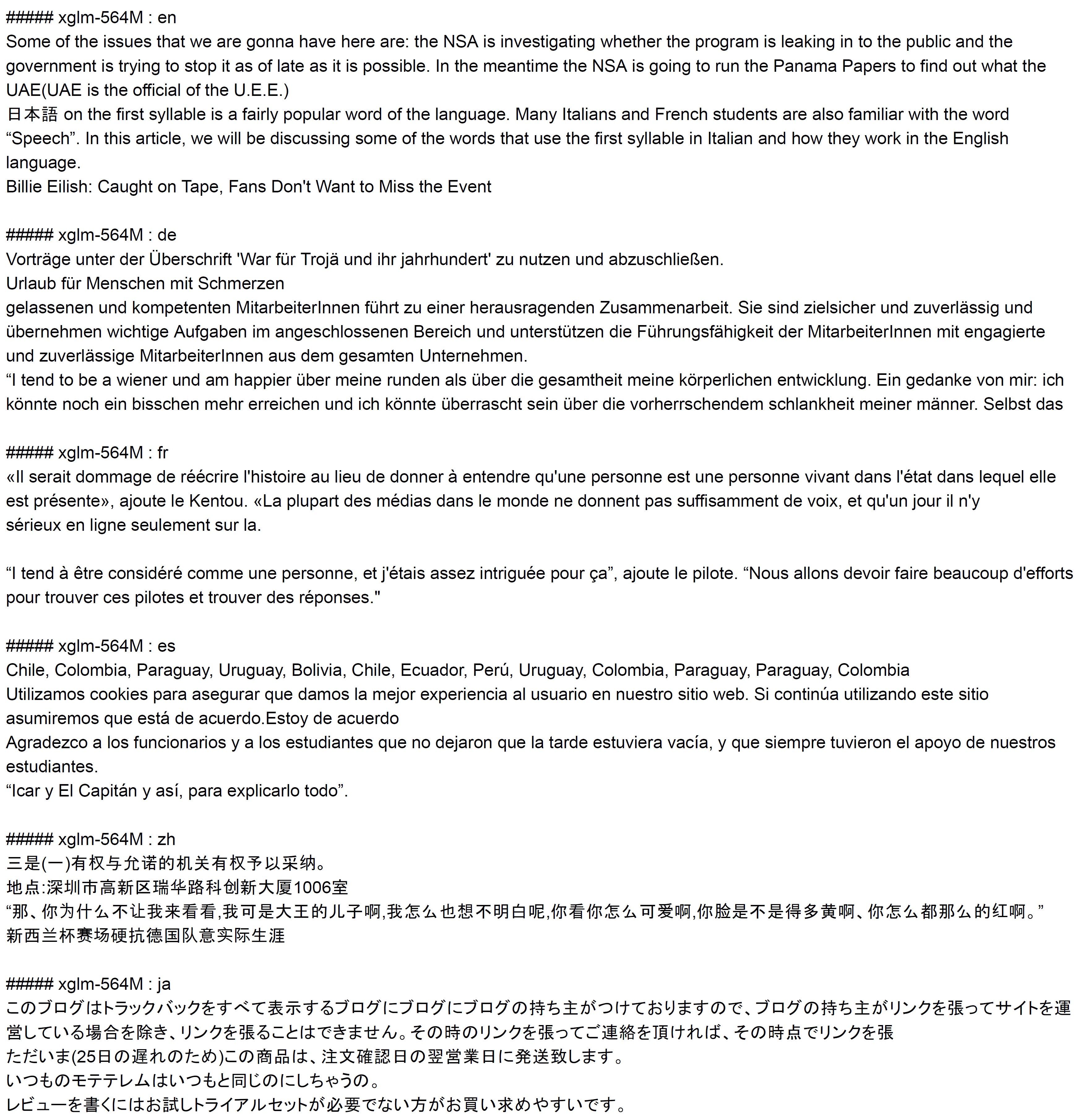}
}
\caption{Model-generated text examples from unconditional text generation settings with the top-1000 and bottom-1000 language-selective neurons intervention.}
\label{text_samples_intervention1}
\end{center}
\end{figure*}

\begin{figure*}[t]
\begin{center}
\fbox{
\includegraphics[width=0.90\linewidth]{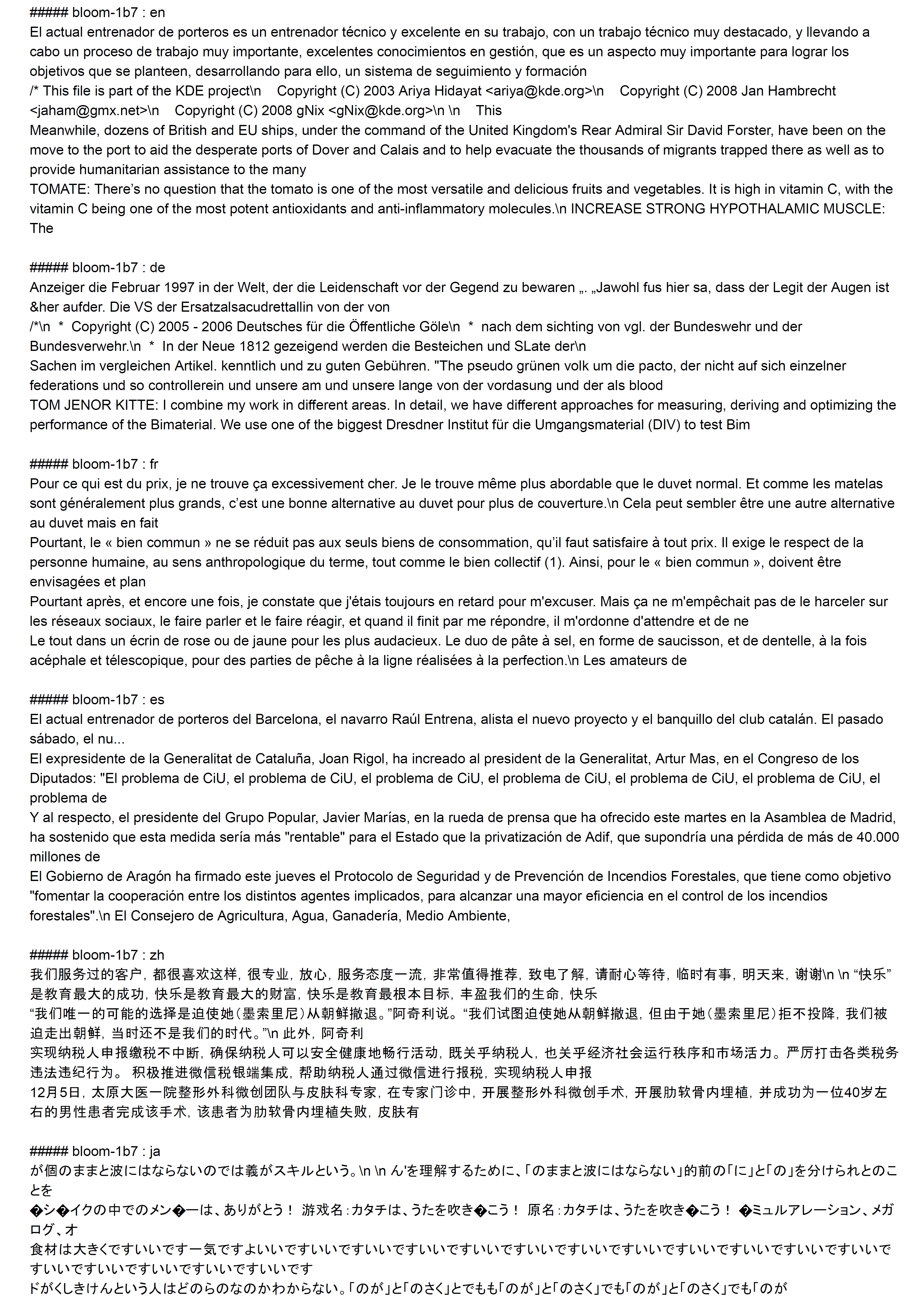}
}
\caption{Model-generated text examples from unconditional text generation settings with top-1000 and bottom-1000 language-selective neurons intervention.}
\label{text_samples_intervention2}
\end{center}
\end{figure*}

\begin{figure*}[t]
\begin{center}
\fbox{
\includegraphics[width=0.90\linewidth]{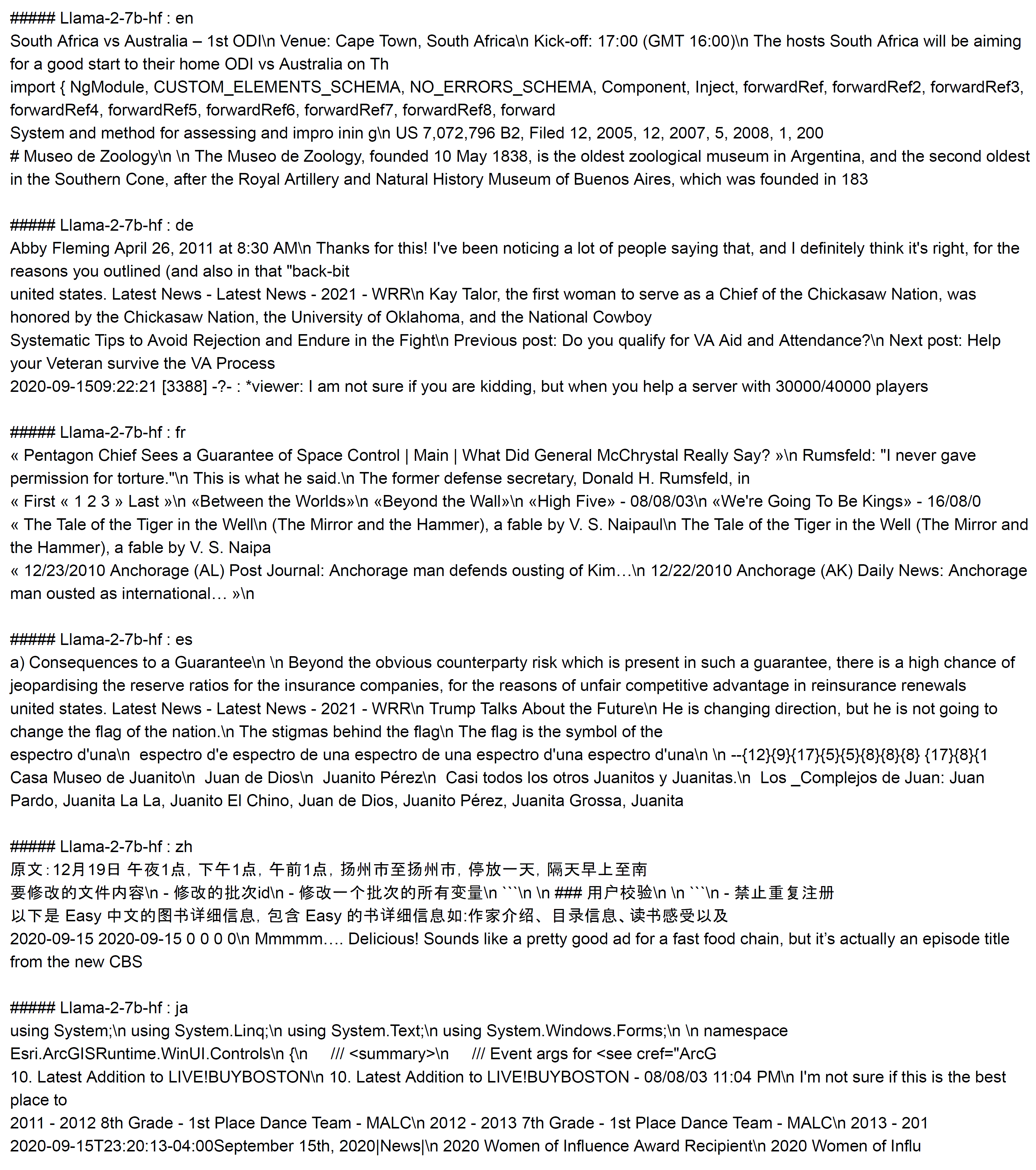}
}
\caption{Model-generated text examples from unconditional text generation settings with top-1000 and bottom-1000 language-selective neurons intervention.}
\label{text_samples_intervention3}
\end{center}
\end{figure*}

\begin{figure*}[t]
\begin{center}
\fbox{
\includegraphics[width=0.90\linewidth]{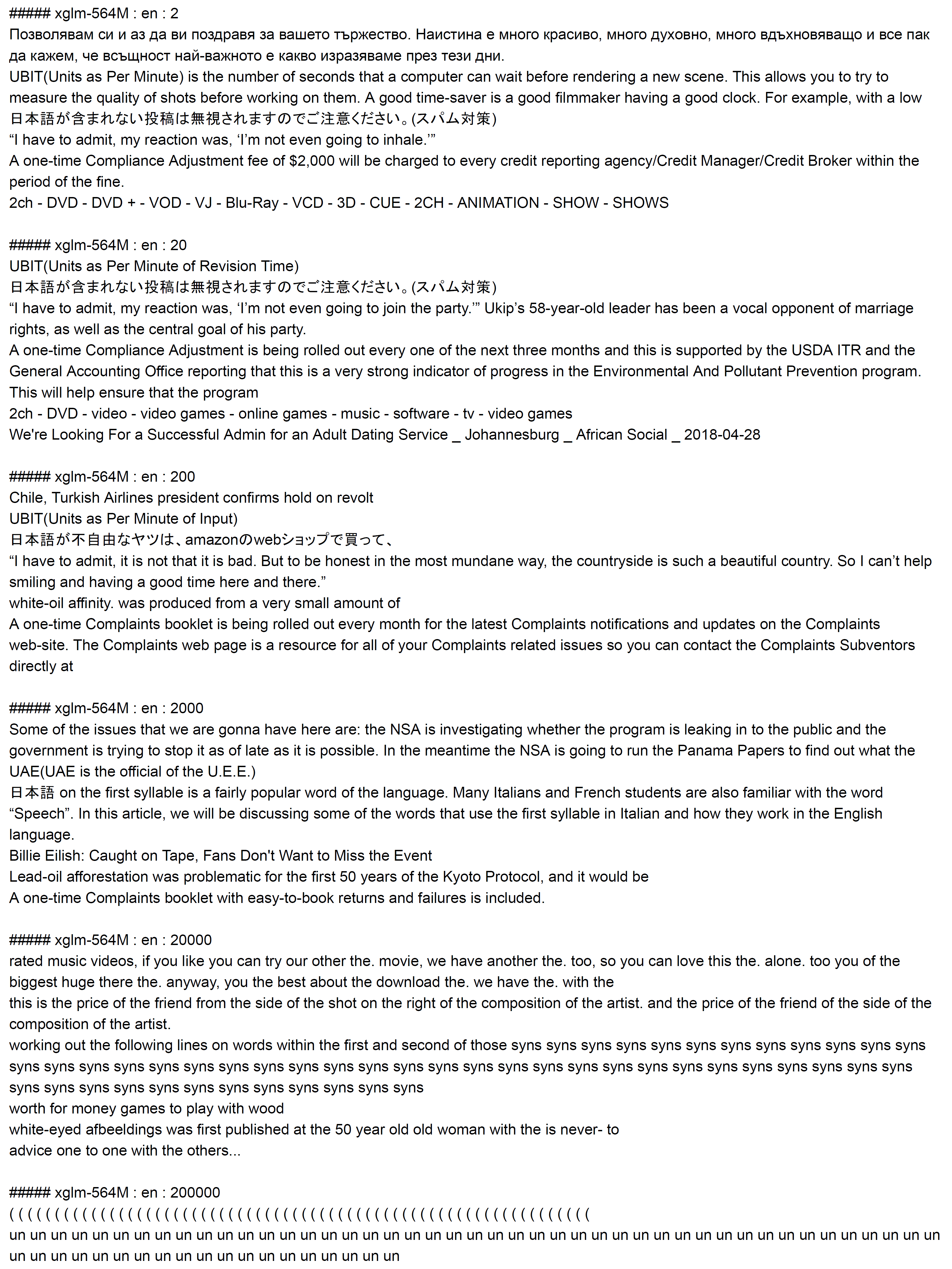}
}
\caption{Model-generated text examples from unconditional text generation settings by varying the number of interventions.}
\label{text_samples_intervention_en}
\end{center}
\end{figure*}

\begin{figure*}[t]
\begin{center}
\fbox{
\includegraphics[width=0.90\linewidth]{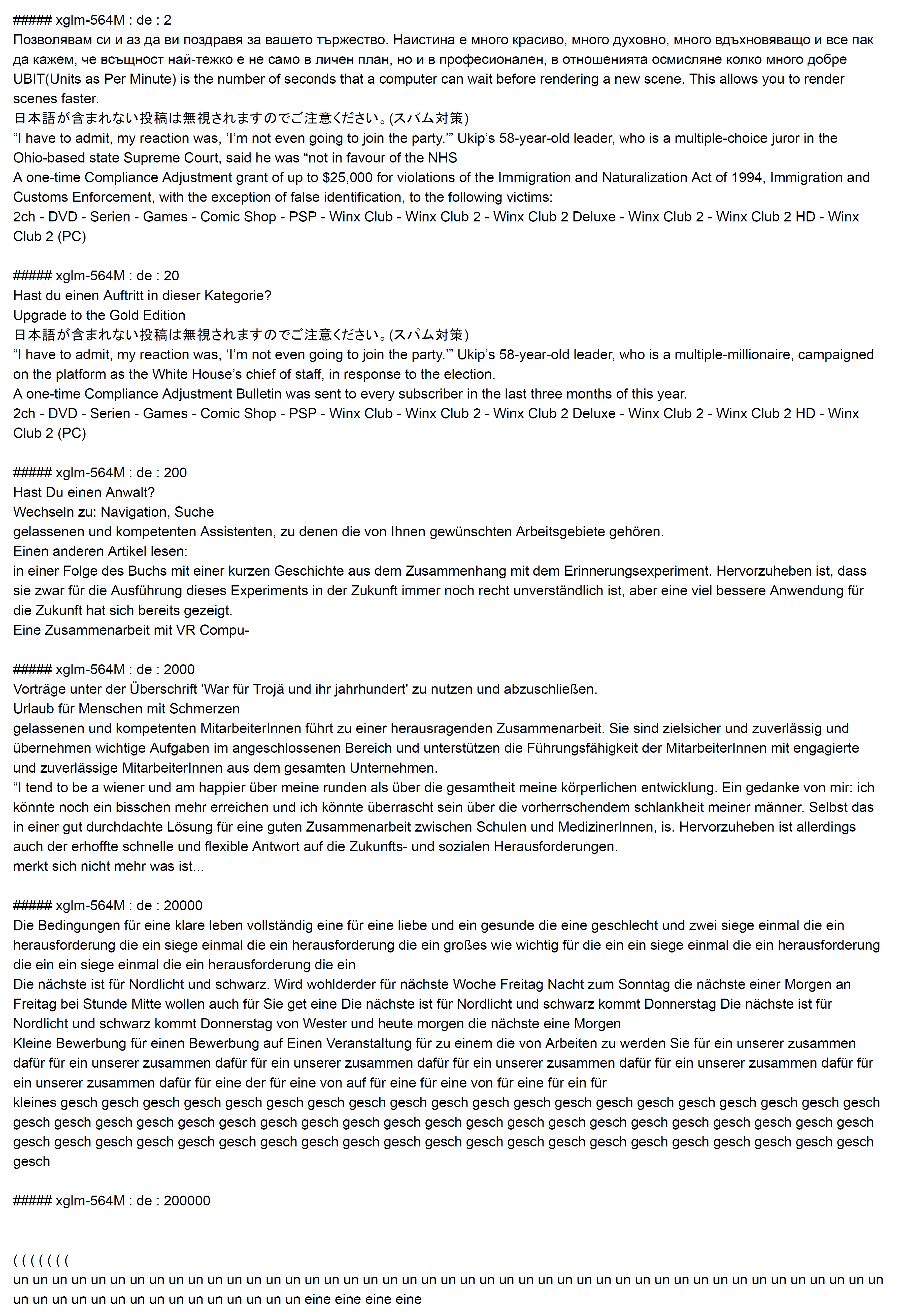}
}
\caption{Model-generated text examples from unconditional text generation settings by varying the number of interventions.}
\label{text_samples_intervention_de}
\end{center}
\end{figure*}

\begin{figure*}[t]
\begin{center}
\fbox{
\includegraphics[width=0.90\linewidth]{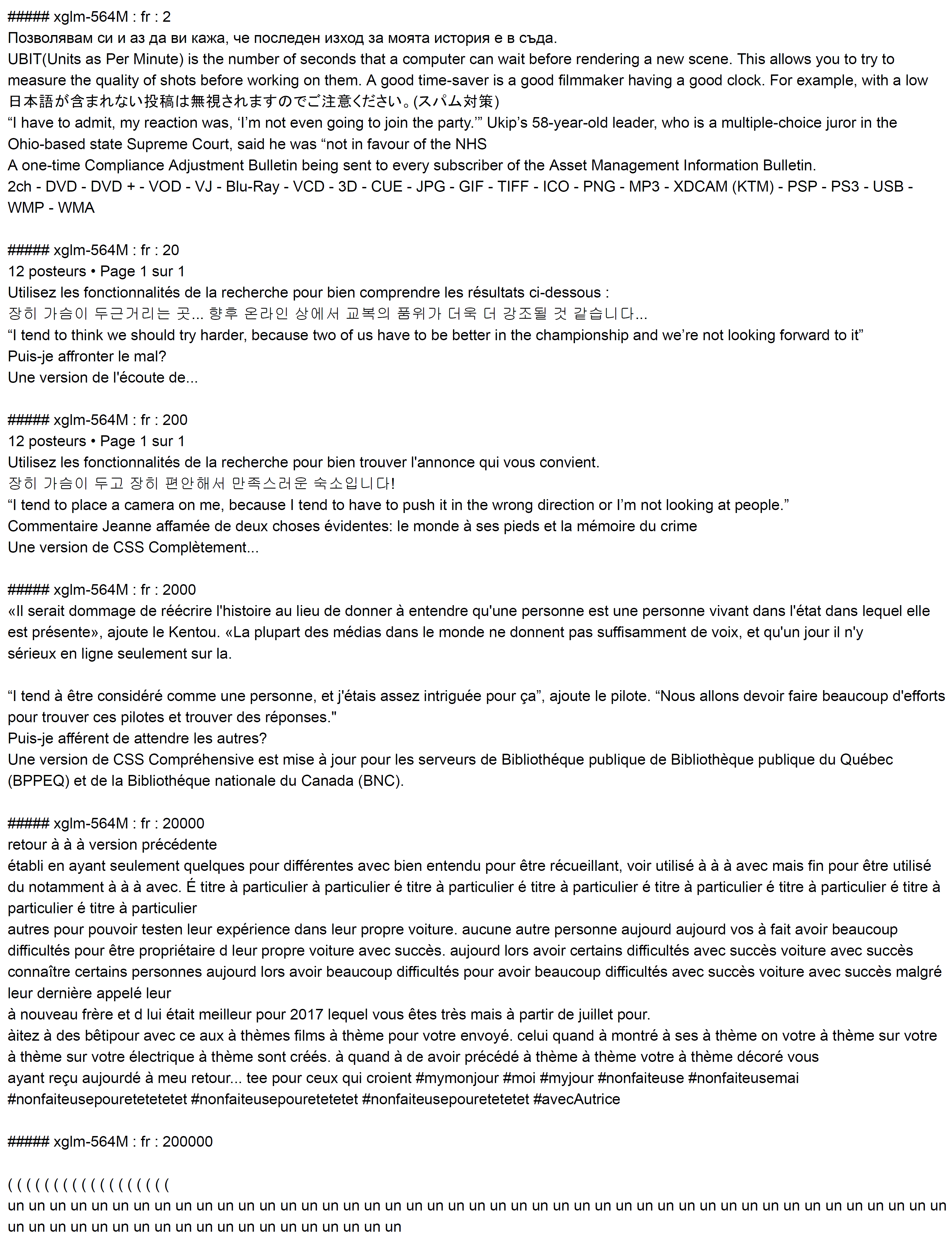}
}
\caption{Model-generated text examples from unconditional text generation settings by varying the number of interventions.}
\label{text_samples_intervention_fr}
\end{center}
\end{figure*}

\begin{figure*}[t]
\begin{center}
\fbox{
\includegraphics[width=0.90\linewidth]{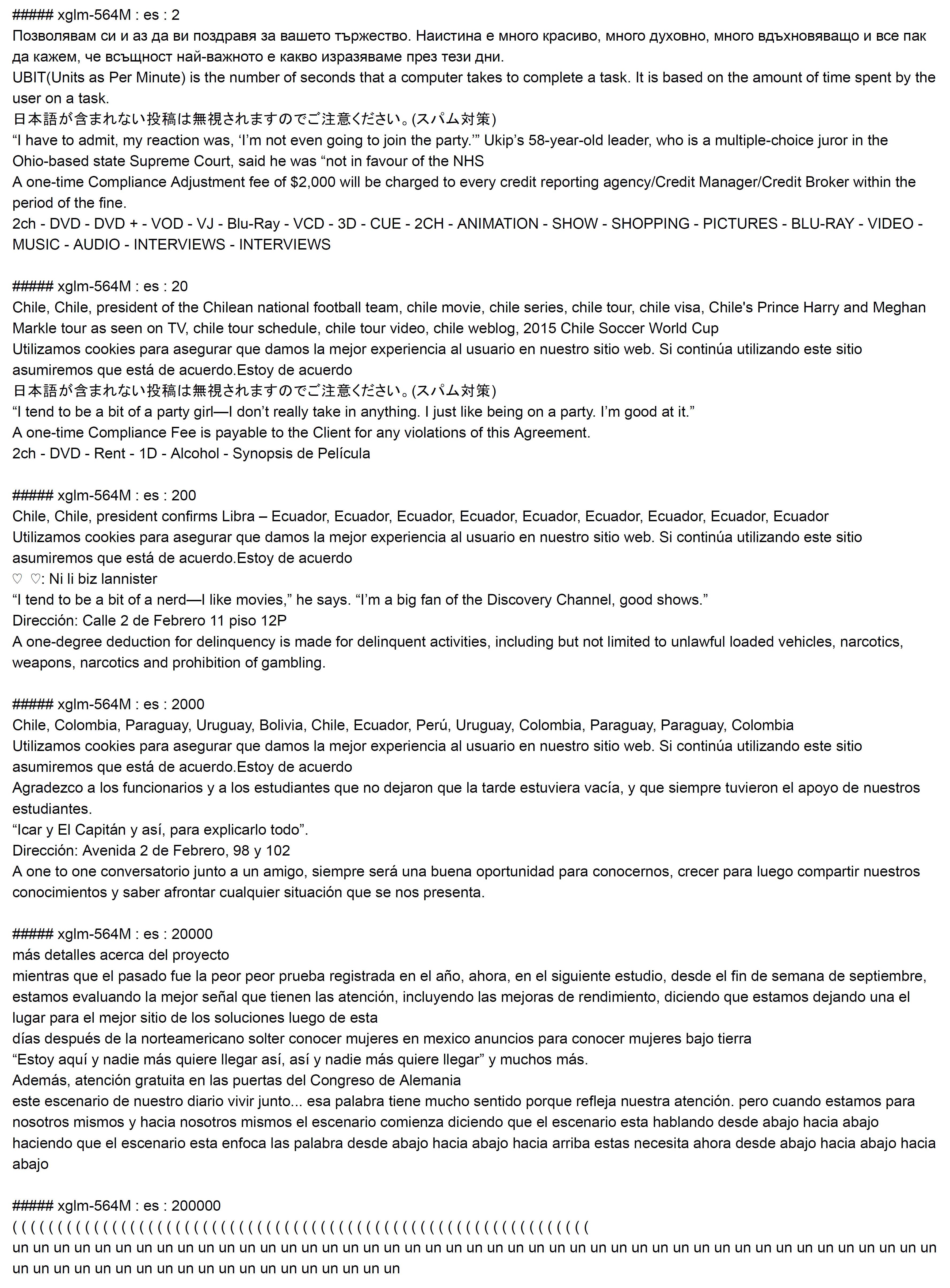}
}
\caption{Model-generated text examples from unconditional text generation settings by varying the number of interventions.}
\label{text_samples_intervention_es}
\end{center}
\end{figure*}

\begin{figure*}[t]
\begin{center}
\fbox{
\includegraphics[width=0.90\linewidth]{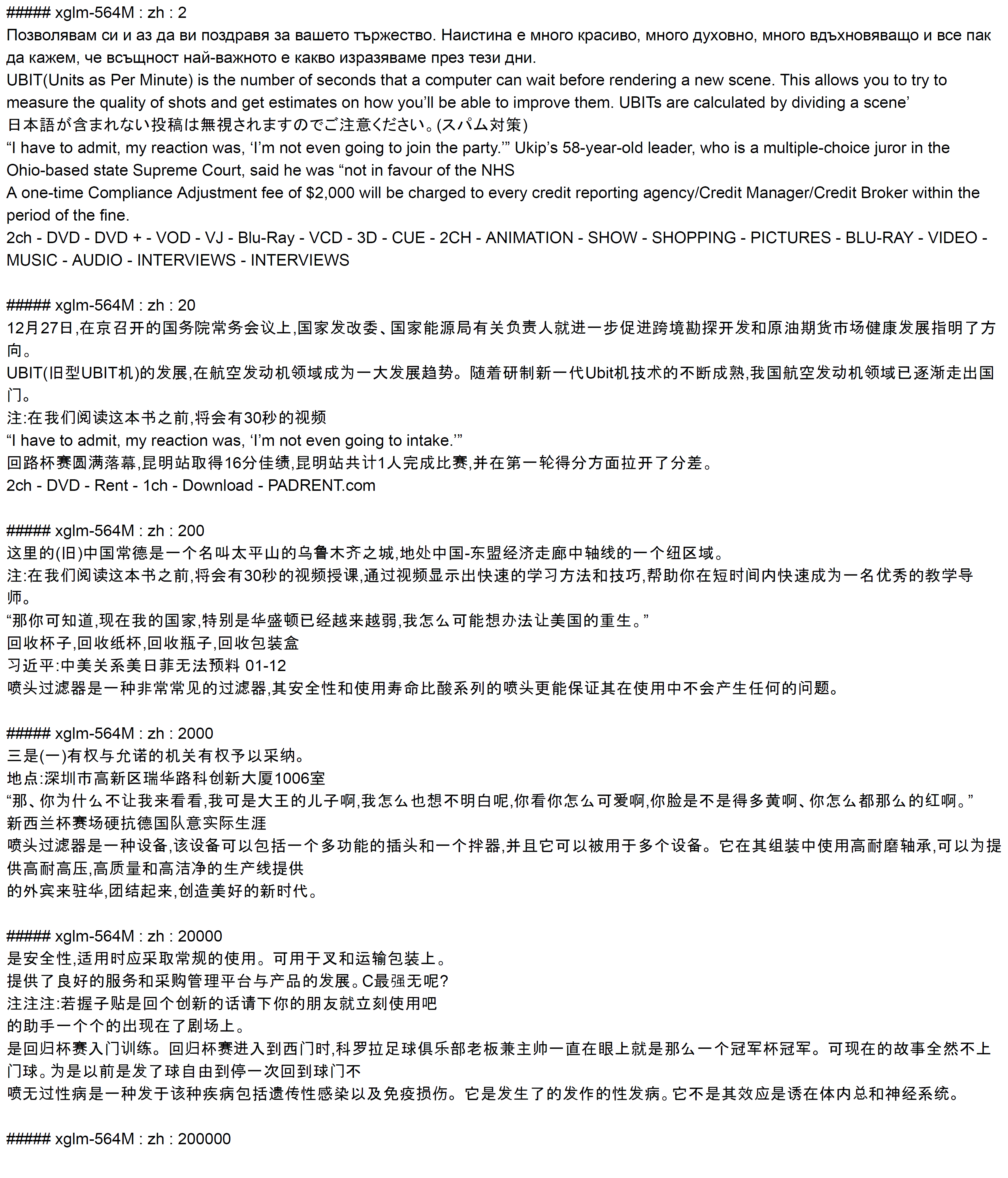}
}
\caption{Model-generated text examples from unconditional text generation settings by varying the number of interventions.}
\label{text_samples_intervention_zh}
\end{center}
\end{figure*}

\begin{figure*}[t]
\begin{center}
\fbox{
\includegraphics[width=0.90\linewidth]{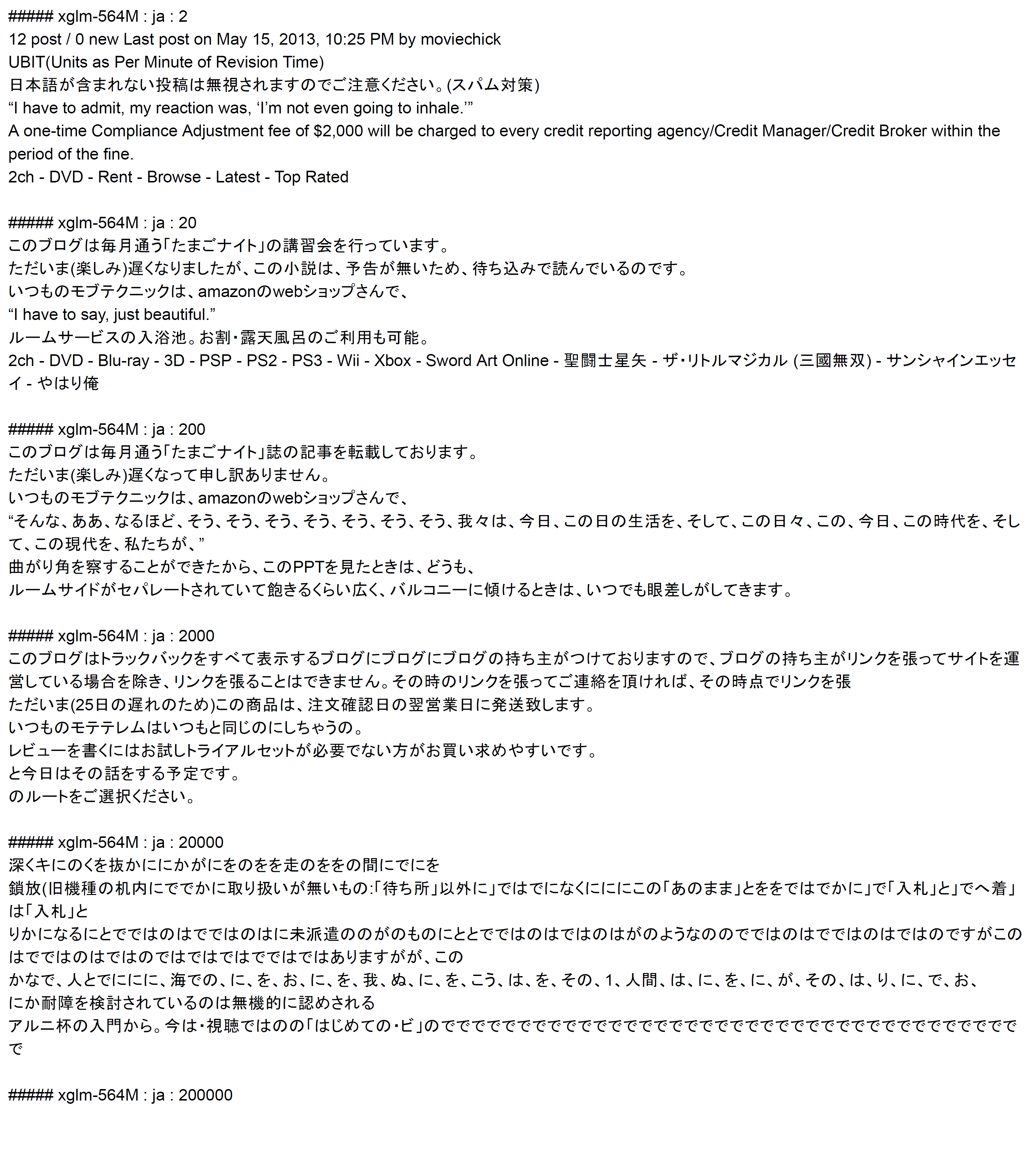}
}
\caption{Model-generated text examples from unconditional text generation settings by varying the number of interventions.}
\label{text_samples_intervention_ja}
\end{center}
\end{figure*}

\begin{itemize}
\setlength{\itemindent}{-2mm}
\item Figure \ref{summary_text_sample_unconditional} describes a summary of model-generated unconditional text examples.
\item Figure \ref{text_samples_natural} lists model-generated text examples without any interventions.
\item Figure \ref{text_samples_intervention1}, \ref{text_samples_intervention2}, \ref{text_samples_intervention3} list model-generated text examples with top-1000 and bottom-1000 neurons intervention.
\item Figure \ref{text_samples_intervention_en}, \ref{text_samples_intervention_de}, \ref{text_samples_intervention_fr}, \ref{text_samples_intervention_es}, \ref{text_samples_intervention_zh}, and \ref{text_samples_intervention_ja} list model-generated text examples by changing the number of neuron interventions.
\end{itemize}

\subsection{Detail Setting of Unconditional Text Generation}

A random sampling decoding method was used for unconditional text generation using the following settings in all experiments for unconditional text generation:
\begin{itemize}
\setlength{\itemindent}{-2mm}
\item temperature: 0.8
\item top-p: 0.9
\item maximum output length: 64
\item prompt (input token for models):
\begin{itemize}
    \setlength{\itemindent}{-9mm}
    \item XGLM: </s> is automatically set.
    \item BLOOM: nothing is automatically set. \\
    We explicitly set </s>.
    \item Llama2: <s> is automatically set.
\end{itemize}
\end{itemize}

\section{Intervention in Neurons for Conditional Text Generation}

\subsection{Effect of Intervention on Conditional Text Generation}

Tables \ref{tab:conditional_generation_iwslt}, \ref{tab:conditional_generation_wmt}, and \ref{tab:conditional_generation_flores200} summarize the probability of target language occurrence in the generated texts before and after the intervention.

\subsection{Model-Generated Text Examples For Conditional Setting}
\label{appx:generated_text_samples_conditional}

\begin{figure*}[t]
\begin{center}
\fbox{
\includegraphics[width=0.90\linewidth]{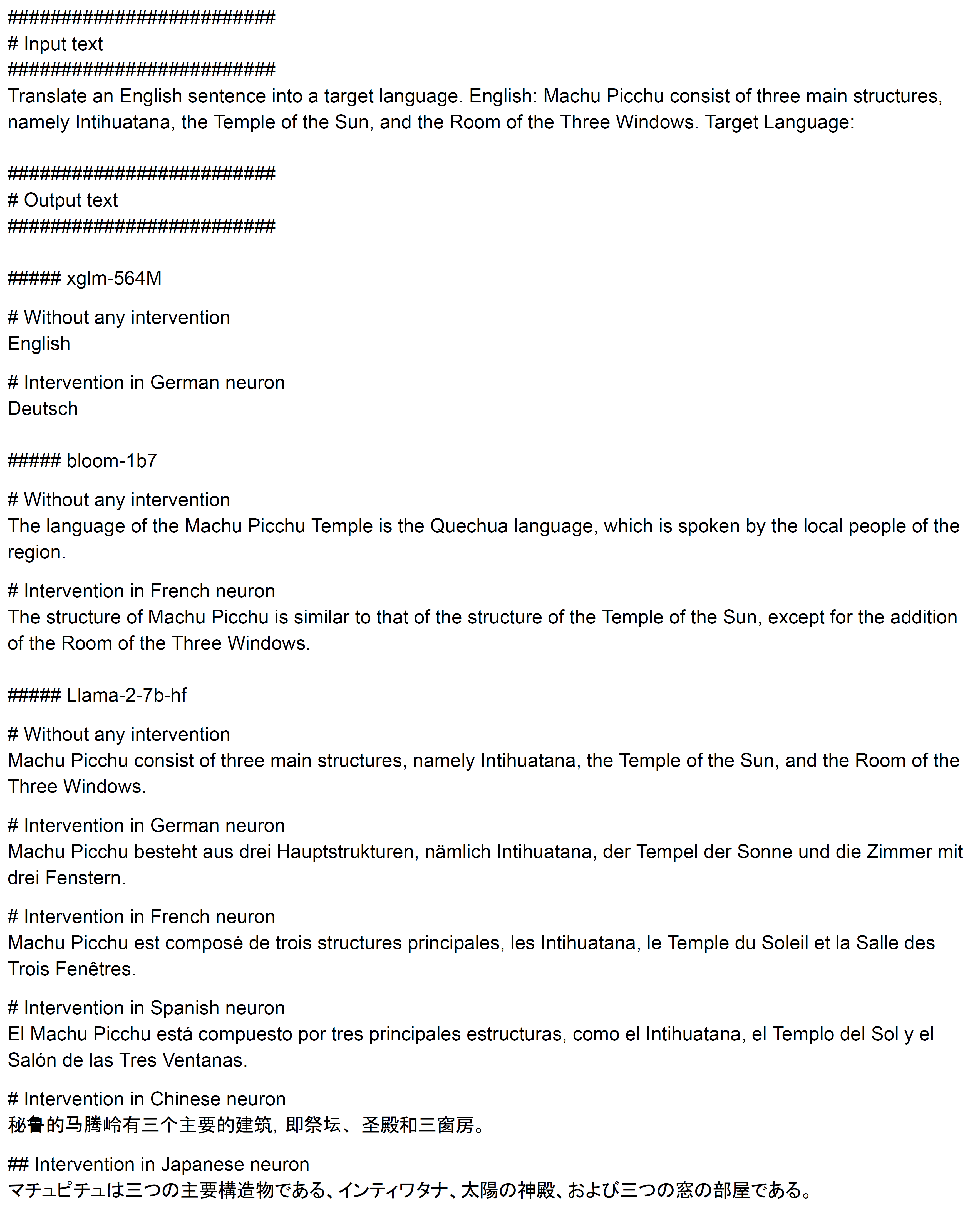}
}
\caption{Summary of model-generated text examples from conditional text generation settings}
\label{summary_text_sample_conditional}
\end{center}
\end{figure*}

\begin{figure*}[t]
\begin{center}
\fbox{
\includegraphics[width=0.90\linewidth]{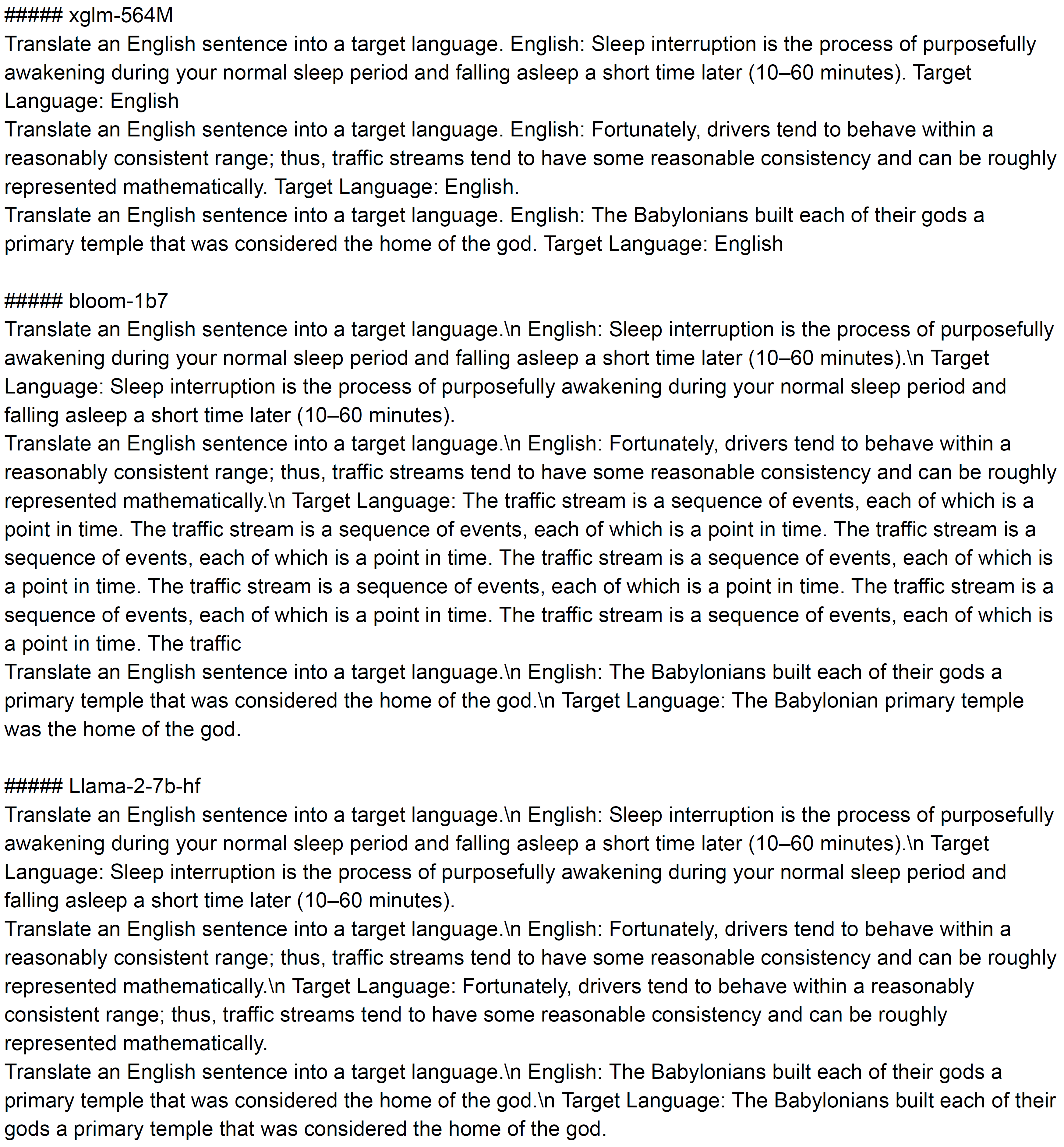}
}
\caption{Model-generated text examples from conditional text generation settings without interventions.}
\label{conditional_text_samples}
\end{center}
\end{figure*}

\begin{figure*}[t]
\begin{center}
\fbox{
\includegraphics[width=0.90\linewidth]{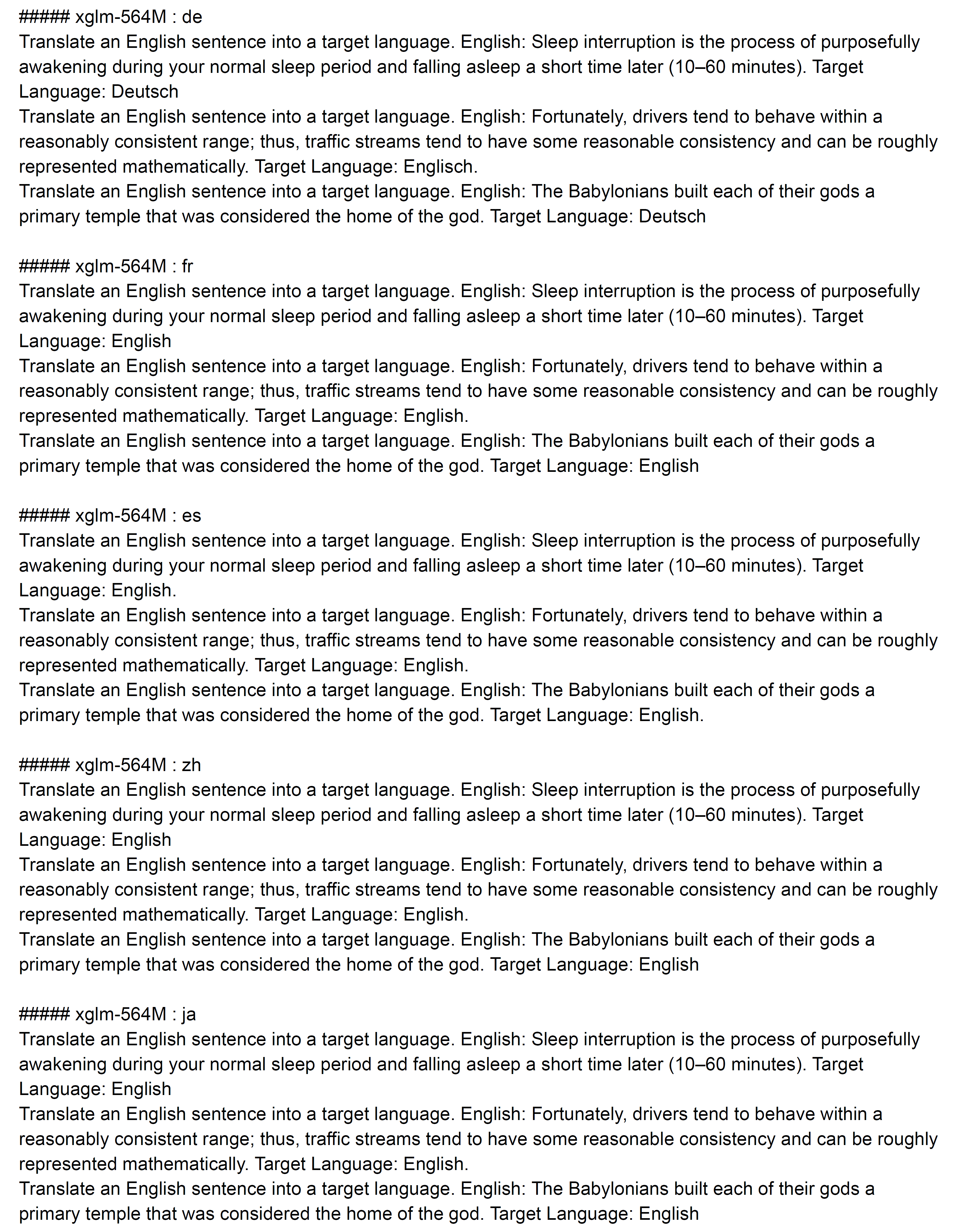}
}
\caption{Model-generated text examples from conditional text generation settings with top-1000 and bottom-1000 language-specific neurons intervention.}
\label{conditional_text_samples_intervention1}
\end{center}
\end{figure*}

\begin{figure*}[t]
\begin{center}
\fbox{
\includegraphics[width=0.90\linewidth]{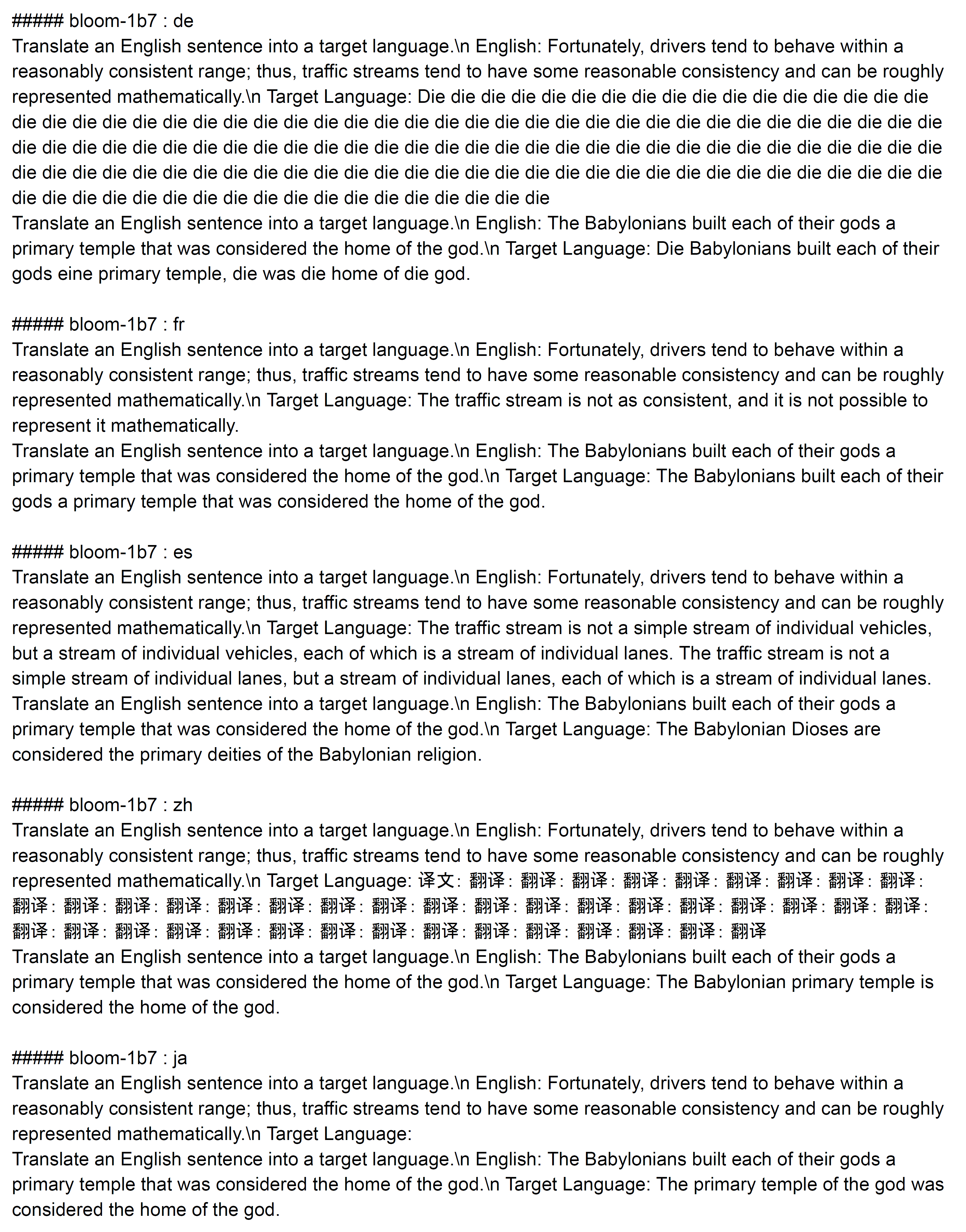}
}
\caption{Model-generated text examples from conditional text generation settings with top-1000 and bottom-1000 language-specific neurons intervention.}
\label{conditional_text_samples_intervention2}
\end{center}
\end{figure*}

\begin{figure*}[t]
\begin{center}
\fbox{
\includegraphics[width=0.90\linewidth]{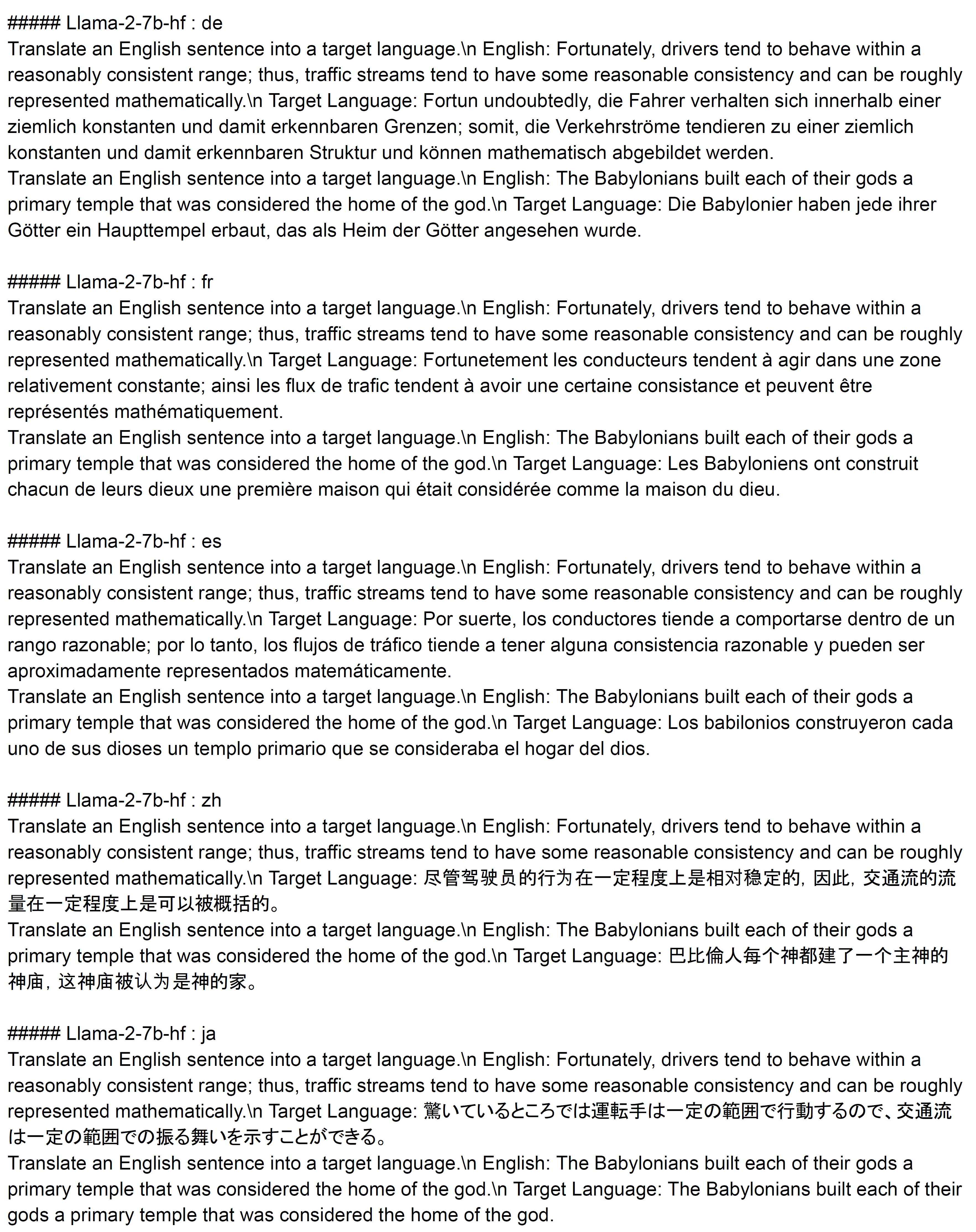}
}
\caption{Model-generated text examples from conditional text generation settings with top-1000 and bottom-1000 language-specific neurons intervention.}
\label{conditional_text_samples_intervention3}
\end{center}
\end{figure*}

\begin{itemize}
\item Figure \ref{summary_text_sample_conditional} describes a summary of model-generated conditional text examples.
\item Figure \ref{conditional_text_samples} lists model-generated text examples without any interventions.
\item Figures \ref{conditional_text_samples_intervention1}, \ref{conditional_text_samples_intervention2}, and \ref{conditional_text_samples_intervention3} list model-generated text examples with top-1000 and bottom-1000 neurons intervention.
\end{itemize}

\subsection{Detail Setting of Conditional Text Generation}

A greedy decoding method was used for conditional text generation.
For evaluation of machine translation, the first line of model-generated text (sentences before the first linebreak code ``\textyen n'') is used as translated sentences to omit useless subsequent sentences.

The following settings were used in all experiments for conditional text generation:

\begin{itemize}
\item maximum output length: 128
\end{itemize}

\section{Detail Setting of Datasets}

To create a language-specific text corpus, we mixed the following two datasets: \textbf{dev} split of PAWS-X \citep{yang-etal-2019-paws} and \textbf{test} split of FLORES-200 \citep{costa2022no}.
To create translation tasks for conditional text generation, we randomly sampled tasks from the \textbf{devtest} split of FLORES200, \textbf{test} split of IWSLT2017 \citep{cettolo2017overview}, and \textbf{test} split of WMT \citep{bojar-EtAl:2014:W14-33,bojar-EtAl:2016:WMT1,bojar-EtAl:2018:WMT1}.
All datasets were downloaded from HuggingFace \citep{wolf2019huggingface}.

\section{Detail Setting of BLEU-4 metrics}

We used NLTK library \citep{bird2009natural} to measure the BLEU  scores for both unconditional and conditional text generation. Specifically, the \href{https://www.nltk.org/api/nltk.translate.bleu_score.html#nltk.translate.bleu_score.sentence_bleu}{sentence\_bleu} function was used with method2 SmoothingFunction option for unconditional text generation. \href{https://www.nltk.org/api/nltk.translate.bleu_score.html#nltk.translate.bleu_score.corpus_bleu}{corpus\_bleu} function was used with method2 SmoothingFunction option for conditional text generation.
To enable the comparison of BLEU scores across models, we tokenized all texts using a multilingual tokenizer, XGLM, whose pre-training corpus includes a large proportion of texts in the six target languages \citep{lin2021few}.

\section{License}

\subsection{Model}

\begin{itemize}
    \setlength{\itemindent}{-2mm}
    \item XGLM: MIT [\href{https://huggingface.co/facebook/xglm-564M}{link}]
    \item BLOOM: bigscience-bloom-rail-1.0 [\href{https://huggingface.co/bigscience/bloom#languages}{link}]
    \item Llama2: Meta license [\href{https://huggingface.co/meta-llama/Llama-2-7b}{link}]
\end{itemize}

\subsection{Dataset}

\begin{itemize}
    \setlength{\itemindent}{-2mm}
    \item PAWS-X: No License (Free to use) [\href{https://huggingface.co/datasets/paws-x}{link}]
    \item FLORES200: cc-by-sa-4.0 [\href{https://huggingface.co/datasets/Muennighoff/flores200}{link}]
    \item IWSLT2017: cc-by-nc-nd-4.0 [\href{https://huggingface.co/datasets/iwslt2017}{link}]
    \item WMT14: Unknown [\href{https://huggingface.co/datasets/wmt14}{link}]
    \item WMT16: Unknown [\href{https://huggingface.co/datasets/wmt16}{link}]
    \item WMT18: Unknown [\href{https://huggingface.co/datasets/wmt18}{link}]
\end{itemize}

\section{Total computation for Experiments}

We executed the experiments mainly for running the inference (both identification and intervention of language-specific neurons) for each model using the following number of A100(40GB) GPUs and approximate computing hours per run.
We run GPUs 60 times per model (6 languages $\times$ (1 for identification of language-specific neurons + 6 for unconditional text generation by changing the number of neurons to intervene + 3 for conditional text generation)) for the production run.
The computational resource of AI Bridging Cloud Infrastructure (ABCI) provided by the National Institute of Advanced Industrial Science and Technology (AIST) was used for the experiments.

\begin{itemize}
    \setlength{\itemindent}{-2mm}
    \item XGLM 564M: 1GPU $\times$ 0.5hrs $\times$ 60 runs.
    \item XGLM 1.7B: 1GPU $\times$ 0.5hrs $\times$ 60 runs.
    \item XGLM 2.9B: 1GPU $\times$ 0.5hrs $\times$ 60 runs.
    \item BLOOM 560M: 1GPU $\times$ 0.5hrs $\times$ 60 runs.
    \item BLOOM 1.7B: 1GPU $\times$ 0.5hrs $\times$ 60 runs.
    \item BLOOM 3B: 1GPU $\times$ 0.5hrs $\times$ 60 runs.
    \item Llama2 7B: 8GPUs $\times$ 1hrs $\times$ 60 runs.
    \item Llama2 13B: 8GPUs $\times$ 3hrs $\times$ 60 runs.
\end{itemize}



\clearpage


\clearpage

\clearpage

\clearpage

\clearpage

\clearpage

\clearpage

\clearpage

\clearpage


\clearpage
\begin{table*}[t]\centering
\scalebox{1.0}[0.90]{
\begin{tabular}{lrrrrr}\toprule
Model &Language &Top &Middle &Bottom \\
\midrule
xglm-564M &en &[0.58, 0.56] &[1.33, 1.45] &[0.68, 0.51] \\
xglm-564M &de &[0.66, 0.61] &[1.29, 1.41] &[0.81, 0.65] \\
xglm-564M &fr &[0.65, 0.53] &[1.4, 1.48] &[0.58, 0.52] \\
xglm-564M &es &[0.51, 0.52] &[1.31, 1.4] &[0.43, 0.53] \\
xglm-564M &zh &[0.65, 0.63] &[1.3, 1.42] &[0.86, 0.66] \\
xglm-564M &ja &[0.62, 0.63] &[1.19, 1.25] &[0.85, 0.78] \\
\midrule
xglm-1.7B &en &[0.68, 0.48] &[1.29, 1.45] &[0.52, 0.5] \\
xglm-1.7B &de &[0.67, 0.54] &[1.34, 1.57] &[0.8, 0.59] \\
xglm-1.7B &fr &[0.72, 0.52] &[1.29, 1.45] &[0.67, 0.51] \\
xglm-1.7B &es &[0.69, 0.49] &[1.22, 1.43] &[0.6, 0.48] \\
xglm-1.7B &zh &[0.63, 0.5] &[1.24, 1.3] &[0.61, 0.47] \\
xglm-1.7B &ja &[0.56, 0.49] &[1.26, 1.26] &[0.56, 0.56] \\
\midrule
xglm-2.9B &en &[0.54, 0.36] &[1.28, 1.43] &[0.84, 0.43] \\
xglm-2.9B &de &[0.5, 0.47] &[1.19, 1.52] &[0.99, 0.56] \\
xglm-2.9B &fr &[0.42, 0.46] &[1.21, 1.47] &[0.56, 0.47] \\
xglm-2.9B &es &[0.38, 0.41] &[1.19, 1.5] &[0.46, 0.4] \\
xglm-2.9B &zh &[0.59, 0.39] &[1.21, 1.44] &[0.69, 0.39] \\
xglm-2.9B &ja &[0.5, 0.37] &[1.14, 1.36] &[0.56, 0.4] \\
\midrule
bloom-560m &en &[0.56, 0.47] &[1.26, 1.22] &[0.48, 0.39] \\
bloom-560m &de &[1.74, 0.66] &[0.94, 1.08] &[1.7, 0.77] \\
bloom-560m &fr &[0.56, 0.48] &[1.2, 1.14] &[0.69, 0.44] \\
bloom-560m &es &[0.52, 0.45] &[1.27, 1.26] &[0.44, 0.4] \\
bloom-560m &zh &[0.69, 0.56] &[1.17, 1.15] &[0.67, 0.52] \\
bloom-560m &ja &[0.46, 0.57] &[1.28, 1.2] &[0.64, 0.68] \\
\midrule
bloom-1b7 &en &[0.54, 0.48] &[1.26, 1.24] &[0.56, 0.37] \\
bloom-1b7 &de &[2.14, 0.62] &[1.1, 1.2] &[1.22, 0.64] \\
bloom-1b7 &fr &[0.61, 0.47] &[1.16, 1.17] &[0.63, 0.4] \\
bloom-1b7 &es &[0.65, 0.46] &[1.24, 1.18] &[0.64, 0.4] \\
bloom-1b7 &zh &[0.8, 0.52] &[1.27, 1.31] &[0.71, 0.53] \\
bloom-1b7 &ja &[0.8, 0.49] &[1.29, 1.23] &[0.8, 0.63] \\
\midrule
bloom-3b &en &[0.56, 0.42] &[1.23, 1.23] &[0.56, 0.36] \\
bloom-3b &de &[2.12, 0.65] &[1.08, 1.15] &[1.35, 0.62] \\
bloom-3b &fr &[0.83, 0.43] &[1.27, 1.26] &[0.67, 0.36] \\
bloom-3b &es &[0.95, 0.44] &[1.2, 1.12] &[0.66, 0.36] \\
bloom-3b &zh &[0.89, 0.48] &[1.19, 1.17] &[0.58, 0.45] \\
bloom-3b &ja &[0.82, 0.47] &[1.28, 1.27] &[0.75, 0.57] \\
\midrule
Llama-2-7b-hf &en &[0.91, 0.33] &[1.28, 1.38] &[0.96, 0.35] \\
Llama-2-7b-hf &de &[1.06, 0.53] &[1.4, 1.63] &[0.75, 0.43] \\
Llama-2-7b-hf &fr &[1.37, 0.56] &[1.2, 1.39] &[1.03, 0.45] \\
Llama-2-7b-hf &es &[1.76, 0.58] &[1.3, 1.69] &[1.17, 0.46] \\
Llama-2-7b-hf &zh &[0.48, 0.7] &[1.2, 1.33] &[0.51, 0.57] \\
Llama-2-7b-hf &ja &[0.6, 0.53] &[1.17, 1.27] &[0.56, 0.42] \\
\midrule
Llama-2-13b-hf &en &[0.65, 0.36] &[1.25, 1.32] &[0.83, 0.39] \\
Llama-2-13b-hf &de &[0.94, 0.53] &[1.26, 1.41] &[0.57, 0.39] \\
Llama-2-13b-hf &fr &[0.98, 0.51] &[1.22, 1.51] &[0.68, 0.39] \\
Llama-2-13b-hf &es &[1.68, 0.61] &[1.29, 1.7] &[1.11, 0.46] \\
Llama-2-13b-hf &zh &[0.48, 0.52] &[1.12, 1.2] &[0.48, 0.48] \\
Llama-2-13b-hf &ja &[0.63, 0.38] &[1.11, 1.21] &[0.55, 0.33] \\
\bottomrule
\end{tabular}
}
\caption{Estimation of Beta distribution parameters from the histogram of neurons across layers in top-1000, middle-1000, and bottom-1000 groups, respectively.}
\label{tab:beta_distribution}
\end{table*}


\clearpage

\begin{table*}[t]\centering
    \begin{tabular}{cc}

      \begin{minipage}[t]{0.45\hsize}
            \subcaption{xglm-564M}
            \label{tab:cross_check_1}
            \scalebox{0.9}{
            \begin{tabular}{lrrrrrrr}\toprule
            &de &en &es &fr &ja &zh \\\midrule
            de &2000 &41 &74 &39 &44 &34 \\
            en &41 &2000 &34 &41 &49 &40 \\
            es &74 &34 &2000 &57 &77 &22 \\
            fr &39 &41 &57 &2000 &21 &93 \\
            ja &44 &49 &77 &21 &2000 &27 \\
            zh &34 &40 &22 &93 &27 &2000 \\
            \bottomrule
            \end{tabular}
            }
      \end{minipage}
      \hspace{0.04\columnwidth} 
      \hfill
      \begin{minipage}[t]{0.45\hsize}
            \subcaption{xglm-1.7B}
            \label{tab:cross_check_2}
            \scalebox{0.9}{
            \begin{tabular}{lrrrrrrr}\toprule
            &de &en &es &fr &ja &zh \\\midrule
            de &2000 &12 &14 &9 &43 &9 \\
            en &12 &2000 &21 &22 &23 &28 \\
            es &14 &21 &2000 &60 &22 &17 \\
            fr &9 &22 &60 &2000 &7 &30 \\
            ja &43 &23 &22 &7 &2000 &30 \\
            zh &9 &28 &17 &30 &30 &2000 \\
            \bottomrule
            \end{tabular}
            }
      \end{minipage} \\\\

      \begin{minipage}[t]{0.45\hsize}
            \subcaption{xglm-2.9B}
            \label{tab:cross_check_3}
            \scalebox{0.9}{
            \begin{tabular}{lrrrrrrr}\toprule
            &de &en &es &fr &ja &zh \\\midrule
            de &2000 &10 &6 &1 &14 &5 \\
            en &10 &2000 &13 &10 &8 &11 \\
            es &6 &13 &2000 &28 &12 &16 \\
            fr &1 &10 &28 &2000 &7 &12 \\
            ja &14 &8 &12 &7 &2000 &30 \\
            zh &5 &11 &16 &12 &30 &2000 \\
            \bottomrule
            \end{tabular}
            }
      \end{minipage}
      \hfill
      \begin{minipage}[t]{0.45\hsize}
            \subcaption{bloom-560m}
            \label{tab:cross_check_4}
            \scalebox{0.9}{
            \begin{tabular}{lrrrrrrr}\toprule
            &de &en &es &fr &ja &zh \\\midrule
            de &2000 &12 &19 &20 &12 &61 \\
            en &12 &2000 &76 &91 &61 &87 \\
            es &19 &76 &2000 &168 &70 &47 \\
            fr &20 &91 &168 &2000 &42 &56 \\
            ja &12 &61 &70 &42 &2000 &5 \\
            zh &61 &87 &47 &56 &5 &2000 \\
            \bottomrule
            \end{tabular}
            }
      \end{minipage} \\\\

      \begin{minipage}[t]{0.45\hsize}
            \subcaption{bloom-1b7}
            \label{tab:cross_check_5}
            \scalebox{0.9}{
            \begin{tabular}{lrrrrrrr}\toprule
            &de &en &es &fr &ja &zh \\\midrule
            de &2000 &10 &22 &15 &20 &59 \\
            en &10 &2000 &55 &88 &26 &59 \\
            es &22 &55 &2000 &140 &28 &10 \\
            fr &15 &88 &140 &2000 &24 &39 \\
            ja &20 &26 &28 &24 &2000 &8 \\
            zh &59 &59 &10 &39 &8 &2000 \\
            \bottomrule
            \end{tabular}
            }
      \end{minipage}
      \hfill
      \begin{minipage}[t]{0.45\hsize}
            \subcaption{bloom-3b}
            \label{tab:cross_check_6}
            \scalebox{0.9}{
            \begin{tabular}{lrrrrrrr}\toprule
            &de &en &es &fr &ja &zh \\\midrule
            de &2000 &8 &12 &12 &15 &43 \\
            en &8 &2000 &64 &45 &34 &46 \\
            es &12 &64 &2000 &98 &14 &26 \\
            fr &12 &45 &98 &2000 &18 &25 \\
            ja &15 &34 &14 &18 &2000 &11 \\
            zh &43 &46 &26 &25 &11 &2000 \\
            \bottomrule
            \end{tabular}
            }
      \end{minipage} \\\\

      \begin{minipage}[t]{0.45\hsize}
            \subcaption{Llama-2-7b}
            \label{tab:cross_check_7}
            \scalebox{0.9}{
            \begin{tabular}{lrrrrrrr}\toprule
            &de &en &es &fr &ja &zh \\\midrule
            de &2000 &20 &22 &12 &7 &15 \\
            en &20 &2000 &16 &14 &11 &11 \\
            es &22 &16 &2000 &34 &17 &8 \\
            fr &12 &14 &34 &2000 &13 &14 \\
            ja &7 &11 &17 &13 &2000 &85 \\
            zh &15 &11 &8 &14 &85 &2000 \\
            \bottomrule
            \end{tabular}
            }
      \end{minipage}
      \hfill
      \begin{minipage}[t]{0.45\hsize}
            \subcaption{Llama-2-13b}
            \label{tab:cross_check_8}
            \scalebox{0.9}{
            \begin{tabular}{lrrrrrrr}\toprule
            &de &en &es &fr &ja &zh \\\midrule
            de &2000 &14 &13 &11 &5 &23 \\
            en &14 &2000 &10 &10 &18 &9 \\
            es &13 &10 &2000 &23 &16 &1 \\
            fr &11 &10 &23 &2000 &9 &34 \\
            ja &5 &18 &16 &9 &2000 &80 \\
            zh &23 &9 &1 &34 &80 &2000 \\
            \bottomrule
            \end{tabular}
            }
      \end{minipage}

    \end{tabular}
    \caption{Cross-table check to count the number of overlapping language-specific neurons between languages.}
\end{table*}


\clearpage

\clearpage

\clearpage

\clearpage

\clearpage

\clearpage


\clearpage
\begin{table*}[t]\centering

\begin{tabular}{cc}

\begin{minipage}[t]{.45\textwidth}
\begin{center}
\begin{tabular}{lrrrrrr}\toprule
& &before &\multicolumn{3}{c}{after} \\\cmidrule{3-6}
& & &Top &Bottom &Both \\\midrule
xglm &en &40.0 &62.0 &77.0 &\textbf{89.0} \\
(564M) &de &0.0 &89.0 &31.0 &\textbf{95.0} \\
&fr &0.0 &86.0 &7.0 &\textbf{90.0} \\
&es &2.0 &71.0 &5.0 &\textbf{78.0} \\
&zh &7.0 &\textbf{82.0} &50.0 &79.0 \\
&ja &7.0 &92.0 &61.0 &\textbf{99.0} \\
&- &9.3 &80.3 &38.5 &\textbf{88.3} \\
\midrule
xglm &en &36.0 &23.0 &\textbf{43.0} &32.0 \\
(1.7B) &de &3.0 &84.0 &10.0 &\textbf{91.0} \\
&fr &1.0 &54.0 &5.0 &\textbf{70.0} \\
&es &3.0 &53.0 &9.0 &\textbf{69.0} \\
&zh &3.0 &59.0 &4.0 &\textbf{65.0} \\
&ja &9.0 &83.0 &17.0 &\textbf{87.0} \\
&- &9.2 &59.3 &14.7 &\textbf{69.0} \\
\midrule
xglm &en &31.0 &28.0 &\textbf{48.0} &42.0 \\
(2.9B) &de &2.0 &\textbf{92.0} &1.0 &88.0 \\
&fr &1.0 &60.0 &3.0 &\textbf{61.0} \\
&es &1.0 &67.0 &5.0 &\textbf{73.0} \\
&zh &5.0 &74.0 &6.0 &\textbf{85.0} \\
&ja &11.0 &\textbf{81.0} &3.0 &80.0 \\
&- &8.5 &67.0 &11.0 &\textbf{71.5} \\
\midrule
bloom &en &50.0 &69.0 &80.0 &\textbf{85.0} \\
(560m) &de &0.0 &34.0 &0.0 &\textbf{72.0} \\
&fr &13.0 &37.0 &85.0 &\textbf{93.0} \\
&es &9.0 &72.0 &69.0 &\textbf{97.0} \\
&zh &0.0 &24.0 &61.0 &\textbf{90.0} \\
&ja &0.0 &60.0 &0.0 &\textbf{74.0} \\
&- &12.0 &49.3 &49.2 &\textbf{85.2} \\
\bottomrule
\end{tabular}
\end{center}
\end{minipage}

\hspace{0.07\columnwidth} 
\hfill

\begin{minipage}[t]{.45\textwidth}
\begin{center}
\begin{tabular}{lrrrrrr}\toprule
& &before &\multicolumn{3}{c}{after} \\\cmidrule{3-6}
& & &Top &Bottom &Both \\\midrule
bloom &en &37.0 &78.0 &67.0 &\textbf{88.0} \\
(1b7) &de &0.0 &60.0 &0.0 &\textbf{86.0} \\
&fr &13.0 &80.0 &72.0 &\textbf{98.0} \\
&es &18.0 &44.0 &94.0 &\textbf{97.0} \\
&zh &6.0 &1.0 &89.0 &\textbf{90.0} \\
&ja &0.0 &67.0 &35.0 &\textbf{97.0} \\
&- &12.3 &55.0 &59.5 &\textbf{92.7} \\
\midrule
bloom &en &32.0 &41.0 &87.0 &\textbf{96.0} \\
(3b) &de &0.0 &44.0 &2.0 &\textbf{55.0} \\
&fr &15.0 &72.0 &72.0 &\textbf{93.0} \\
&es &19.0 &60.0 &94.0 &\textbf{95.0} \\
&zh &7.0 &24.0 &\textbf{91.0} &90.0 \\
&ja &0.0 &85.0 &1.0 &\textbf{87.0} \\
&- &12.2 &54.3 &57.8 &\textbf{86.0} \\
\midrule
Llama-2 &en &83.0 &82.0 &\textbf{89.0} &\textbf{89.0} \\
(7b) &de &0.0 &2.0 &6.0 &\textbf{23.0} \\
&fr &2.0 &1.0 &\textbf{8.0} &7.0 \\
&es &1.0 &4.0 &4.0 &\textbf{35.0} \\
&zh &0.0 &2.0 &4.0 &\textbf{50.0} \\
&ja &1.0 &1.0 &\textbf{12.0} &10.0 \\
&- &14.5 &15.3 &20.5 &\textbf{35.7} \\
\midrule
Llama-2 &en &64.0 &90.0 &81.0 &\textbf{94.0} \\
(13b) &de &3.0 &2.0 &3.0 &\textbf{16.0} \\
&fr &0.0 &\textbf{9.0} &1.0 &8.0 \\
&es &1.0 &1.0 &\textbf{5.0} &\textbf{5.0} \\
&zh &3.0 &\textbf{10.0} &6.0 &5.0 \\
&ja &2.0 &\textbf{6.0} &1.0 &4.0 \\
&- &12.2 &19.7 &16.2 &\textbf{22.0} \\
\bottomrule
\end{tabular}
\end{center}
\end{minipage}

\end{tabular}

\caption{Probability of language occurrence of generated texts before and after intervention. Top: intervention to only top-1000 neurons. Bottom: intervention to only bottom-1000 neurons. Both: intervention to both  top- and bottom-1000 neurons. The metric is accuracy measured by the FastText language identifier.}
\label{tab:intervention_language_appx}

\end{table*}


\clearpage

\clearpage

\clearpage


\clearpage

\clearpage

\clearpage

\clearpage

\clearpage

\clearpage

\clearpage

\clearpage

\clearpage

\clearpage

\clearpage


\clearpage
\begin{table*}[t]\centering
\begin{tabular}{lrrrrr}\toprule
model &language &\multicolumn{2}{c}{Accuracy} &\multicolumn{2}{c}{BLEU} \\\cmidrule{1-6}
& &Before &After &Before &After \\\midrule
xglm-564M &de &0.0 &\textbf{15.0} &0.0 &0.0 \\
xglm-564M &ja &0.0 &0.0 &0.0 &0.0 \\
xglm-564M &fr &0.0 &\textbf{3.0} &0.0 &0.0 \\
xglm-564M &zh &0.0 &2.0 &0.0 &0.0 \\
xglm-564M &- &0.0 &\textbf{5.0} &0.0 &0.0 \\
\midrule
xglm-1.7B &de &0.0 &\textbf{18.0} &0.0 &\textbf{1.4} \\
xglm-1.7B &ja &0.0 &\textbf{11.0} &0.0 &0.0 \\
xglm-1.7B &fr &0.0 &0.0 &0.0 &0.0 \\
xglm-1.7B &zh &0.0 &0.0 &0.0 &0.0 \\
xglm-1.7B &- &0.0 &\textbf{7.3} &0.0 &\textbf{0.3} \\
\midrule
xglm-2.9B &de &0.0 &\textbf{3.0} &0.0 &0.0 \\
xglm-2.9B &ja &0.0 &0.0 &0.0 &0.0 \\
xglm-2.9B &fr &0.0 &0.0 &0.0 &0.0 \\
xglm-2.9B &zh &0.0 &0.0 &0.0 &0.0 \\
xglm-2.9B &- &0.0 &\textbf{0.8} &0.0 &0.0 \\
\midrule
bloom-560m &de &0.0 &\textbf{4.0} &0.3 &\textbf{0.4} \\
bloom-560m &ja &0.0 &0.0 &0.0 &0.0 \\
bloom-560m &fr &0.0 &0.0 &0.0 &0.0 \\
bloom-560m &zh &0.0 &0.0 &0.0 &0.0 \\
bloom-560m &- &0.0 &\textbf{1.0} &0.1 &0.1 \\
\midrule
bloom-1b7 &de &0.0 &\textbf{35.0} &1.0 &\textbf{1.8} \\
bloom-1b7 &ja &0.0 &\textbf{8.0} &0.1 &\textbf{0.2} \\
bloom-1b7 &fr &0.0 &\textbf{2.0} &1.0 &\textbf{1.5} \\
bloom-1b7 &zh &0.0 &\textbf{3.0} &0.2 &\textbf{0.3} \\
bloom-1b7 &- &0.0 &\textbf{12.0} &0.6 &\textbf{0.9} \\
\midrule
bloom-3b &de &0.0 &\textbf{32.0} &0.7 &\textbf{1.0} \\
bloom-3b &ja &0.0 &\textbf{4.0} &0.1 &0.1 \\
bloom-3b &fr &0.0 &\textbf{6.0} &0.4 &\textbf{0.7} \\
bloom-3b &zh &0.0 &\textbf{1.0} &0.2 &0.2 \\
bloom-3b &- &0.0 &\textbf{10.8} &0.3 &\textbf{0.5} \\
\midrule
Llama-2-7b-hf &de &0.0 &\textbf{48.0} &1.2 &\textbf{12.5} \\
Llama-2-7b-hf &ja &1.0 &\textbf{57.0} &0.2 &\textbf{4.5} \\
Llama-2-7b-hf &fr &0.0 &\textbf{32.0} &1.0 &\textbf{11.1} \\
Llama-2-7b-hf &zh &3.0 &\textbf{82.0} &0.6 &\textbf{7.8} \\
Llama-2-7b-hf &- &1.0 &\textbf{54.8} &0.8 &\textbf{9.0} \\
\midrule
Llama-2-13b-hf &de &0.0 &\textbf{37.0} &0.6 &\textbf{10.0} \\
Llama-2-13b-hf &ja &4.0 &\textbf{75.0} &0.7 &\textbf{6.1} \\
Llama-2-13b-hf &fr &0.0 &\textbf{9.0} &0.7 &\textbf{4.7} \\
Llama-2-13b-hf &zh &40.0 &\textbf{96.0} &5.8 &\textbf{9.6} \\
Llama-2-13b-hf &- &11.0 &\textbf{54.3} &1.9 &\textbf{7.6} \\
\bottomrule
\end{tabular}
\caption{Results of conditional text generation for IWSLT2017.}
\label{tab:conditional_generation_iwslt}
\end{table*}

\begin{table*}[t]\centering
\begin{tabular}{lrrrrr}\toprule
model &language &\multicolumn{2}{c}{Accuracy} &\multicolumn{2}{c}{BLEU} \\\cmidrule{1-6}
& &Before &After &Before &After \\\midrule
xglm-564M &de &0.0 &\textbf{17.0} &0.0 &0.0 \\
xglm-564M &fr &0.0 &\textbf{1.0} &0.0 &0.0 \\
xglm-564M &zh &0.0 &\textbf{2.0} &0.0 &0.0 \\
xglm-564M &- &0.0 &\textbf{6.7} &0.0 &0.0 \\
\midrule
xglm-1.7B &de &0.0 &\textbf{16.0} &0.0 &\textbf{0.2} \\
xglm-1.7B &fr &0.0 &0.0 &0.0 &0.0 \\
xglm-1.7B &zh &0.0 &0.0 &0.0 &0.0 \\
xglm-1.7B &- &0.0 &\textbf{5.3} &0.0 &\textbf{0.1} \\
\midrule
xglm-2.9B &de &0.0 &0.0 &0.0 &0.0 \\
xglm-2.9B &fr &0.0 &0.0 &0.0 &0.0 \\
xglm-2.9B &zh &0.0 &0.0 &0.0 &0.0 \\
xglm-2.9B &- &0.0 &0.0 &0.0 &0.0 \\
\midrule
bloom-560m &de &0.0 &\textbf{4.0} &\textbf{1.4} &1.2 \\
bloom-560m &fr &0.0 &0.0 &0.5 &\textbf{0.6} \\
bloom-560m &zh &0.0 &0.0 &0.1 &0.1 \\
bloom-560m &- &0.0 &\textbf{1.3} &\textbf{0.7} &0.6 \\
\midrule
bloom-1b7 &de &0.0 &\textbf{37.0} &\textbf{2.9} &1.7 \\
bloom-1b7 &fr &0.0 &\textbf{9.0} &1.7 &\textbf{2.7} \\
bloom-1b7 &zh &0.0 &\textbf{34.0} &0.5 &\textbf{0.6} \\
bloom-1b7 &- &0.0 &\textbf{26.7} &1.7 &1.7 \\
\midrule
bloom-3b &de &0.0 &\textbf{19.0} &\textbf{3.1} &1.4 \\
bloom-3b &fr &0.0 &\textbf{7.0} &1.2 &\textbf{4.0} \\
bloom-3b &zh &0.0 &\textbf{4.0} &0.4 &\textbf{1.0} \\
bloom-3b &- &0.0 &\textbf{10.0} &1.5 &\textbf{2.1} \\
\midrule
Llama-2-7b-hf &de &2.0 &\textbf{53.0} &5.3 &\textbf{15.2} \\
Llama-2-7b-hf &fr &0.0 &\textbf{36.0} &2.1 &\textbf{13.2} \\
Llama-2-7b-hf &zh &12.0 &\textbf{86.0} &2.4 &\textbf{11.3} \\
Llama-2-7b-hf &- &4.7 &\textbf{58.3} &3.3 &\textbf{13.2} \\
\midrule
Llama-2-13b-hf &de &4.0 &\textbf{32.0} &3.3 &\textbf{9.7} \\
Llama-2-13b-hf &fr &1.0 &\textbf{15.0} &2.2 &\textbf{6.6} \\
Llama-2-13b-hf &zh &57.0 &\textbf{99.0} &13.5 &\textbf{18.9} \\
Llama-2-13b-hf &- &20.7 &\textbf{48.7} &6.3 &\textbf{11.7} \\
\bottomrule
\end{tabular}
\caption{Results of conditional text generation for WMT.}
\label{tab:conditional_generation_wmt}
\end{table*}

\begin{table*}[t]\centering
\begin{tabular}{lrrrrr}\toprule
model &language &\multicolumn{2}{c}{Accuracy} &\multicolumn{2}{c}{BLEU} \\\cmidrule{1-6}
& &Before &After &Before &After \\\midrule
xglm-564M &de &0.0 &\textbf{38.0} &0.0 &0.0 \\
xglm-564M &es &0.0 &\textbf{3.0} &0.0 &0.0 \\
xglm-564M &ja &0.0 &0.0 &0.0 &0.0 \\
xglm-564M &fr &0.0 &0.0 &0.0 &0.0 \\
xglm-564M &zh &0.0 &\textbf{1.0} &0.0 &0.0 \\
xglm-564M &- &0.0 &\textbf{8.4} &0.0 &0.0 \\
\midrule
xglm-1.7B &de &0.0 &\textbf{21.0} &0.0 &\textbf{1.3} \\
xglm-1.7B &es &0.0 &0.0 &0.0 &0.0 \\
xglm-1.7B &ja &0.0 &\textbf{4.0} &0.0 &0.0 \\
xglm-1.7B &fr &0.0 &0.0 &0.0 &0.0 \\
xglm-1.7B &zh &0.0 &0.0 &0.0 &0.0 \\
xglm-1.7B &- &0.0 &\textbf{5.0} &0.0 &\textbf{0.3} \\
\midrule
xglm-2.9B &de &0.0 &0.0 &0.0 &0.0 \\
xglm-2.9B &es &0.0 &0.0 &0.0 &0.0 \\
xglm-2.9B &ja &0.0 &0.0 &0.0 &0.0 \\
xglm-2.9B &fr &0.0 &0.0 &0.0 &0.0 \\
xglm-2.9B &zh &0.0 &0.0 &0.0 &0.0 \\
xglm-2.9B &- &0.0 &0.0 &0.0 &0.0 \\
\midrule
bloom-560m &de &0.0 &\textbf{6.0} &\textbf{0.4} &0.3 \\
bloom-560m &es &0.0 &\textbf{9.0} &0.2 &\textbf{0.6} \\
bloom-560m &ja &0.0 &\textbf{5.0} &0.0 &0.0 \\
bloom-560m &fr &0.0 &0.0 &0.5 &\textbf{0.6} \\
bloom-560m &zh &0.0 &0.0 &0.3 &0.3 \\
bloom-560m &- &0.0 &\textbf{4.0} &0.3 &0.3 \\
\midrule
bloom-1b7 &de &0.0 &\textbf{56.0} &1.3 &1.3 \\
bloom-1b7 &es &0.0 &\textbf{2.0} &1.2 &1.2 \\
bloom-1b7 &ja &0.0 &\textbf{6.0} &\textbf{0.2} &0.1 \\
bloom-1b7 &fr &0.0 &\textbf{16.0} &1.7 &\textbf{2.8} \\
bloom-1b7 &zh &0.0 &\textbf{21.0} &\textbf{0.3} &0.2 \\
bloom-1b7 &- &0.0 &\textbf{20.2} &0.9 &\textbf{1.1} \\
\midrule
bloom-3b &de &0.0 &\textbf{31.0} &\textbf{1.4} &0.8 \\
bloom-3b &es &0.0 &\textbf{7.0} &1.4 &\textbf{2.3} \\
bloom-3b &ja &0.0 &\textbf{7.0} &0.2 &0.2 \\
bloom-3b &fr &0.0 &\textbf{1.0} &1.8 &\textbf{1.8} \\
bloom-3b &zh &1.0 &\textbf{2.0} &0.4 &\textbf{0.4} \\
bloom-3b &- &0.2 &\textbf{9.6} &1.0 &\textbf{1.1} \\
\midrule
Llama-2-7b-hf &de &0.0 &\textbf{66.0} &2.6 &\textbf{17.7} \\
Llama-2-7b-hf &es &4.0 &\textbf{77.0} &3.3 &\textbf{16.6} \\
Llama-2-7b-hf &ja &0.0 &\textbf{58.0} &0.3 &\textbf{10.4} \\
Llama-2-7b-hf &fr &1.0 &\textbf{58.0} &4.1 &\textbf{21.5} \\
Llama-2-7b-hf &zh &1.0 &\textbf{76.0} &1.0 &\textbf{11.5} \\
Llama-2-7b-hf &- &1.2 &\textbf{67.0} &2.3 &\textbf{15.5} \\
\midrule
Llama-2-13b-hf &de &0.0 &\textbf{22.0} &1.5 &\textbf{8.8} \\
Llama-2-13b-hf &es &2.0 &\textbf{14.0} &1.8 &\textbf{4.3} \\
Llama-2-13b-hf &ja &7.0 &\textbf{54.0} &2.4 &\textbf{11.0} \\
Llama-2-13b-hf &fr &0.0 &\textbf{23.0} &1.6 &\textbf{10.5} \\
Llama-2-13b-hf &zh &20.0 &\textbf{93.0} &4.4 &\textbf{19.1} \\
Llama-2-13b-hf &- &5.8 &\textbf{41.2} &2.3 &\textbf{10.8} \\
\bottomrule
\end{tabular}
\caption{Results of conditional text generation for FLORES200.}
\label{tab:conditional_generation_flores200}
\end{table*}


\clearpage

\clearpage

\clearpage

\clearpage

\clearpage

\end{document}